\documentclass{article}

\PassOptionsToPackage{numbers, compress}{natbib}

\usepackage[preprint]{preprint}

\AtBeginDocument{\newgeometry{margin=1in}}  %

\usepackage[utf8]{inputenc}
\usepackage[T1]{fontenc}
\usepackage{hyperref}
\usepackage{url}
\usepackage{booktabs}
\usepackage{nicefrac}
\usepackage{microtype}
\usepackage{xcolor}

\usepackage{amsmath}
\usepackage{amssymb}
\usepackage{mathtools}
\usepackage{bm}

\usepackage{graphicx}
\usepackage{subcaption}

\usepackage{multirow}

\usepackage{tikz}
\usetikzlibrary{arrows.meta, positioning, calc, fit, backgrounds}

\usepackage{enumitem}
\usepackage{placeins}  %
\usepackage{float}     %

\newcommand{\sst}{\textsc{SST}}
\newcommand{\lsc}{\textsc{LSC}}

\title{State Stream Transformer (SST) V2: Parallel Training of Nonlinear Recurrence for Latent Space Reasoning}

\author{%
  Thea Aviss \\
  Fifth Dimension AI \\
  \texttt{thea@fifthdimensionai.com}
}

\begin{document}

\maketitle

\begin{abstract}
Current transformers discard their rich latent residual stream between positions, reconstructing latent reasoning context at each new position and leaving potential reasoning capacity untapped. The State Stream Transformer (SST) V2 enables parameter-efficient reasoning in continuous latent space through an FFN-driven nonlinear recurrence at each decoder layer, where latent states are streamed horizontally across the full sequence via a learned blend. This same mechanism supports continuous latent deliberation per position at inference time, dedicating additional FLOPs to exploring abstract reasoning before committing to a token. A two-pass parallel training procedure approximates the sequential recurrence, making co-training computationally practical. Hidden state analysis shows that the state stream facilitates reasoning through sharp, content-dependent reorganisations in continuous latent space; the LM head exposes the resulting latent belief states through the output distribution, while the state stream carries them forward to influence future positions. A learned probe shows that at the first generated token, the latent state already predicts whether the eventual answer will survive or break under additional latent computation for every subsequent position. Co-trained into an existing 27B backbone using only a small dataset of GSM8K examples and evaluated using an oracle to route each question to a depth of one to four recurrent forward passes per token, the \sst{} achieves an architectural capacity bound of $61.11\%$ on out-of-distribution GPQA-Diamond, a $+15.15$ point gain over a fine-tuning-matched baseline, and cuts that same baseline’s remaining GSM8K errors by $46\%$. Together, these results provide a method for efficiently training a nonlinear recurrence and show that the state stream provides additional reasoning capacity beyond that of an otherwise training-matched transformer.
\end{abstract}

\section{The Axes of Latent Space Reasoning}
\label{sec:intro}

When a person writes, both the written record and an evolving thought contribute to the next word. In the brain, these correspond to functionally dissociated systems: the language network handles linguistic comprehension and production but remains largely inactive during reasoning tasks, which instead engage a separate multiple demand network responsible for executive function, mathematics, and novel problem-solving~\cite{fedorenko2024language}. Transformers already mirror this dissociation architecturally, with the decoder layers computing rich latent representations through the residual stream~\cite{elhage2021circuits} at every position (the reasoning substrate) while the language model head maps these to token probabilities (the linguistic output). This supports the premise of latent reasoning: the computation that matters for reasoning is already happening in the residual stream, not in token space. But transformers lack the temporal continuity the biological system maintains between outputs. Recent neuroimaging of language production reveals that the brain's higher-level representations persist across word boundaries, evolving continuously rather than resetting at each word, while lower-level representations cycle rapidly, producing a hierarchy of temporal dynamics in which multiple representational levels overlap simultaneously~\cite{zhang2025thought}.

These findings suggest two axes along which latent reasoning can operate~\cite{zhu2025latent}, both potentially important for reasoning capacity. The \emph{vertical} axis provides depth of computation at each position, giving the model the ability to deliberate in continuous latent space before committing to a token. The \emph{horizontal} axis provides temporal continuity across positions, so that computation at one position shapes computation at the next through the evolving latent state, not only through the tokens it produced. The vertical axis determines how deeply the model can reason at each step; the horizontal axis determines whether that reasoning carries forward through the latent state or must be reconstructed at each new position. The architectures discussed below each pursue one axis.

Transformers perform rich latent computation through the residual stream at every position, but the residual stream flows only vertically through the layer stack, with no horizontal continuation between positions. At each new position, the model rebuilds its latent representations from the KV cache via the attention mechanism, constructing a new residual stream by querying into frozen projections of previous positions. This reconstruction via attention, without continuous latent computation between positions, may leave reasoning capacity untapped, as the neurological parallel above suggests. Preserving the latent computation across positions offers a new axis of compute within the transformer, orthogonal to both scaling and token-space reasoning, that could potentially compound with either.

The dominant approaches to improving reasoning in language models have focused on scaling model capacity and scaling inference-time computation. Scaling model parameters~\cite{kaplan2020scaling}, whether through dense models or mixture-of-experts architectures~\cite{shazeer2017moe}, increases per-pass capacity at the cost of larger models requiring more memory and compute. Scaling the reasoning process itself through extended chain-of-thought sequences~\cite{wei2022chain,openai2024o1,deepseek2025r1} improves reasoning through token space at the cost of substantially more compute per query as chains grow with problem difficulty. Inference-time compute scaling~\cite{snell2024scaling} also allocates additional compute budget at test time through strategies such as best-of-$n$ sampling.

Recent work on latent reasoning within transformers has pursued different axes separately. Geiping et al.~\cite{geiping2025scaling} iterate a shared recurrent block within a single position to deepen per-token computation before emission; Universal Transformers~\cite{dehghani2018universal} and looped transformers~\cite{giannou2023looped} similarly add depth by repeating layers within a token. Coconut~\cite{hao2024coconut} takes a different direction, inserting a dedicated deliberation phase between the question and the answer: within this phase, the final-layer hidden state at each position is fed forward as the input embedding for the next position, propagating latent state horizontally. Once token generation begins, however, the latent state is discarded between positions as in a standard transformer. Outside the language domain, HRM~\cite{wang2025hrm} supports the premise that latent recurrence and temporal continuity are powerful reasoning mechanisms, achieving complex task performance through hierarchical multi-timescale computation in small models trained from scratch. Its coupled modules exhibit both computational depth and temporal state continuity, but since neither operates via autoregressive generation, neither maps directly onto the per-position or cross-position axes defined here. These approaches all identify latent space as the site of reasoning, but none combines vertical compute at each position with horizontal latent continuity during token generation itself.

State space models such as Mamba~\cite{gu2024mamba} pursue horizontal continuity by replacing attention with selective state spaces whose input-dependent linear dynamics carry context across positions. Mamba's success demonstrates horizontal state persistence as a viable architectural approach. The \sst{} differs in two respects: its horizontal recurrence evolves through each layer's feedforward network, making it nonlinear, and it augments a transformer rather than replacing attention, preserving both mechanisms.

Prior work~\cite{aviss2025sst} introduced the state stream by modifying a frozen Llama 3.1 8B backbone and demonstrated both axes simultaneously, with reasoning improvements over the prompted base model. Without co-training, however, the model required multiple iterations per token for coherent output, meaning vertical iteration was needed for stability rather than being available purely for deliberation. The present work makes this co-training feasible through an improved architecture and a two-pass parallel training method for the nonlinear cross-position recurrence, together with a mechanistic analysis that traces how latent computation through the state stream causally shapes reasoning (improvements over the prior version detailed in Appendix~\ref{app:v1_comparison}).

\newpage
\section{Architecture of the SST}
\label{sec:architecture}

Our State Stream Transformer (SST) V2 is a latent space reasoning architecture that operates on both axes from a single mechanism (Figure~\ref{fig:architecture}). The \sst{} adds a state stream to each decoder layer of a standard transformer: a persistently evolving latent representation, produced by the layer's own feedforward network and streamed between forward passes alongside the KV cache, meaning the state stream flows between every forward pass. At a single pass per token, this carries the evolving latent state forward through the sequence, and at additional passes on the same position it dedicates additional FLOPs to deliberation, exploring abstract reasoning in latent space before committing to a token. The blend coefficients that govern this horizontal evolution are learned per-dimension during training, adding $L \times d = 333{,}312$ blend parameters plus an equal number of state-normalisation parameters to the base model while the reasoning computation itself is performed entirely by the existing pretrained feedforward weights.

The recurrence is nonlinear, passing through each layer's FFN at every step, so the state stream carries transformed reasoning rather than a compressed summary. This is what enables the \sst{} to be co-trained into an existing pretrained transformer backbone rather than requiring pretraining from scratch: rather than scaling parameters to increase per-pass capacity, the \sst{} extends the existing pretrained feedforward computation into reasoning by giving the network a horizontal state stream. The language modelling head sits outside the recurrence, connected only at the final emission point, keeping the cost of each iteration limited to the forward pass through the layer stack with no vocabulary projection or softmax at each step. At 27 billion parameters, this makes 1--4 iterations per position feasible without prohibitive overhead. At inference, the state stream stores a single $d$-dimensional vector per layer, $651$\,KB in bf16, a fixed cost that does not grow with sequence length.

\definecolor{blendcol}{HTML}{4A90D9}
\definecolor{ffncol}{HTML}{E8913A}
\definecolor{statecol}{HTML}{5BAA5B}
\definecolor{cascadecol}{HTML}{888888}
\definecolor{layerbg}{HTML}{F5F5F5}
\definecolor{itercol}{HTML}{7B68AE}

\begin{figure}[H]
\centering
\begin{tikzpicture}[
    scale=0.85, transform shape,
    >=Stealth,
    box/.style={draw, rounded corners=2pt, minimum height=0.65cm, minimum width=1.5cm,
                font=\footnotesize, thick, align=center},
    lscbox/.style={draw, rounded corners=2pt, minimum height=0.5cm, minimum width=1.1cm,
                   font=\scriptsize, thick, fill=statecol!15, text=statecol!80!black},
    arr/.style={->, thick},
    sarr/.style={->, thick, statecol, line width=1.1pt},
    every node/.style={font=\small},
  ]

  \begin{scope}[local bounding box=leftpanel]

    \node[font=\footnotesize\bfseries, anchor=south] at (2.2, 4.6) {(a) State stream at layer $l$};

    \node[font=\footnotesize, gray] at (0.7, 4.15) {position $t{-}1$};
    \node[font=\footnotesize, gray] at (3.7, 4.15) {position $t$};

    \node[font=\scriptsize] (attn0) at (0.0, 3.85) {from attention};
    \node[font=\tiny, gray, anchor=west] at (0.15, 3.55) {(+ residual)};
    \node[box, fill=blendcol!15, text=blendcol!80!black] (blend0) at (0.0, 2.9) {Blend ($\boldsymbol{\alpha}_l$)};
    \node[box, fill=ffncol!15, text=ffncol!80!black] (ffn0) at (0.0, 1.9) {FFN};
    \node[font=\tiny, gray, anchor=west] at (0.15, 1.4) {(+ residual)};
    \node[font=\scriptsize] (out0) at (0.0, 0.9) {to layer $l{+}1$};

    \draw[arr] (attn0) -- (blend0);
    \draw[arr] (blend0) -- (ffn0);
    \draw[arr] (ffn0) -- (out0);

    \node[lscbox, minimum width=1.3cm] (lsc) at (2.2, 1.9) {$\mathbf{C}_l$};

    \draw[arr, statecol] (ffn0.east) -- (lsc.west);

    \node[font=\scriptsize] (attnt) at (4.4, 3.85) {from attention};
    \node[font=\tiny, gray, anchor=west] at (4.55, 3.55) {(+ residual)};
    \node[box, fill=blendcol!15, text=blendcol!80!black] (blendt) at (4.4, 2.9) {Blend ($\boldsymbol{\alpha}_l$)};
    \node[box, fill=ffncol!15, text=ffncol!80!black] (ffnt) at (4.4, 1.9) {FFN};
    \node[font=\tiny, gray, anchor=west] at (4.55, 1.4) {(+ residual)};
    \node[font=\scriptsize] (outt) at (4.4, 0.9) {to layer $l{+}1$};

    \draw[arr] (attnt) -- (blendt);
    \draw[arr] (blendt) -- (ffnt);
    \draw[arr] (ffnt) -- (outt);

    \draw[sarr] (lsc.north) |- (blendt.west);

    \node[font=\large, statecol!60!black] at (-1.6, 2.9) {$\cdots$};
    \draw[sarr] (-1.3, 2.9) -- (blend0.west);

    \draw[sarr] (ffnt.east) -- ++(0.5, 0);
    \node[font=\large, gray] at (5.5, 1.7) {$\cdots$};

  \end{scope}

  \begin{scope}[xshift=8.0cm, local bounding box=rightpanel]

    \node[font=\footnotesize\bfseries, anchor=south] at (1.4, 4.6) {(b) Full model cascade};

    \node[font=\scriptsize] (emb) at (1.4, 4.1) {token embedding};

    \node[box, fill=layerbg, minimum width=2.8cm, minimum height=0.6cm]
      (layer0) at (1.4, 3.4) {\footnotesize Blend $\rightarrow$ FFN};
    \node[font=\scriptsize, gray, anchor=east] at ($(layer0.west) + (-0.15, 0)$) {$l{=}0$};
    \draw[sarr]
      ($(layer0.east) + (0.05, 0)$) -- ++(0.6, 0)
      node[right, font=\scriptsize, statecol!80!black] {$\mathbf{C}_{0}$};

    \draw[arr] (emb) -- (layer0);

    \node[box, fill=layerbg, minimum width=2.8cm, minimum height=0.6cm, below=0.4cm of layer0]
      (layermid) {\footnotesize Blend $\rightarrow$ FFN};
    \node[font=\scriptsize, gray, anchor=east] at ($(layermid.west) + (-0.15, 0)$) {$l{=}1{\ldots}60$};
    \draw[sarr]
      ($(layermid.east) + (0.05, 0)$) -- ++(0.6, 0)
      node[right, font=\scriptsize, statecol!80!black] {$\mathbf{C}_{1{\ldots}60}$};
    \draw[arr, cascadecol] (layer0) -- (layermid);

    \node[box, fill=layerbg, minimum width=2.8cm, minimum height=0.6cm, below=0.4cm of layermid]
      (layerN) {\footnotesize Blend $\rightarrow$ FFN};
    \node[font=\scriptsize, gray, anchor=east] at ($(layerN.west) + (-0.15, 0)$) {$l{=}61$};
    \draw[sarr]
      ($(layerN.east) + (0.05, 0)$) -- ++(0.6, 0)
      node[right, font=\scriptsize, statecol!80!black] {$\mathbf{C}_{61}$};
    \draw[arr, cascadecol] (layermid) -- (layerN);

    \node[font=\scriptsize\itshape, statecol!80!black, rotate=-90, anchor=south]
      at ($(layermid.east) + (1.55, 0)$) {state stream};

    \begin{scope}[on background layer]
      \fill[itercol!8, rounded corners=5pt]
        ($(layer0.north west) + (-0.6, 0.2)$) rectangle ($(layerN.south east) + (0.15, -0.2)$);
    \end{scope}
    \node[font=\scriptsize, itercol, anchor=north east]
      at ($(layerN.south east) + (0.1, -0.05)$) {$\times\, i$};

    \node[box, fill=white, minimum width=1.8cm, below=0.6cm of layerN] (lmhead) {\footnotesize lm\_head};
    \draw[arr] (layerN) -- (lmhead);
    \node[font=\scriptsize, below=0.4cm of lmhead] (tok) {output token};
    \draw[arr] (lmhead) -- (tok);

  \end{scope}

\end{tikzpicture}
\caption{The state stream mechanism. Green arrows denote the \emph{state stream}: the per-layer latent states ($\mathbf{C}_l$) that persist across positions and iterations.
\textbf{(a)}~At each layer, the \lsc{} stores the post-feedforward output. At the next position, this state is blended into the hidden representation with learned per-dimension strength $\boldsymbol{\alpha}_l$ before the feedforward network. The feedforward output updates the \lsc{} for the next position.
\textbf{(b)}~The mechanism operates at every layer with unique parameters. Each layer's output feeds into the next, creating a vertical cascade of $L$ coupled recurrences. The state stream flows horizontally at every layer simultaneously. At inference, the full model is run $i$ times per token without advancing the position index; previous positions' KV entries are unchanged, while the current position's entries are updated as the state stream evolves. The state stream carries information between iterations. The language modelling head produces a token only after the final iteration.}
\label{fig:architecture}
\end{figure}
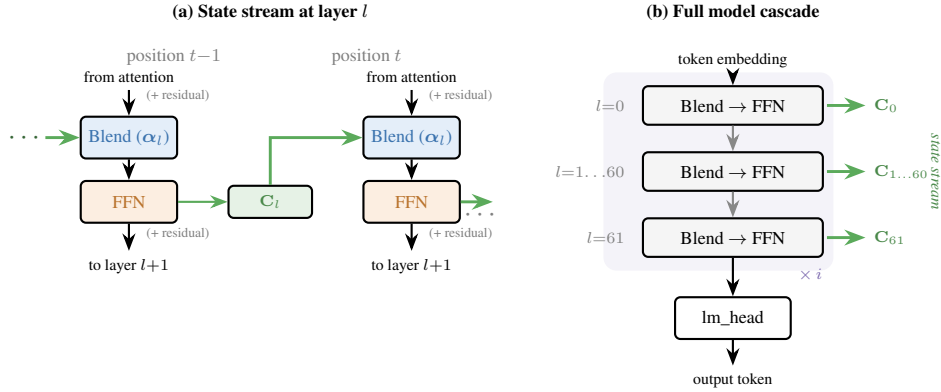

\subsection{State stream recurrence}
\label{sec:recurrence}

We describe the mechanism for a single decoder layer $l \in \{0, \ldots, L{-}1\}$ at sequence position $t$. Let $\mathbf{x}_{l,t} \in \mathbb{R}^d$ denote the input to layer $l$ (the output of layer $l{-}1$, or the token embedding for $l = 0$), $\mathbf{h}_{l,t} \in \mathbb{R}^d$ the post-attention hidden state, $\mathbf{C}_{l,t} \in \mathbb{R}^d$ the latent state cache (\lsc{}) at layer $l$ and position $t$, and $\boldsymbol{\alpha}_l \in \mathbb{R}^d$ the learned per-dimension blend strength for layer $l$.

\paragraph{Post-attention residual.}
The attention output is added to the residual in the standard way:
\begin{equation}
  \mathbf{h}_{l,t} = \mathbf{x}_{l,t} + \mathrm{Attn}_l\!\bigl(\mathrm{RMSNorm}(\mathbf{x}_{l,t})\bigr).
  \label{eq:post_attn}
\end{equation}

\paragraph{Latent state blend.}
The \lsc{} linearly interpolates (or \emph{blends}) between the post-attention hidden state and the previous position's latent state, element-wise:
\begin{equation}
  \tilde{\mathbf{h}}_{l,t}
    = \bigl(\mathbf{1} - \boldsymbol{\alpha}_l\bigr) \odot \mathbf{h}_{l,t}
    + \boldsymbol{\alpha}_l \odot \mathrm{RMSNorm}\!\bigl(\mathbf{C}_{l,t-1}\bigr),
  \label{eq:blend}
\end{equation}
where $\mathbf{C}_{l,t-1} \in \mathbb{R}^d$ is the latent state from position $t{-}1$ at layer~$l$, and $\boldsymbol{\alpha}_l \in \mathbb{R}^d$ is a per-dimension learned blend vector (Section~\ref{sec:alpha}). The blend is applied \emph{post-attention, pre-feedforward}: the feedforward network processes the blended representation $\tilde{\mathbf{h}}_{l,t}$, not the raw attention output.

\paragraph{Feedforward network and state update.}
The feedforward network operates on the blended hidden state:
\begin{equation}
  \mathbf{o}_{l,t}
    = \tilde{\mathbf{h}}_{l,t}
    + \mathrm{FFN}_l\!\bigl(\mathrm{RMSNorm}(\tilde{\mathbf{h}}_{l,t})\bigr).
  \label{eq:ffn}
\end{equation}
The latent state is then updated with the full post-feedforward output:
\begin{equation}
  \mathbf{C}_{l,t} = \mathbf{o}_{l,t}.
  \label{eq:state_update}
\end{equation}
The state update stores only the most recent position's post-feedforward output. There is no explicit retention of earlier states, but the influence of earlier positions persists through transformation. Position $t{-}2$'s state was blended into position $t{-}1$'s hidden representation and processed by $t{-}1$'s feedforward network. The result of that computation is what position $t$ receives as its state. The influence of position $t{-}k$ on position $t$ has passed through $k$ successive blend-feedforward cycles, each transforming it through the nonlinearity of a different position's computation. The effect is a sliding window of weighted decay across positions~\citep{aviss2025sst}, but unlike a fixed exponential decay, the rate and character of the attenuation is content-dependent; determined by what the feedforward network computed at each intervening position.

\paragraph{Vertical cascade.}
The output $\mathbf{o}_{l,t}$ serves as the input $\mathbf{x}_{l+1,t}$ to the next layer. Because each layer blends its own state before its own feedforward network, the perturbation introduced at layer~$l$ propagates through all subsequent layers' attention and feedforward computations. A blend at layer~0 affects the input to layers 1 through $L{-}1$. This vertical cascade is what couples the $L$ independent horizontal recurrences into a single coherent computation.

\paragraph{Per-dimension learned blend strength.}
\label{sec:alpha}
Each layer $l$ has a learned blend vector $\boldsymbol{\alpha}_l \in \mathbb{R}^d$, parameterised as:
\begin{equation}
  \boldsymbol{\alpha}_l
    = \alpha_{\min}
    + \bigl(\alpha_{\max} - \alpha_{\min}\bigr)
      \cdot \sigma\!\bigl(\boldsymbol{\theta}_l\bigr),
  \label{eq:alpha}
\end{equation}
where $\boldsymbol{\theta}_l \in \mathbb{R}^d$ is a learned logit vector and $\sigma$ is the sigmoid function, constraining $\boldsymbol{\alpha}_l \in [\alpha_{\min}, \alpha_{\max}]$. We set $\alpha_{\min} = 0.015$ and $\alpha_{\max} = 0.10$. The logits are initialised at $\theta_l^{(0)} = -1.8$, corresponding to $\sigma(-1.8) \approx 0.142$ and an initial blend strength of approximately $0.027$ per dimension. This initialisation was identified through an exploratory untrained ablation on two different base models and is validated by a trained ablation (Appendix~\ref{app:ablations}); optimising the bias value is not within the scope of this work, and other values may yield different performance.

The blend is non-bypassable: every dimension always blends at least $\alpha_{\min}$. Across $L = 62$ layers, the architecture introduces $L \times d = 333{,}312$ learned blend parameters, plus $L \times d$ RMSNorm parameters for state normalisation.

\subsection{Iterative refinement at inference}
\label{sec:inference_iteration}

Due to the recurrence of the state stream every forward pass, the architecture supports \emph{iterative refinement}: multiple forward passes through the full $L$-layer stack per token, without advancing the position index. Previous positions' KV cache entries remain unchanged and the current position's entries are updated on each iteration as the evolving latent states produce different hidden representations at each layer. As each iteration repeats the blend--feedforward--update cycle at every layer, using the states produced by the previous iteration, this continues computation and dedicates more FLOPs per position. This subsequently compounds across positions via the state stream.

Without access to the output distribution during iterative refinement, the model has no native mechanism to signal completion in its $d$-dimensional latent space, unlike approaches that detect convergence through token-space metrics such as KL divergence on the output distribution~\cite{geiping2025scaling}; there is no analogue of a \texttt{</think>} token. The number of iterations is therefore controlled externally. We address this limitation through a staged compute evaluation methodology that measures the model's full capacity across iteration depths (Section~\ref{sec:evaluation}), and through a learned halting probe that demonstrates the feasibility of autonomous iteration depth selection from the position~0 latent state alone, without requiring token-space evaluation (Section~\ref{sec:halting_probe}).

\section{Training}
\label{sec:training}
As the state stream recurrence (Section~\ref{sec:recurrence}) introduces a nonlinear full-sequence cross-position dependency at every layer, training cannot be simply parallelised across positions. Training as-is would therefore have to unroll the full sequence end-to-end via BPTT which would be prohibitively expensive (Appendix~\ref{app:parallelisation} analyses the problem and alternative approaches in more detail). This is the same sequential bottleneck that motivated the move from recurrent architectures to parallelisable transformers~\cite{vaswani2017attention}.

However, an exact solution to the parallelisation problem may not be necessary. Supervised fine-tuning already accepts an approximation of the autoregressive recurrence: teacher forcing substitutes ground-truth tokens for the model's own predictions, breaking the sequential dependency across positions and allowing all positions to be computed in parallel. The resulting train-inference mismatch (exposure bias~\cite{ranzato2015sequence}) is well understood and widely tolerated.

The state stream's cross-position recurrence is a second sequential dependency in the same model. If the token-level recurrence can be approximated by substituting ground truth, the state-level recurrence can be approximated in the same spirit. But if approximate states are available at every position, the blend becomes a linear interpolation between the hidden state and a known quantity, and the cross-position propagation of these approximate states reduces to a linear recurrence. Linear recurrences can be parallelised efficiently via the associative scan that underlies S4~\cite{gu2021s4}, S5~\cite{smith2022s5}, and Mamba~\cite{gu2024mamba}. Section~\ref{sec:two_pass} repurposes a basic form of this into a concrete two-pass training method and analyses the approximation error.

The model is fine-tuned on a synthetic CodeACT~\cite{wang2024codeact} reformulation of GSM8K~\cite{cobbe2021gsm8k} (6{,}579 training examples) on a single NVIDIA RTX PRO 6000, trained via QLoRA~\cite{dettmers2023qlora} with the \lsc{} parameters at full precision. Dataset details, training setup, and full hyperparameters are given in Appendix~\ref{app:training}.

\subsection{Two-pass parallel training}
\label{sec:two_pass}
The two-pass method replaces the sequential state recurrence with a parallelisable approximation. The first pass runs the model without the blend, producing post-feedforward outputs at every layer and position simultaneously. These are propagated across positions via a per-layer associative scan. The second pass re-runs the model with the blend enabled, using the propagated states, completing the approximation. Loss is computed on the second pass only, but gradients are collected for both passes.

\paragraph{Pass 1: approximate states.}
The first forward pass disables the blend step (Eq.~\ref{eq:blend}), eliminating the cross-position state dependency. Every position at every layer computes the standard post-attention residual and feedforward operations (Eqs.~\ref{eq:post_attn},~\ref{eq:ffn}) without a latent state. Each layer~$l$ stores its post-feedforward output $\mathbf{o}^{(1)}_{l,t}$ for all $T$ positions.

\paragraph{State propagation.}
The stored outputs are propagated across positions at each layer~$l$ via a linear recurrence:
\begin{equation}
  \mathbf{S}_{l,t} = \mathbf{A}_l \odot \mathbf{S}_{l,t-1} + \mathbf{B}_{l,t}, \quad \mathbf{S}_{l,0} = \mathbf{0},
  \label{eq:scan_recurrence}
\end{equation}
parallelised with an associative scan in $O(\log T)$ steps. The $L$ per-layer scans are independent and can be executed in parallel. We set $\mathbf{A}_l = \mathbf{0}$ and $\mathbf{B}_{l,t} = \mathbf{o}^{(1)}_{l,t}$, and shift the result right by one position: position~$t$ receives position~$t{-}1$'s pass~1 output as its state, and position~0 receives a zero vector. This delivers each position's pass-1 output to the next position, mirroring the state update of the sequential recurrence (Eq.~\ref{eq:state_update}). The scan provides the approximate states that enable pass~2 to execute the recurrence in parallel.

\paragraph{Pass 2: forward with blend.}
The second forward pass completes the recurrence approximation by re-running the full model with the blend enabled, using the propagated state $\mathbf{S}_{l,t-1}$ as the recurrent input in Eq.~\ref{eq:blend}:
\begin{equation}
  \tilde{\mathbf{h}}^{(2)}_{l,t}
    = \bigl(\mathbf{1} - \boldsymbol{\alpha}_l\bigr) \odot \mathbf{h}^{(2)}_{l,t}
    + \boldsymbol{\alpha}_l \odot \mathrm{RMSNorm}\!\bigl(\mathbf{S}_{l,t-1}\bigr).
  \label{eq:pass2_blend}
\end{equation}
Because the scan states are pre-computed and available at every position, all positions can again be processed in parallel. The vertical cascade (Section~\ref{sec:recurrence}) is preserved: layer~$l$'s blended output is the input to layer~$l{+}1$'s attention, so the effect of blending compounds through the full depth of the model. Loss is computed on pass~2's output through the language modelling head.

\paragraph{Approximation error.}
In the sequential recurrence, position~$t$'s state $\mathbf{C}_{l,t}$ depends on the full recursive history: $\mathbf{C}_{l,t-1}$ was itself computed with a blend from $\mathbf{C}_{l,t-2}$. The two-pass method substitutes pass~1 outputs $\mathbf{o}^{(1)}_{l,t}$, which were computed without the blend. Omitting the blend perturbs the FFN input by $O(\alpha)$. The FFN with residual connection is a composition of linear maps (finite weight matrices) and a Lipschitz activation (GELU~\cite{hendrycks2016gelu}, tanh approximation; $L < 1.13$), giving it a finite Lipschitz constant $L_l$, so the post-FFN output perturbation is also $O(\alpha)$:
\begin{equation}
  \mathbf{o}^{(1)}_{l,t} = \mathbf{C}_{l,t} + O(\alpha).
  \label{eq:approx_bound}
\end{equation}
When pass~2 blends this $O(\alpha)$ approximation with weight $\boldsymbol{\alpha}_l$, the error is $\boldsymbol{\alpha}_l \odot O(\alpha) = O(\alpha^2)$. With learned $\alpha \in [0.024, 0.035]$, $\alpha^2 \in [5.8 \times 10^{-4},\, 1.2 \times 10^{-3}]$. The two-pass blend matches the true sequential computation to first order in~$\alpha$. The full derivation is given in Appendix~\ref{app:approx_bound}.

\paragraph{Co-adaptation.}
The $O(\alpha^2)$ bound assumes fixed weights. In practice, the LoRA~\cite{hu2021lora} adapters, blend parameters, and state normalisation weights are all trained jointly across both passes. Gradients flow from pass~2's loss through the blend, through the scan, and into pass~1's feedforward outputs. The model learns pass~1 outputs that are maximally useful when propagated and blended into pass~2. The two passes co-adapt rather than one approximating the other, tightening the effective approximation beyond the fixed-weight bound. This allows any suitable pretrained transformer with a gated-FFN backbone to be turned into an \sst{} via fine-tuning.

\paragraph{Validation against exact recurrence.}
To test whether the local approximation remains faithful after propagation through the complete 62-layer stack, we directly compare the trained checkpoint's two-pass computation with the exact nonlinear recurrence. We use the first four examples in validation-file order, truncated to the first 24 supervised tokens of each example, giving 96 supervised tokens and 92 positions at which the recurrence is active. Both paths use full-sequence causal attention and bf16 arithmetic, matching training. The exact reference serialises only the within-layer blend--FFN recurrence across supervised positions, avoiding confounds from KV-cache versus full-sequence attention kernels. An instrumented implementation of the two-pass path was bitwise identical to the native training implementation on all four examples.

\begin{table}[H]
\centering
\small
\caption{\textbf{Two-pass approximation versus exact nonlinear recurrence.} Aggregate results across 96 supervised tokens and 92 recurrent positions from the first four validation examples.}
\label{tab:exact_recurrence_validation}
\begin{tabular}{lc}
\toprule
Measurement & Result \\
\midrule
Local substitution error / recurrent signal & $0.491\%$ \\
Propagated blended-input error / recurrent signal & $31.702\%$ \\
Final hidden-state relative RMS & $2.683\%$ \\
Final-logit relative RMS & $1.670\%$ \\
Mean KL divergence & $1.95 \times 10^{-4}$ \\
Greedy argmax agreement & $92/92$ \\
Mean absolute teacher-forced loss difference & $0.00210$ \\
\bottomrule
\end{tabular}
\end{table}

The local measurement isolates the substituted recurrent state while holding the exact upstream hidden state fixed; the propagated measurement includes approximation differences accumulated through preceding layers. The internal difference is measurably amplified through the stack, reaching $31.70\%$ of the recurrent signal at the blended inputs, but this does not produce a correspondingly large end-to-end mismatch: final-logit relative RMS remains $1.67\%$, mean KL divergence is $1.95 \times 10^{-4}$, and all $92$ greedy argmax decisions agree. The mean absolute loss difference is $0.00210$. The comparison includes the numerical difference introduced by evaluating recurrent FFNs position-by-position rather than as a sequence batch. These results directly validate the two-pass approximation after propagation through the full trained model on the evaluated sequences. They are a checkpoint-level fidelity test, not a uniform error bound over arbitrary sequence lengths, checkpoints, or backbones.

\paragraph{Measured training cost.}
Training cost comprises the second transformer pass, the retained recurrent tensors, the scan workspace, and the split backward pass. We therefore compare the complete optimised training paths rather than inferring cost from the number of forward passes. Both the \sst{} and matched baseline use chunked cross-entropy, offloaded gradient checkpointing, and tiled MLP forward passes. Table~\ref{tab:training_cost} reports their measured cost on the same NVIDIA RTX PRO 6000 across 310 paired optimiser steps with identical 16-example accumulated batches.

\begin{table}[H]
\centering
\small
\caption{\textbf{Measured cost of the optimised training paths.} Step-level measurements use 310 paired optimiser steps with identical accumulated batches.}
\label{tab:training_cost}
\begin{tabular}{@{}lccc@{}}
\toprule
Measurement & Baseline & \sst{} & Ratio \\
\midrule
Transformer passes / micro-batch & 1 & 2 & $2.00\times$ \\
Mean optimiser-step time & $12.76$\,s & $26.51$\,s & $2.08\times$ \\
Mean peak allocated VRAM & $24.60$\,GiB & $25.70$\,GiB & $1.04\times$ \\
Peak VRAM at $T=6005$ & $42.92$\,GiB & $51.42$\,GiB & $1.20\times$ \\
Mean validation/checkpoint interval & $122.10$\,s & $288.35$\,s & $2.36\times$ \\
Active runtime to best checkpoint & $1.94$\,h & $6.29$\,h & $3.24\times$ \\
\bottomrule
\end{tabular}
\end{table}

The end-to-end optimiser step is therefore $2.08\times$ the baseline: the scan, recurrent tensor management, split backward, and training diagnostics together add only $0.08$ baseline-step equivalents beyond the second transformer pass. Despite this additional machinery, mean peak allocation increases by $1.10$\,GiB ($4.5\%$), while the longest sequence encountered in the paired steps increases peak allocation by $8.50$\,GiB ($19.8\%$). The additional memory comes primarily from retaining pass~1 outputs and propagated states at every layer, so it grows linearly with sequence length: these tensors occupy $7.46$\,GiB for the longest sequence encountered and $10.17$\,GiB at the configured maximum length. During the split backward, detached scan-state gradients add transient storage, while the checkpointed scan avoids retaining its internal autograd intermediates by recomputing them during backward.

The active-runtime ratio is not a per-step architectural multiplier: it also reflects that the selected \sst{} checkpoint occurs at step~410 rather than step~280 and includes 41 rather than 28 validation/checkpoint intervals. The measurements conservatively describe the runs that produced the evaluated checkpoints. In particular, the \sst{} path also computed scan-state norms, blend magnitudes, and full-vocabulary pass-1/pass-2 top-$k$ overlap diagnostics during every training micro-batch; these diagnostics are not required by the training method. Appendix~\ref{app:why_two_passes} discusses why additional training passes would be counterproductive.

\subsection{Matched fine-tuned baseline}
\label{sec:baseline_ft}

To attribute downstream evaluation results to the \lsc{} mechanism rather than the fine-tuning data, we train a matched baseline that applies QLoRA adapters to the unmodified Gemma~3 backbone (no \lsc{} parameters, no two-pass forward). The baseline matches the \sst{} on every training factor we hold constant: dataset, optimiser, learning-rate schedules, deterministic example ordering, and attention implementation (eager attention throughout training; scaled dot-product attention is used only at inference). Architecture is the only varied factor, so the \sst{}--baseline delta on downstream evaluations (Section~\ref{sec:evaluation}) isolates the contribution of the state stream mechanism. Both models converge to comparable validation loss (Appendix Figure~\ref{fig:sst_vs_baseline_loss}).

\subsection{Learned blend coefficients}
\label{sec:learned_alpha}

To verify that training exploited the per-dimension, per-layer freedom of the \lsc{} parameterisation, we examine the learned $\alpha_l$ relative to their initialisation. Because every coefficient begins at $\alpha_{\mathrm{init}}$ (Section~\ref{sec:alpha}), any deviation $\alpha_{l,d} - \alpha_{\mathrm{init}}$ is by construction a move from the no-learning state; we use this fixed reference (not the data-derived per-layer or across-layer mean) for all subsequent measurements of adaptation.

\begin{figure}[H]
  \centering
  \begin{subfigure}[t]{0.49\linewidth}
    \centering
    \includegraphics[width=\linewidth]{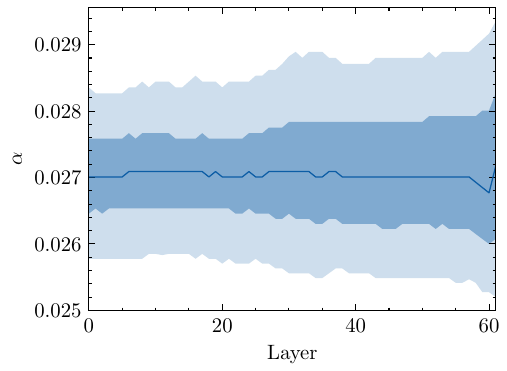}
    \caption{Per-layer $\alpha$ distributions}
    \label{fig:alpha_quantile_ribbon}
  \end{subfigure}
  \hfill
  \begin{subfigure}[t]{0.49\linewidth}
    \centering
    \includegraphics[width=\linewidth]{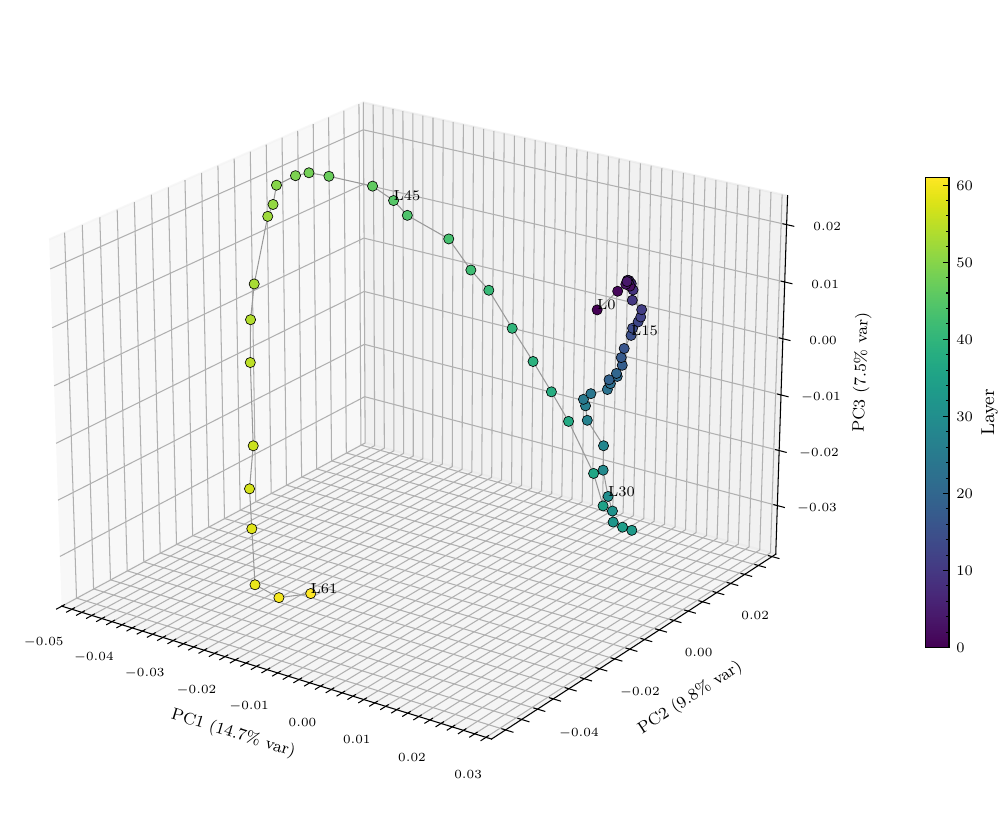}
    \caption{Depth-ordered trajectory in PC space}
    \label{fig:alpha_pca_trajectory}
  \end{subfigure}
  \caption{\textbf{Learned blend coefficient structure.} \textbf{(a)} Within-layer percentile bands (p5--p95 outer, p25--p75 inner, median line) of the 5{,}376 per-dimension blend coefficients $\alpha_{l,d}$ at each of the 62 layers. At initialisation, every value was $\alpha_{\mathrm{init}} \approx 0.027$; the visible spread is the consequence of training. \textbf{(b)} Each layer's deviation vector $\boldsymbol{\alpha}_l - \alpha_{\mathrm{init}}$ projected onto the top three principal components (PC1, PC2, PC3 explain 14.7\%, 9.8\%, 7.5\% of layer-to-layer variance). Points are coloured by layer index and connected by line segments; the smooth depth-ordered trajectory indicates stack-wide coherence rather than independent optimisation.}
  \label{fig:alpha_combined}
\end{figure}

Training preserved the magnitude prior set by initialisation but exploited the per-dimension freedom on top of it: per-layer means stay near $\alpha_{\mathrm{init}}$ at every depth while the within-layer distributions widen non-monotonically (Figure~\ref{fig:alpha_quantile_ribbon}). The architecture's $L \times d$ degrees of freedom were not used to rescale blend strength globally; they were used to develop a per-dimension pattern at every layer. The adaptation is layer-specific, not a single shared per-dimension bias replicated across layers. Inspecting the raw $\alpha_l$ vectors at representative depths reveals distinct per-dimension textures at each layer, with the magnitude of adaptation varying systematically and non-monotonically with depth (Appendix Figures~\ref{fig:alpha_panels_raw_selected},~\ref{fig:alpha_panels_deviation_selected},~\ref{fig:alpha_specialisation_profile}).

Projected onto the top three principal components of their deviation matrix $D_{l,d} = \alpha_{l,d} - \alpha_{\mathrm{init}}$, the 62 per-layer adaptation patterns form a smooth depth-ordered trajectory (Figure~\ref{fig:alpha_pca_trajectory}), with adjacent layers neighbouring each other in parameter space. This is consistent with stack-wide coherence through gradient coupling across the forward cascade.

Together, these observations establish that the per-dimension, per-layer freedom of the \lsc{} parameterisation was used by training as designed. Each of the 62 layers learned a distinct adaptation pattern (Appendix~\ref{app:alpha_analysis}), the magnitude of those adaptations varies systematically with depth, and the resulting patterns are organised into a coherent stack-wide structure.

\FloatBarrier

\section{Latent computation dynamics}
\label{sec:mechanism}

The per-layer state stream substantially alters the model's behaviour, with the \sst{} producing fundamentally different output from the unmodified backbone and different iteration depths yielding substantively different responses. Through $L$ coupled nonlinear recurrences, the latent space actively shapes computation at every position, playing a more direct role than in a standard transformer where no per-layer information persists between positions. As the model performs computational work in continuous $d$-dimensional space before ever projecting into discrete tokens, characterising the mechanics of this latent computation is essential to understanding whether and how the state stream performs structured reasoning. This section traces the causal chain from the state stream's effect on hidden state representations through to changes in the output distribution, and what those dynamics reveal about the nature of latent space reasoning.

\subsection{Punctuated equilibrium in latent space}
\label{sec:punctuated_equilibrium}

\begin{figure}[H]
  \centering
  \includegraphics[width=\linewidth]{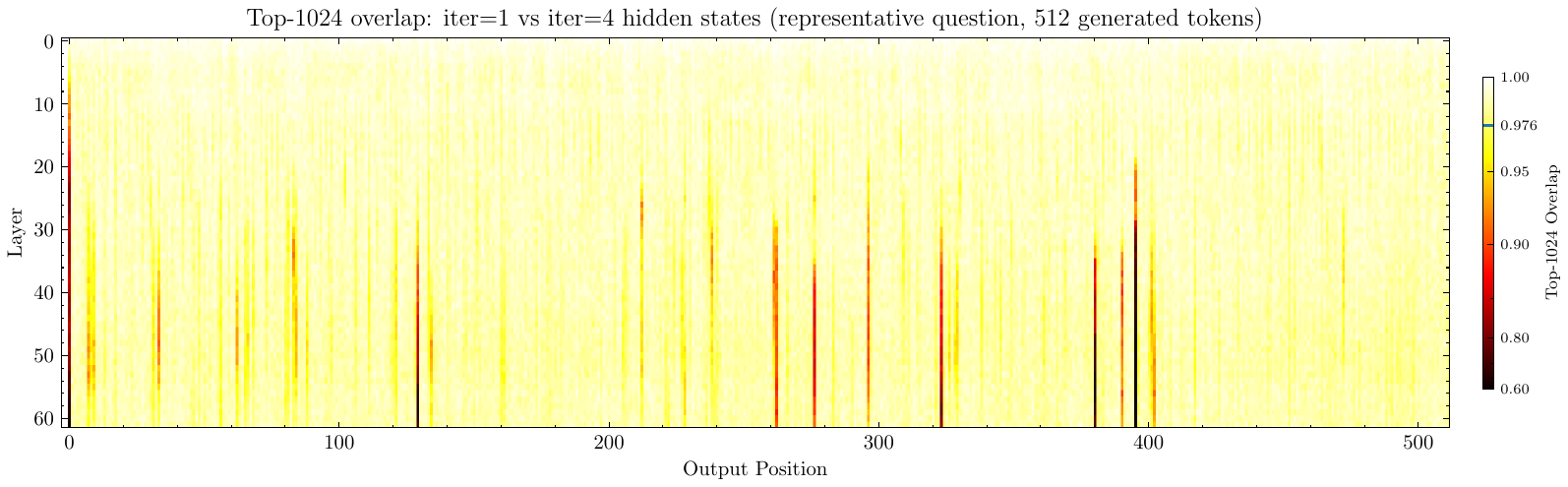}
  \caption{\textbf{Top-1024 overlap between iter=1 and iter=4 hidden states across the layer stack and generated sequence}, for a representative GPQA-Diamond question. Bright cells = stable; dark cells = low-overlap reorganisation. Low-overlap positions appear as vertical streaks cascading through the depth of the model; stable positions remain near 1.0 at every layer. The Gaussian mixture model (GMM) crossover threshold of 0.976 is marked on the colour bar. A zoom of the first 10 positions for a separate question is given in Appendix Figure~\ref{fig:overlap_heatmap_zoom}.}
  \label{fig:overlap_heatmap}
\end{figure}

Examining hidden state representations across iteration depths reveals an immediate visual pattern (Figure~\ref{fig:overlap_heatmap}): at most positions in a sequence, the hidden states are remarkably stable across iterations, with increasing the iteration depth from 1 to 4 producing barely perceptible changes in the representation. At specific positions, however, the representation reorganises dramatically across the full layer stack, appearing as dark vertical streaks cascading through the depth of the model.

\paragraph{Methodology.} Isolating the state stream's contribution to these dynamics is not straightforward. The state stream and the attention mechanism are deeply entangled: at each layer, the blend alters the hidden state before the feedforward network, and the resulting output becomes the input to the next layer's attention. Attention at layer $l{+}1$ processes a representation that was shaped by the blend at layer $l$, and the post-attention hidden state that the blend receives at layer $l$ was itself produced by attention over the full token history. This coupling operates vertically through the layer cascade and horizontally through autoregressive generation, where each position's latent state and token output jointly shape the computation at all subsequent positions. There is no way to attribute any particular aspect of the model's output to the state stream versus the attention mechanism; the two are inseparable within normal generation. Comparing to the unmodified backbone would introduce a different confound: the \sst{} and the baseline have different LoRA weights and different fine-tuning, so any difference could be attributed to the adapted parameters rather than the state stream mechanism itself.

Comparing the same model across iteration depths provides a window into the state stream's effect on the latent dynamics. The model, weights, and prompt are identical across runs at different depths. The only controlled variable is the number of blend-feedforward-update cycles per position: the state stream has processed each prior position through more iterations, producing different latent states that propagate forward through the blend.

Comparing across iteration depths introduces its own constraint. Once any run at any iteration depth produces a different argmax token, all subsequent positions attend to different token histories through the KV cache. After divergence, changes in the output distribution reflect both the state stream and attention over different tokens, and the two contributions cannot be separated. The analysis below therefore uses different scope choices depending on the measurement. The overlap analysis of this subsection characterises the representational structure across all positions. Aggregate basin-shift measurements, including the layer profile, restrict to \emph{pre-divergence positions} (those before the first argmax disagreement between any pair of iteration-depth runs), where the token history is identical across depths and any output change is causally attributable to the state stream alone. This matches the population used by the causal analysis of Section~\ref{sec:logit_dynamics}. For measurements requiring causal attribution beyond the divergence point, the analysis compares between iterations within individual runs, where the token history is shared across iterations by construction (Appendix~\ref{app:logit_methodology}).

Even at pre-divergence positions, cross-run comparisons do not distinguish between the two axes along which the state stream operates. The vertical axis is the local blend-feedforward-update cycle at a given position; the horizontal axis is the propagation of iterated states from prior positions through the blend. At any position beyond position~0, these two axes contribute simultaneously: the iter${} = 2$ run's prior positions have each undergone an additional iteration cycle, and the resulting states have propagated forward through the blend, so the output distribution already differs from the iter${} = 1$ run before any local iteration difference at the current position. To attribute changes specifically to local iteration when characterising the gap distributions~(3) and output-distribution reorganisation~(5) of the two overlap regimes (Section~\ref{sec:logit_dynamics}), the analysis uses within-run comparisons where horizontal propagation is shared across all iterations. Position~0 serves as a natural control: with no prior positions, horizontal propagation is zero.

We measure the top-1024 overlap between hidden state representations at iteration~1 and iteration~$i$ at each layer and position (formal definition in Appendix~\ref{app:topk_overlap}). The metric captures which dimensions are most active in each representation, providing a content-sensitive measure of representational similarity. We conduct this analysis on all 198 GPQA-Diamond questions (Section~\ref{sec:evaluation}), an out-of-distribution benchmark where larger headroom for improvement and wider effect ranges across iteration depths make the dynamics more visible. The punctuated equilibrium structure is present across the full generation (Figure~\ref{fig:overlap_heatmap} shows a representative 512-position trace); we record hidden states at the first 10 generated positions per question across all four iteration depths and all 62 layers, as most questions diverge in argmax within this window.

\paragraph{Bimodal overlap distribution.} The overlap quantifies the visual pattern in Figure~\ref{fig:overlap_heatmap}: stable positions show overlap above 0.99, while low-overlap positions drop sharply, to as low as 0.60 for this particular example. Across all 198 questions, the representational change concentrates at positions where the hidden state undergoes a sharp reorganisation, while surrounding positions remain stable. A Gaussian mixture model fitted to the overlap distribution ($N = 122{,}760$ position-layer pairs; methodology in Appendix~\ref{app:gmm}) identifies two components: a \emph{stable} component (86.2\%, mean overlap $0.990 \pm 0.004$) and a \emph{low-overlap} component (13.8\%, mean overlap $0.869 \pm 0.092$), separated at a crossover threshold of 0.976. The order-of-magnitude difference in standard deviation reflects the two distinct regimes: stable positions cluster tightly, while low-overlap positions span a wide range of divergence magnitudes, with overlap as low as 0.30.

\paragraph{Content-dependent distribution.} The distribution of low-overlap positions within this 10-position window is content-dependent. Across the 198 questions, the number of low-overlap positions per question ranges from 1 to 7 (median 3), and the per-position rates vary from 4\% (position~3) to 42\% (position~1). If low-overlap were a property of position alone, every question would have the same positions flagged and the count would be fixed; it is not. If low-overlap were independent of position, whether arising from a uniform random process or a fixed per-position probability, the per-position rates would be similar; they differ by an order of magnitude. The distribution is neither purely positional nor position-independent: which positions are low-overlap depends on the interaction of position and question content. The gaps between consecutive low-overlap positions range from 1 to 9 with a coefficient of variation of 0.84, ruling out periodicity. This heterogeneity across questions means that basin shifts cannot be reduced to identifiable surface-level cues or token categories; they are triggered by the interaction of content and position within the latent representation itself.

\paragraph{Floating-point precision rules out artefact.} A possible concern is whether the between-iteration deltas could arise from deterministic floating point error propagation rather than from the FFN computing on different inputs. The blend modifies each dimension of the FFN input by $\alpha_{l,d} \times \Delta_d$ (Eq.~\ref{eq:blend}), where $\Delta_d$ is the state-hidden difference. For this perturbation to be indistinguishable from bf16 rounding error ($|h_d| \times \epsilon$, where $\epsilon = 2^{-7}$), the blend coefficient would need to fall below $\epsilon$. The learned coefficients range from $0.024$ to $0.035$ across all $333{,}312$ dimensions, placing every value above $3\epsilon$. At basin shift positions (Section~\ref{sec:punctuated_equilibrium}), which drive the largest representational changes and the most consequential downstream effects on the output distribution, the realised per-dimension deltas grow through the vertical cascade as each layer's FFN amplifies the perturbation from the layer below: the median delta-to-precision ratio rises from $2.9\times$ at layer~$0$ to $28.9\times$ at layer~$61$, with $91.3\%$ of per-dimension measurements exceeding the precision floor across all $62$ layers ($p < 10^{-300}$; Appendix~\ref{app:precision_proof}). The representational changes observed throughout this section are therefore the product of $62$ layers of feedforward computation on genuinely different inputs, compounded through the vertical cascade.

\begin{figure}[H]
  \centering
  \includegraphics[width=\linewidth]{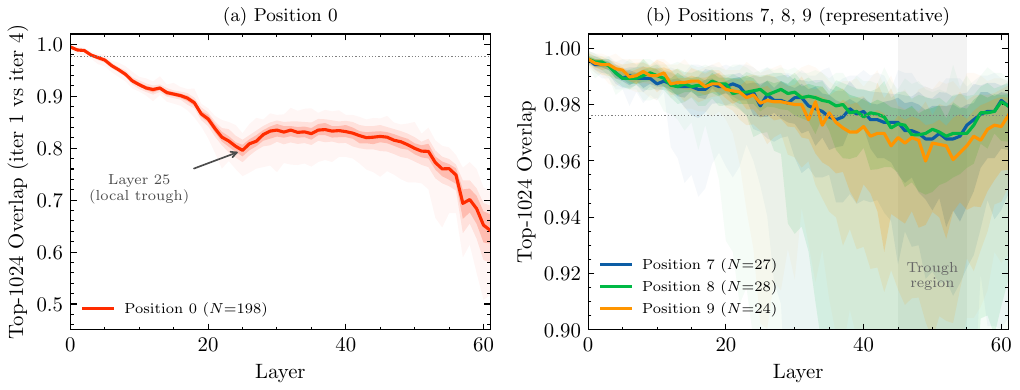}
  \caption{\textbf{Layer profile of top-1024 overlap (iter=1 vs iter=4) at low-overlap positions.} \textbf{(a)} Position 0 ($N=198$): universal basin shift, with the local trough at the layer~25 feedforward activity band. \textbf{(b)} Positions 7, 8, 9 (representative): divergence onset at $\sim$layer~25, trough in the middle-to-late layers ($\sim$layer~50), rising overlap at deep layers. Solid line = median; shaded bands = IQR, p10--p90, and p5--p95.}
  \label{fig:layer_profile}
\end{figure}

\paragraph{Layer profile structure.} The \emph{median} layer profile across 356 pre-divergent low-overlap positions is non-monotonic, showing structured variation across the model depth rather than a uniform accumulation of divergence (Figure~\ref{fig:layer_profile}). The feedforward networks are not uniformly active: gate activity peaks in layers 22--30 (where 30--40\% of neurons are active), and gate flip rates between iterations are highest in layers 15--35. In the median trend, overlap begins near 1.0 at the early layers (median $0.995$, IQR $0.994$--$0.997$ at layer~0) and descends through layers 5--25, reaching a local minimum at layer~25 (median $0.816$, IQR $0.794$--$0.984$) within this peak activity band. The median then increases through layers 25--30 (median $0.853$, IQR $0.834$--$0.982$ at layer~30) before a second descent in the deep layers (50--61), reaching a global minimum at layer~61 (median $0.689$, IQR $0.637$--$0.978$). However, the per-position spread is substantial: individual positions range from below 0.60 to above 0.95 at layer~61, and the median increase from layer~25 to~30 ($+0.036$) is small relative to the IQR at those layers. The non-monotonic shape characterises the typical behaviour across low-overlap positions, not a universal per-position trajectory. At stable positions, overlap remains flat (median above 0.988 at all layers). Extended layer profile data, separately for each of the first 10 generated positions, are given in Appendix Figure~\ref{fig:layer_profile_per_position}.

\paragraph{Universal basin shift at position 0.} The low-overlap effect at position~0 is universal. Position~0 is the first generated token. At iter=1, position~0 performs a single forward pass through the model, producing the first latent state to be stored in the \lsc{}. This single-pass computation is no different from a standard transformer forward pass. At iter=2 and beyond, position~0 undergoes additional blend-feedforward-update cycles, each refining the representation using the state produced by the previous iteration. Position~0 is therefore the cleanest point of comparison between iteration depths: it is the first generated token, and the only difference between runs is the number of state stream iterations. All 198 questions show low top-1024 overlap at position~0 between iter=1 and iter=2 (mean $0.635$, std $= 0.042$), regardless of content or difficulty: the basin shift is present at the minimum additional iteration depth, not just cumulatively. (Elsewhere we compare iter=1 to iter=4 for the full accumulated effect.) Even at the very first generated token, additional iterations substantially reshape the hidden state representation.

The punctuated equilibrium in the hidden states raises a direct question: do these sharp representational changes actually affect what the model outputs? If the low-overlap positions change the hidden states but leave the output distribution unchanged, they would be computationally inert. The following analysis restricts to pre-divergence positions as defined above, where output changes are causally attributable to the state stream alone.

\subsection{Logit dynamics}
\label{sec:logit_dynamics}

The following analyses characterise how these latent dynamics shape the output distribution, establish the causal mechanics underlying them, and rule out null hypotheses that would attribute the patterns to artefact. All measurements operate on the same first-10-position-per-question window as Section~\ref{sec:punctuated_equilibrium}'s overlap analysis.

\paragraph{(1) Argmax changes concentrate at low-overlap positions.}
At pre-divergence positions, low-overlap positions have $3.8\times$ higher odds of changing the argmax than stable positions ($12.0\%$ vs $3.5\%$; OR$=3.82$, $p<0.001$; Table~\ref{tab:argmax_concentration}). The hidden state reorganisations of Section~\ref{sec:punctuated_equilibrium} produce a measurably different effect on the output distribution than stable positions do.

\paragraph{(2) The argmax changes are not decision boundary exploitation.}
Only $11.6\%$ of argmax changes occur at exact ties (gap $= 0$; the smallest non-zero gap observed in the data is $0.121$ nats, so the boundary between exact ties and non-ties is unambiguous); the remaining $88.4\%$ override a token with strictly higher probability. The logprob shift at argmax-changing positions (median $0.87$ nats; $e^{0.87} = 2.39\times$ probability multiplier) represents a substantial restructuring of the output distribution, compared to the marginal variation at non-changing positions (median $0.05$ nats; $e^{0.05} = 1.05\times$). The latent computation produces large, content-specific shifts that override real preferences, not small perturbations catching coincidental ties (Table~\ref{tab:decision_boundary}).

\paragraph{(3) Low-overlap positions override larger preferences.}
At stable positions, argmax changes occur almost exclusively at exact ties or the minimum representable gap ($54.3\%$ exact ties, range $0.00$--$0.125$ nats). At low-overlap positions, however, the distribution shifts toward larger gaps ($5.0\%$ exact ties, range $0.00$--$6.375$ nats; Mann--Whitney U test, $p < 0.001$), and low-overlap positions are simultaneously far less likely to be exact ties (Fisher's exact OR$=0.044$, $p<0.001$). The positions that concentrate argmax changes are the positions where ties are least likely to explain them (Table~\ref{tab:gap_distributions}).

\paragraph{(4) The hidden state change is causally prior.}
Across all 198 questions, low top-1024 overlap either precedes or coincides with the first argmax divergence between iteration depths, and never occurs in the reverse order. Of the 57 questions where divergence falls outside the 10-position observation window, full trace comparison confirms all 57 produce different text at different iteration depths; the divergence occurs later, not never. The latent state change is the cause, and the output change is the consequence (Table~\ref{tab:causal_ordering}).

\paragraph{(5) Two distinct regimes of output-distribution reorganisation.}
The LM head projects the model's current latent belief state into an output distribution over the next token, making the changes induced by recurrent computation directly observable. The resulting changes separate into two regimes. At low-overlap positions where the argmax differs between iterations within the same run ($N=439$), the output distribution reorganises comprehensively; roughly half the top-100 tokens are replaced (median 54), the original winner is always suppressed, and the new winner can come from as deep as rank 50 in the original distribution. At stable positions where the argmax differs ($N=81$), the output distribution is functionally preserved; the local iteration replaces median 1 token (max 3), suppresses the original winner by median $0.0625$ nats, and the new winner is always rank 2.

A cross-run comparison using the same iter${=}1$ vs iter${=}2,3,4$ structure measures the total state stream effect across both axes. At stable positions, the cross-run output distribution differs by median 20 tokens (IQR 16--26) at all pre-divergence positions, yet the within-run local iteration replaces median 1 (IQR 1--2). This difference is present whether or not the argmax changes (cross-run mean 22.0 at all positions, 23.0 at argmax changes). Because the within-run comparison controls for horizontal propagation and shows stable positions replace only median 1 token locally, the median 20-token difference in the cross-run measurement is attributable to low-overlap positions at upstream positions propagating their reorganisation through the state stream. The stable positions remain in the output-distribution regime that the upstream reorganisation pushed them into, with only minor local perturbation, until the next low-overlap position. The within-run sample includes post-divergence positions excluded from the cross-run, yet the distributions are indistinguishable, confirming that the pre-divergence measurement window is representative of the behaviour across the full generation, consistent with the punctuated equilibrium observed across 512 positions in Section~\ref{sec:punctuated_equilibrium}.

The two regimes of output-distribution reorganisation are not the same process operating at different magnitudes. At low-overlap positions, the model transitions to an entirely different output-distribution regime. The combination of a sharp representational shift at specific positions followed by stability at subsequent positions constitutes a semantic basin shift: the model moves to a new attractor region and remains there as the state stream carries the representation forward. It does not denote an independently decoded semantic category. This is what produces the punctuated equilibrium observed in the hidden states (Section~\ref{sec:punctuated_equilibrium}): the basin structure itself, with shifts at specific content-dependent positions and stability between them. This punctuated equilibrium is an emergent property of the state stream architecture, not an explicitly trained behaviour. The same mechanism operates at iter${=}1$ with identical horizontal propagation; iteration depth is the instrument that makes the basin dynamics measurable, not the mechanism that creates them. Full statistical detail for both regimes is given in Tables~\ref{tab:reorg_lo_argmax}--\ref{tab:reorg_st_all}.

\paragraph{(6) Later iterations do real computational work.}
Later iterations continue to reshape the output distribution in structured, regime-dependent ways. Within the iter$=4$ run, basin shifts take three forms: sustained across all three consecutive iteration pairings, present only at $1\!\rightarrow\!2$, and genuinely new basin shifts that emerge at later pairings from positions that were stable at $1\!\rightarrow\!2$. Within the iter${=}4$ run, later iterations produce 37 additional argmax changes (19 at $2\!\rightarrow\!3$, 18 at $3\!\rightarrow\!4$), vs 88 at $1\!\rightarrow\!2$; token history is shared within the run by construction and the state stream is therefore the sole source of variation.

The character of this continued computation differs between regimes. At basin shift positions, later iterations amplify into the new basin: logit trajectories continue in the same direction, and positions that changed argmax at $1\!\rightarrow\!2$ reinforce the new basin, but can still change again at later transitions. At stable positions, the dynamics are heterogeneous, with logits splitting across reversal, retention, and continuation in roughly equal proportions (ratio $0.878$ vs $1.081$ at basin shift; $p < 10^{-56}$). The heterogeneous per-logit dynamics can tip the relative ordering among the immediate frontrunners (rank 1--3, as established above). $61.4\%$ of tracked logits reverse direction between transitions, but the cumulative ratio of $0.95$ rules out a period-2 orbit.

These dynamics propagate forward through the state stream's horizontal evolution: each subsequent position's blend--FFN--update cycle operates on the propagated latent state. Across all questions, $18.7\%$ of those that changed argmax at $1\!\rightarrow\!2$ show further argmax changes at later transitions, cascading through the KV cache into completely different generated output sequences (Tables~\ref{tab:basin_shift_activity}--\ref{tab:later_iter_computation}).

\paragraph{(7) Position 0 encodes whether iteration-depth divergence happens here or later.}
All 198 questions show a semantic basin shift at position 0 and from this same mechanism, three distinct outcomes across pairwise iteration-depth comparisons: 75 questions ($37.9\%$) diverge at position 0 itself, 66 ($33.3\%$) diverge at positions 1--9, and 57 ($28.8\%$) diverge beyond the 10-position observation window. Full trace comparison confirms all 57 late-divergence questions produce different text at different iteration depths, meaning that the divergence occurs later.

The same basin shift at the same position produces three distinct patterns of downstream divergence across the four iteration depths. As established above, argmax changes at basin shift positions are not determined by the top-1/top-2 gap alone: the new winner can come from as deep as rank 50 in the original output distribution, and the entire top-100 restructures with a median of 54 logits replaced (IQR 47--60). Whether the basin shift at position 0 crosses the argmax boundary immediately, propagates through subsequent positions before crossing, or produces a different generation sequence without crossing within the observation window depends on the full structure of the latent state at that position, which propagates forward through the state stream and shapes the computation at every subsequent position. The latent state at position 0 therefore potentially carries information about the downstream generation trajectory in its full $5{,}376$-dimensional representational structure, propagated forward through the state stream's horizontal evolution (Table~\ref{tab:pos0_divergence}).

\paragraph{(8) Convergent reasoning across computational paths.}
Interestingly, we observe that a substantial number of questions arrive at the same correct answer at every iteration depth. Is this convergence a coincidence on a 4-option multiple choice benchmark? To test this, we restrict the analysis to questions where all four flat iteration depths produced pairwise distinct text traces, so the observed convergence reflects different semantic paths rather than deterministic repetition where basin shifts failed to cross the argmax threshold. The chance null is $(1/4)^4$: four independent random answers on a 4-option MCQ.\footnote{A null based on the product of empirical per-depth accuracies would test mutual independence of correctness events across depths, not chance coincidence; however, the events are not empirically independent, as the observed all-four-correct rate is $60/198 = 0.303$ versus $0.55 \times 0.47 \times 0.47 \times 0.43 = 0.053$ under independence, and a $\chi^2$ test against this independence prediction rejects at $\chi^2 = 248.55$, $p = 5.38 \times 10^{-56}$.} At each iteration depth, the nonlinear recurrence transforms the latent state before the LM head produces the output distribution. Outputs at different depths therefore arise from different latent computations, rather than from stochastic sampling or reranking of a fixed distribution. The rate at which all four depths arrive at the correct answer is compared against this chance baseline in Table~\ref{tab:convergent_reasoning}. Of the 198 questions, 53 ($26.77\%$) produced completely distinct text traces across all four flat depths while universally arriving at the correct answer. This observed rate of convergent reasoning exceeds the $0.39\%$ chance baseline by $68.53\times$ ($p = 7.62 \times 10^{-80}$). The latent computation of the state stream is therefore functional rather than mechanical, producing exploration along distinct deliberative trajectories that resolve to correct answers at rates that cannot arise by chance (Table~\ref{tab:convergent_reasoning}).

The latent computation isolated in this section is structured, content-dependent, causally prior to the output, and convergent across different computational paths. This is incompatible with the null hypothesis that the variation between iteration depths is equivalent to stochastic sampling from a fixed distribution. The state stream architecture enables structured computation in continuous horizontal latent space across an entire generation sequence, with each iteration depth producing a different computation through this space.

\FloatBarrier
\section{Quantitative evaluation}
\label{sec:evaluation}

\begin{table}[H]
\centering
\caption{\textbf{Headline architectural comparison across benchmarks.} Each architecture is evaluated at its own compute budget: one forward pass per token for the matched fine-tuned baseline (Section~\ref{sec:baseline_ft}), up to $i_{\max} = 4$ iterations per token for the \sst{} at its latent compute capacity via staged compute (Section~\ref{sec:capacity}). Error correction is the fraction of the baseline's errors recovered by the \sst{}: $(\text{SST} - \text{baseline}) / (1 - \text{baseline})$.}
\label{tab:main_results}
\small
\begin{tabular}{lrcccc}
\toprule
Benchmark & $N$ & Baseline & \shortstack{\sst{}\\(staged compute, $i_{\max}{=}4$)} & $\Delta$ & \shortstack{Error\\correction} \\
\midrule
GSM8K & 1{,}319 & 94.77\% & \textbf{97.19\%} & $+2.43$pp & $+46.38\%$ \\
MATH-500 & 500 & 83.40\% & \textbf{89.80\%} & $+6.40$pp & $+38.55\%$ \\
GPQA-Diamond & 198 & 45.96\% & \textbf{61.11\%} & $+15.15$pp & $+28.04\%$ \\
HumanEval & 164 & 87.2\% & \textbf{89.0\%} & $+1.8$pp & $+14.3\%$ \\
\bottomrule
\end{tabular}
\end{table}

The evaluation compares the \sst{} against its matched fine-tuned baseline (Section~\ref{sec:baseline_ft}) under greedy argmax decoding at iteration depths 1 through 4. All generation is fully deterministic; the evaluation methodology and rationale are given in Appendix~\ref{app:eval_methodology}.

Table~\ref{tab:main_results} summarises the headline comparison across all four benchmarks. The remainder of this section develops the methodology behind these numbers (Section~\ref{sec:capacity}), compares the \sst{}'s accuracy against published results for other models (Section~\ref{sec:external}), characterises overthinking regression under uniform iteration depth (Section~\ref{sec:overthinking}), and reports output stability (Section~\ref{sec:output_stability}).

\subsection{Latent compute capacity via staged compute}
\label{sec:capacity}

The \sst{}'s iteration mechanism requires a new evaluation methodology. For a standard autoregressive model under greedy deterministic decoding, a single run directly measures the architecture's full deployable capacity on a benchmark, because iter${} = 1$ is the only depth the architecture offers within this regime. The \sst{} extends the architecture with an additional latent compute axis, and a single-depth evaluation captures only a slice of that axis. Measuring across iteration depths therefore characterises the architectural feature rather than introducing a confound. Section~\ref{sec:logit_dynamics} established that different iteration depths actively reorganise the output distribution, suppressing the original argmax, promoting tokens from deeper in the distribution, and replacing much of the top-100 candidate set at basin shift positions. A single-depth evaluation measures the model's capability at that depth alone, not the capacity of the architecture across the compute axis it provides. This is consistent with the broader finding that performance saturation in recurrent-depth architectures is task-dependent, with different tasks and problems saturating at different recurrence depths~\cite{geiping2025scaling}; evaluating such architectures at a single fixed depth systematically understates their capacity.

We therefore evaluate the \sst{} with a staged compute methodology at $i_{\max} = 4$, measuring the architecture's capability across the different computations enabled by its iteration depths. The architecture is evaluated sequentially in an escalatory pattern from iter${} = 1$ through iter${} = 4$, with each stage testing whether deeper computation reveals capability the shallower depth did not. Many questions that the architecture solves at one depth are also solved at deeper depths through convergent reasoning across different computational paths (Section~\ref{sec:logit_dynamics}); staged compute resolves these at the shallowest sufficient depth. The capacity at each stage is the fraction of questions the architecture solves through that depth.

The experiment isolates latent-space computation in a backbone without an extended-reasoning policy, without training an explicit CoT policy, and without eliciting extended reasoning at evaluation. Ordinary reasoning text remains permitted.

\begin{table}[H]
\centering
\caption{\textbf{Staged compute capacity at each depth.} Each question is evaluated sequentially from iter${} = 1$ through iter${} = 4$. Each column is the fraction of questions the architecture solves through that stage. These are staged capacities, not flat per-question iteration depths; behaviour under flat iteration is reported separately (Section~\ref{sec:overthinking}).}
\label{tab:staged_breakdown}
\small
\begin{tabular}{lccccc}
\toprule
Benchmark & Stage $i{=}1$ & Stage $i{=}2$ & Stage $i{=}3$ & Stage $i{=}4$ & Capacity \\
\midrule
GSM8K ($N = 1{,}319$) & 95.75\% & 97.12\% & 97.19\% & 97.19\% & \textbf{97.19\%} \\
MATH-500 ($N = 500$) & 83.80\% & 87.60\% & 89.00\% & 89.80\% & \textbf{89.80\%} \\
GPQA-Diamond ($N = 198$) & 51.01\% & 56.57\% & 57.58\% & 61.11\% & \textbf{61.11\%} \\
HumanEval ($N = 164$) & 86.0\% & 89.0\% & 89.0\% & 89.0\% & \textbf{89.0\%} \\
\bottomrule
\end{tabular}
\end{table}

Staged compute might be confused with pass@$k$ because both involve evaluating the architecture across multiple runs, but the structures differ. Pass@$k$ draws $k$ samples from a single output distribution at a fixed compute budget, relying on sampling variance for diversity while the underlying computation remains fixed. Staged compute varies the computation itself by increasing iteration depth, with each stage applying additional nonlinear recurrent computation before the LM head produces the output distribution (Section~\ref{sec:logit_dynamics}). The iter${} = 4$ output is not another sample from the iter${} = 1$ distribution but the result of deeper latent computation. The two methodologies therefore vary along different axes and answer different questions.

This separation has published precedent in recurrent-depth evaluation. Fu et al. compare correctness-routed and fixed latent depths within one model, while Fan et al. compare oracle-routed loop counts with fixed-depth baselines~\cite{fu2026think,fan2025looped}. Neither adds stochastic best-of-$n$ sampling to the fixed-depth control to equalise total inference compute.

Behaviour under uniform iteration depth is characterised separately (Section~\ref{sec:overthinking}).

This evaluation selects iteration depth at the question level. Per-position depth selection is a separate design, since it would require deciding whether to halt at each generated token rather than once per question; we discuss this limitation in Section~\ref{sec:limitations}.

\subsection{External comparison}
\label{sec:external}

The comparison with the matched fine-tuned baseline (Table~\ref{tab:main_results}) isolates the contribution of the state stream mechanism. To place the resulting accuracy in the context of published results, we compare the \sst{}'s staged compute capacity with reported numbers for other models on GSM8K (Table~\ref{tab:external_gsm8k}), MATH-500, and GPQA-Diamond (Table~\ref{tab:external_gpqa}). All \sst{} numbers are produced under zero-shot prompting with greedy argmax decoding and no chain-of-thought exemplars. External comparisons are inherently uncontrolled, with reported protocols varying in shot count, chain-of-thought use, sampling temperature, and sample-size choices, often without full disclosure. These tables place the \sst{}'s staged-compute capacity result in the context of published benchmark results. The matched baseline in Section~\ref{sec:baseline_ft} is the controlled architectural comparison.

\begin{table}[H]
\centering
\small
\caption{\textbf{External comparison on GSM8K.} The \sst{} at $i_{\max} = 4$ exceeds all four reported comparisons, including Llama 3.1 405B at 15$\times$ its parameter count, under stricter prompting conditions.}
\label{tab:external_gsm8k}
\begin{tabular}{lccl}
\toprule
Model & Parameters & GSM8K & Conditions \\
\midrule
Llama 3.1 70B Instruct & 70B & 95.1\% & 8-shot CoT~\cite{llama31_70b} \\
Qwen 2.5 72B Instruct & 72B & 95.8\% & Self-reported~\cite{qwen25} \\
Gemma 3 27B-IT & 27B & 95.9\% & 8-shot CoT~\cite{gemma3} \\
Llama 3.1 405B Instruct & 405B & 96.8\% & 8-shot CoT~\cite{llama31_405b} \\
\textbf{\sst{} (staged compute $i_{\max}{=}4$)} & \textbf{27B} & \textbf{97.19\%} & \textbf{0-shot, greedy} \\
\bottomrule
\end{tabular}
\end{table}

\begin{table}[H]
\centering
\small
\caption{\textbf{External comparison on GPQA-Diamond.} Reported accuracies and prompting conditions. The \sst{} at $i_{\max} = 4$ answers more questions correctly than DeepSeek V3 at 671B, Gemini 2.0 Flash, and several 70B+ open-weight models, while using zero-shot greedy decoding at 27B parameters; DeepSeek V3 has roughly $25\times$ the parameter count.}
\label{tab:external_gpqa}
\begin{tabular}{lccl}
\toprule
Model & Parameters & GPQA-Diamond & Conditions \\
\midrule
Gemma 3 27B-IT & 27B & 42.4\% & 5-shot CoT~\cite{gemma3} \\
Qwen 2.5 72B Instruct & 72B & 49.0\% & Self-reported~\cite{qwen25} \\
Llama 3.3 70B Instruct & 70B & 50.5\% & 0-shot CoT~\cite{llama33} \\
GPT-4o & Proprietary & 56.1\% & pass@1~\cite{gpt4o} \\
DeepSeek V3 & 671B & 59.1\% & pass@1~\cite{deepseekv3} \\
Gemini 2.0 Flash & Proprietary & 60.1\% & pass@1~\cite{gemini20flash} \\
\textbf{\sst{} (staged compute $i_{\max}{=}4$)} & \textbf{27B} & \textbf{61.11\%} & \textbf{0-shot, greedy} \\
Gemini 2.0 Pro & Proprietary & 64.7\% & pass@1~\cite{gemini20pro} \\
PhD domain experts & --- & 69.7\% & \cite{rein2023gpqa} \\
Gemini 2.5 Pro & Proprietary & 84.0\% & pass@1~\cite{gemini25pro} \\
\bottomrule
\end{tabular}
\end{table}

On MATH-500~\cite{hendrycks2021math,lightman2023verify}, the matched baseline regresses from the unmodified Gemma~3 27B-IT's reported $89.0\%$~\cite{gemma3} to $83.40\%$, likely because the CodeACT fine-tuning specialises the model toward Python-based problem solving at the cost of competition-level mathematical capability. The \sst{} at iter${} = 1$ shows a similar regression ($83.80\%$), confirming this is a fine-tuning effect rather than a checkpoint artefact. With staged compute, the \sst{} recovers to $89.80\%$ ($+6.40$pp over the matched baseline), achieving slightly higher accuracy than the unmodified backbone.

On HumanEval~\cite{chen2021humaneval}, the \sst{}'s $89.0\%$ exceeds the reported Gemma 3 27B-IT result of $87.8\%$ under $0$-shot pass@1~\cite{gemma3}. HumanEval is near-saturated for models at this scale and has a small test set ($N = 164$); we include it here for completeness rather than as a central comparison.

Together, these comparisons place the \sst{}'s accuracy in the published model landscape, where reported results necessarily differ in prompting, decoding, evaluation harness, and reporting conventions. The \sst{} is evaluated under the deterministic zero-shot argmax protocol justified in Appendix~\ref{app:eval_methodology}; within that landscape, staged compute yields accuracy competitive with much larger and proprietary models at 27B parameters.

\subsection{Overthinking regression}
\label{sec:overthinking}

The \sst{} checkpoint evaluated in this paper has no native halting mechanism: iteration occurs only at inference, not during training (Section~\ref{sec:training}), so no halting criterion can be learned end-to-end. Iteration depth must therefore be chosen externally at inference. Staged compute (Section~\ref{sec:capacity}) measures what the architecture can achieve given per-question depth allocation, but it cannot measure overthinking: once a question passes, staged compute stops and therefore cannot reveal whether further iteration would make that answer regress. Measuring overthinking requires complete flat evaluations in which every question is evaluated independently at every uniform depth. We perform these evaluations on GPQA-Diamond at iter${} = 1, 2, 3, 4$.

The flat iter${} = 1$ evaluation obtains $109/198$ ($55.05\%$), whereas the iter${} = 1$ stage of the headline staged-compute evaluation obtains $101/198$ ($51.01\%$). The flat-depth sweep used a higher-throughput batch configuration because running every question at every depth under the original configuration was infeasible within the available single-GPU evaluation budget. With the fused attention backend required for tractable evaluation, different batch configurations can follow different floating-point reduction paths and thereby alter greedy autoregressive trajectories. In a paired comparison, 14 questions pass only in the staged configuration and 22 only in the flat configuration; the aggregate difference is not significant (exact McNemar paired test, $p = 0.243$), providing no evidence that either configuration systematically increases accuracy. We use the flat iter${} = 1$ result here because the overthinking analysis requires a flat reference at every depth. The four flat runs are diagnostic evaluations, not candidate headline runs from which we select the highest aggregate score. We retain the lower staged iter${} = 1$ result in all headline comparisons because it belongs to the single fixed protocol used to measure capacity under perfect depth routing; substituting the higher flat result would mix evaluation protocols after observing their outcomes and inflate the reported headline performance.

\begin{table}[H]
\centering
\small
\caption{\textbf{Overthinking regression on GPQA-Diamond ($N = 198$).} In the flat-depth experiment, applying additional iterations to every question produces accuracy below flat iter${} = 1$ at every deeper depth tested. The staged-compute result is shown as a separate headline reference and is not used to calculate the flat-depth deltas.}
\label{tab:overthinking}
\begin{tabular}{lcc}
\toprule
Condition & Accuracy & $\Delta$ vs flat iter${} = 1$ \\
\midrule
\sst{} staged compute ($i_{\max}{=}4$; headline reference) & \textbf{61.11\%} & --- \\
\midrule
\sst{} flat iter${} = 1$ & 55.05\% & --- \\
\sst{} flat iter${} = 2$ & 46.97\% & $-8.08$pp \\
\sst{} flat iter${} = 3$ & 47.47\% & $-7.58$pp \\
\sst{} flat iter${} = 4$ & 42.93\% & $-12.12$pp \\
\midrule
Matched baseline (reference) & 45.96\% & \\
\bottomrule
\end{tabular}
\end{table}

\begin{figure}[htbp]
  \centering
  \includegraphics[width=\linewidth]{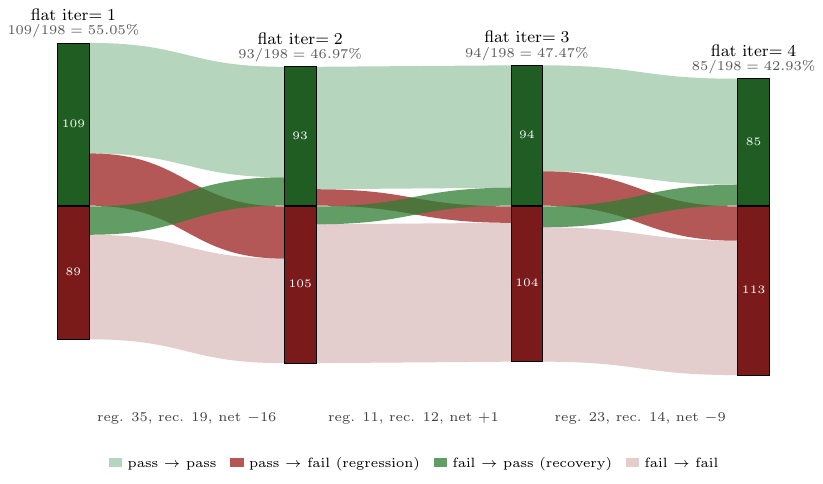}
  \caption{\textbf{Per-question pass/fail trajectories across flat iteration depths on GPQA-Diamond ($N = 198$).} Flow of the 198 questions between pass (green) and fail (red) columns in the independent flat iter${} = 1, 2, 3, 4$ evaluations. Each transition decomposes into four ribbons: pass${\to}$pass (stable correct), pass${\to}$fail (regression), fail${\to}$pass (recovery), and fail${\to}$fail (stable wrong). Accuracy above each column shows the aggregate; regression and recovery counts below each transition show the per-question flips. Every transition involves flips in both directions, and the pass column shrinks from 109 at flat iter${} = 1$ to 85 at flat iter${} = 4$ through the accumulation of more regressions than recoveries.}
  \label{fig:overthinking_sankey}
\end{figure}

Every deeper flat iteration depth produces accuracy below flat iter${} = 1$ (Figure~\ref{fig:overthinking_sankey}). At flat iter${} = 4$, 36 of the 109 questions that pass at flat iter${} = 1$ regress to failure ($33.0\%$), while only 12 of the 89 flat iter${} = 1$ failures are recovered (exact McNemar $p < 0.001$). Flat iter${} = 4$ answers fewer questions correctly ($42.93\%$) than the matched baseline ($45.96\%$): the flat iter${} = 1$ advantage over the baseline ($+9.09$pp) is entirely consumed by overthinking regressions, followed by a further $3.03$pp loss. Compared against staged compute capacity ($61.11\%$), flat iter${} = 4$ loses $18.18$pp (McNemar $\chi^2 = 27.0$, $p < 0.001$, with 42 questions correctly answered under staged compute that flat iter${} = 4$ misses versus 6 in the reverse direction).

This phenomenon has direct counterparts in the token-space test-time compute scaling literature. Zhou et al.~\cite{zhou2026morethinking} study overthinking under extended chain-of-thought on GPQA-Diamond and other benchmarks and report the same behavioural pattern: beyond a problem-dependent compute budget, additional reasoning causes previously correct answers to become incorrect, which they term \emph{negative flips}. They find that optimal thinking length varies across problem difficulty and conclude that uniform compute allocation is suboptimal. Our $36/109$ regression rate at flat iter${} = 4$ is the latent-space analogue of this effect, and the $18.18$pp staged-minus-flat gap is the same ``uniform allocation is suboptimal'' finding expressed along our architecture's compute depth axis rather than along token budget. Hägele et al.~\cite{hagele2026hotmess} observe that failures at longer reasoning become dominated by incoherence rather than systematic error. Hakim~\cite{hakim2026brevity} documents the same overreasoning pattern along a different axis of scale, where large models underperform smaller ones by an average of $28.4$pp on $7.7\%$ of benchmark problems and brevity constraints recover $26.3$pp of accuracy. Across token-space reasoning, model scale, and the \sst{}'s latent compute depth, the same pattern recurs: additional compute helps some questions and hurts others, and the model has no explicitly designed signal for deciding when to stop.

Other latent reasoning architectures provide converging evidence. Universal Transformers~\cite{dehghani2018universal} compared per-position adaptive halting against fixed-depth computation and found that halting improved accuracy on several structured algorithmic and linguistic inference tasks while marginally degrading results on machine translation. The improvement on structured tasks is direct evidence that uniform depth allocation is suboptimal for variable-depth reasoning architectures. Coconut~\cite{hao2024coconut} trains with a multi-stage curriculum that matches continuous-thought count to each problem type (up to $k = 6$ on ProsQA, where 6 is the maximum reasoning depth of the benchmark) and reports monotonic accuracy improvement up to the trained depth; their experiments do not push past this regime and therefore do not probe the overthinking behaviour that manifests in our flat iteration tests. Geiping et al.~\cite{geiping2025scaling} observe that their recurrent-depth model's latent state does not always converge monotonically, exhibiting orbits and drifts in PCA space, and that running at $r = 32$ underperforms their KL-divergence-based early-exit criterion on GSM8K ($44.8\%$ vs $46\%$), though they frame test-time compute as sigmoidal in the number of additions and characterise the effect as diminishing returns rather than performance regression. These architectures operate on the vertical axis alone (Geiping et al., Universal Transformers) or on horizontal state propagation within a bounded deliberation phase (Coconut), differing structurally from the \sst{}'s combination of vertical compute with horizontal state continuity during token generation. The consistency of the pattern across these differing latent reasoning architectures supports overthinking regression as a property of variable-depth compute scaling rather than of any single architectural choice.

The practical consequence is that realising the staged compute capacity in deployment requires a mechanism for selecting per-question iteration depth autonomously. Section~\ref{sec:halting_probe} develops such a mechanism.

\subsection{Output stability}
\label{sec:output_stability}

\sst{} V1 applied the state stream to frozen pretrained weights without training, requiring multiple iterations per token to produce coherent output~\cite{aviss2025sst}. The two-pass training procedure (Section~\ref{sec:two_pass}) resolves this by making the nonlinear recurrence trainable for the first time. The backbone adapts to the state stream, producing stable single-pass generation and freeing iterations from a stability requirement into a deliberation choice.

To verify this, we measure sentence-level repetition across three evaluation benchmarks. Each benchmark example is a multi-turn CodeACT interaction in which the model produces several text turns interleaved with tool outputs. For each model turn, we compute the fraction of sentences (minimum 20 characters) that appear more than once within that turn, then average across turns within the example. Measuring within individual turns rather than across the concatenated conversation avoids false positives from formulaic phrases that naturally recur across turns in structured agent interactions. The 20-character minimum filters short structural sentences; below this length, manual inspection confirmed that repeated sentences are predominantly formatting rather than degenerate. Both the \sst{} and the matched baseline (Section~\ref{sec:baseline_ft}) are evaluated on identical question sets under identical generation conditions: greedy decoding, with no repetition penalty. The cross-benchmark comparison uses staged compute for the \sst{} and iter${}=1$ for the baseline; on GPQA-Diamond, we additionally evaluate the same \sst{} checkpoint at flat iter${}=1$ and flat iter${}=4$ to separate the state stream from unnecessary iteration depth.

\begin{table}[H]
\centering
\small
\caption{\textbf{Sentence-level repetition by benchmark and inference configuration.} Repetition is measured within each model turn and averaged across turns. The GPQA-Diamond rows include flat iter${}=1$ and flat iter${}=4$ runs on the same \sst{} checkpoint, isolating single-pass state-stream behaviour from unnecessary iteration depth.}
\begin{tabular}{llccc}
\toprule
Benchmark & Configuration & $N$ & Sent. rep. & Loops \\
\midrule
GSM8K & \sst{} staged & 1{,}319 & 0.35\% & 2/1{,}319 \\
GSM8K & Baseline iter${}=1$ & 1{,}319 & 0.07\% & 0/1{,}319 \\
HumanEval & \sst{} staged & 164 & 1.35\% & 0/164 \\
HumanEval & Baseline iter${}=1$ & 164 & 1.72\% & 0/164 \\
\midrule
GPQA-Diamond & Baseline iter${}=1$ & 198 & 11.13\% & 8/198 \\
GPQA-Diamond & \sst{} flat iter${}=1$ & 198 & 11.21\% & 6/198 \\
GPQA-Diamond & \sst{} staged & 198 & 11.63\% & 13/198 \\
GPQA-Diamond & \sst{} flat iter${}=4$ & 198 & 17.03\% & 26/198 \\
\bottomrule
\end{tabular}
\label{tab:repetition}
\end{table}

Because the question is whether the state stream introduces a \emph{meaningful amount} of repetition, we report the magnitude of the difference, with a confirmatory significance test included for transparency. The matched staged-vs-baseline comparison shows no evidence of a meaningful repetition increase: rates are near zero on GSM8K and HumanEval, and on GPQA-Diamond the staged \sst{} rate remains close to the matched baseline. The direct single-iteration GPQA-Diamond comparison isolates the architecture itself: \sst{} flat iter${}=1$ has $11.21\%$ sentence repetition versus $11.13\%$ for the matched baseline (Mann--Whitney $p = 0.942$), with fewer detected loops (6 vs 8). Repetition increases only when the same \sst{} checkpoint is forced to continue to flat iter${}=4$ for every question ($17.03\%$, 26 loops), which also drops accuracy from $61.1\%$ under staged compute to $42.9\%$. This indicates that with the trained \sst{} V2, iterations function as a per-question deliberation budget rather than a blanket requirement for coherence, resolving the instability found in V1.

\section{Adaptive halting probe}
\label{sec:halting_probe}

Section~\ref{sec:overthinking} documented an $18.18$pp gap between flat iter${} = 4$ and the staged compute capacity on GPQA-Diamond, driven entirely by regressions on questions that did not need additional iteration. Staged compute (Section~\ref{sec:capacity}) already offers a viable deployment strategy by allowing external depth selection based on output quality, analogous to the selectable reasoning effort in chain-of-thought reasoning models~\cite{openai2024o1}. However, autonomous depth selection from the model's own latent state would be a desirable further step, removing the need for external evaluation altogether. Section~\ref{sec:punctuated_equilibrium} established that every generation begins with a universal basin shift into a content-dependent region of the latent space at position~$0$, and Section~\ref{sec:logit_dynamics} that this basin-shifted representation carries information about the trajectory the generation will take. This section asks a sharper question of the same representation: does the position-$0$ latent state at iteration depth $d$ predict whether the computational trajectory it initiates will still lead to the correct answer if every subsequent position of the generation is iterated to depth $d{+}1$?

The experiment is scoped to feasibility. A deployed halting head would be calibrated post-training on labels generated at scale from the training data itself, which is a separate engineering effort not within scope of this paper. Answering the question posed above requires more than training a probe to high accuracy. The section therefore presents two complementary analyses: a mechanistic analysis of the probe's weights (Section~\ref{sec:halting_probe_structure}), identifying the specific hidden-state dimensions the probe reads; and a held-out generalisation test (Section~\ref{sec:halting_probe_generalisation}), asking whether that feature transfers to questions the probe was never trained on. Memorisation of question-identity patterns is the null hypothesis both analyses rule out in the course of answering these questions.

\subsection{Halt signal as overthinking-regression detection}

Prior work on adaptive halting in iterative transformers learns the stopping criterion end-to-end with the task loss (Adaptive Computation Time; \citet{graves2016act}, applied in the Universal Transformer; \citet{dehghani2018universal}). End-to-end halting is not available to the \sst{} during SFT, as training with iteration under teacher forcing could incentivise the model to defer reasoning to later iterations, producing a model that requires multiple passes for baseline competence rather than one that benefits from them (Section~\ref{sec:two_pass}). The two-pass method therefore trains at iter${} = 1$ equivalent by design, and the halting mechanism must be a post-hoc probe trained on separate labels. The choice of training target becomes a design decision. The natural target is convergence monitoring: has the computation settled? Whether phrased as ``am I done?'', ``should I continue?'', ``must I halt?'', or ``have I converged?'', these are semantically different expressions of the same underlying measurement. All prior approaches to adaptive halting, including ACT, fall in this family.

\begin{figure}[htbp]
  \centering
  \begin{subfigure}[t]{\linewidth}
    \centering
    \includegraphics[width=0.85\linewidth]{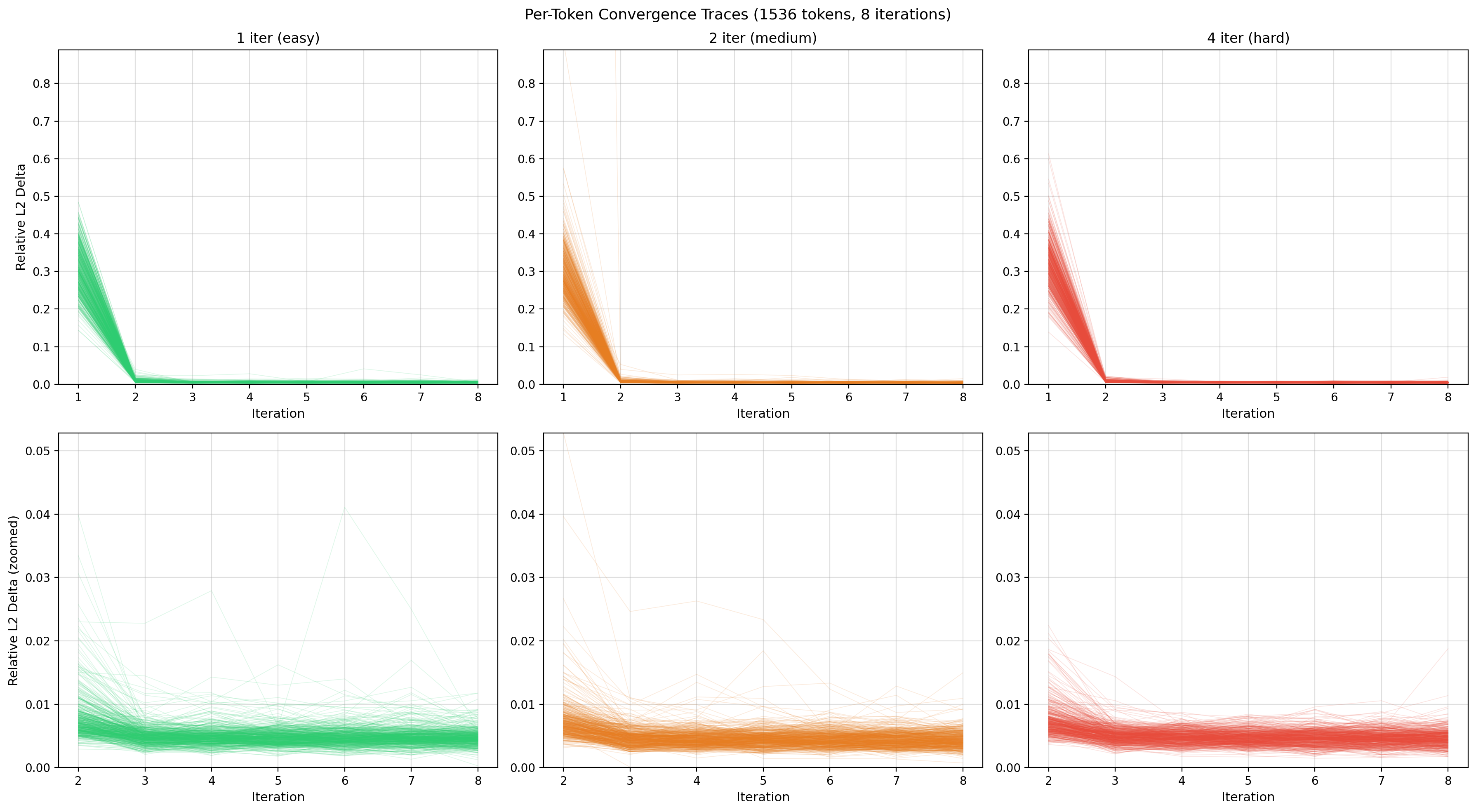}
    \caption{Per-token L2 delta between successive iterations for $1{,}536$ positions across $8$ iterations, grouped by correct iteration depth (iter${} = 1$ green, iter${} = 2$ orange, iter${} = 4$ red). Top: all three difficulty groups collapse to the same near-zero profile by iter${} = 2$, with L2 unable to distinguish questions needing one iteration from those needing four. Bottom (zoomed): outlier traces continue exploring at later iterations.}
  \end{subfigure}
  \vspace{0.5em}
  \begin{subfigure}[t]{\linewidth}
    \centering
    \includegraphics[width=0.85\linewidth]{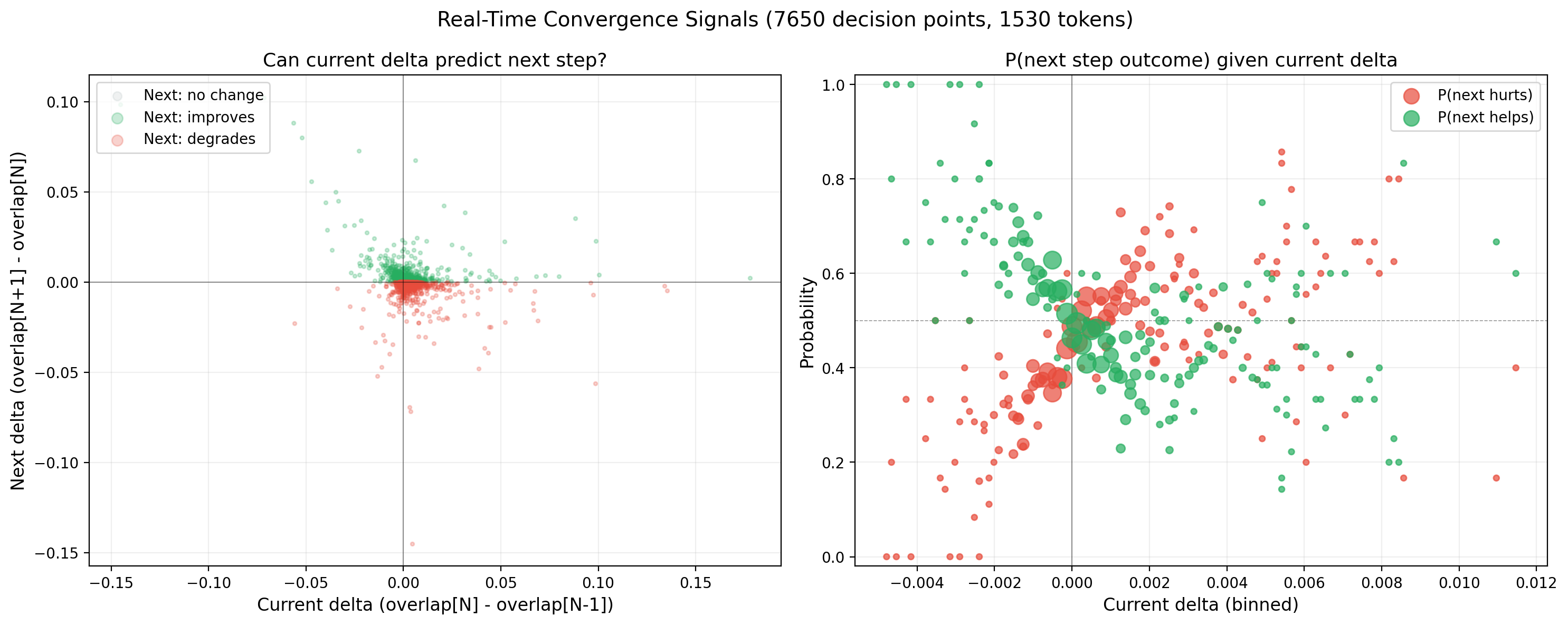}
    \caption{Predictive power of current L2 delta over the next iteration's outcome. Left: current delta (x-axis) vs next delta (y-axis), coloured by whether the next step improves (green), degrades (red), or has no change (grey). The vertical axis encodes the outcome; the horizontal axis is the signal available at decision time. Right: $P(\text{next outcome})$ given current delta, showing both outcomes are completely interleaved at every current-delta value.}
  \end{subfigure}
  \caption{\textbf{L2 convergence monitoring fails for the \sst{}.} The nonlinear recurrence does not converge to a fixed point: all difficulty groups show the same L2 profile (a), and current L2 delta has no predictive power over whether the next iteration will help or hurt (b).}
  \label{fig:convergence_monitoring}
\end{figure}

The natural first step is to look for convergence in the iterative recurrence, and L2 delta between successive iterations is the most direct way to measure it. Figure~\ref{fig:convergence_monitoring}a plots this for $1{,}536$ positions across $8$ iterations, grouped by correct iteration depth. All three difficulty groups (questions correct at iter${} = 1$, iter${} = 2$, and iter${} = 4$) collapse to the same near-zero profile by iter${} = 2$; L2 cannot distinguish a question that needs one iteration from one that needs four. On the surface this resembles fixed-point convergence, but the zoomed view (bottom row) reveals outlier traces whose L2 delta rises and falls across later iterations, superficially resembling period-2 oscillations. L2 measures only the magnitude of change, not its content, so these traces are indistinguishable from an unstable system bouncing between two attractors, a reading that would conventionally suggest stopping at iter${} = 2$. The mechanistic analysis of Section~\ref{sec:logit_dynamics} shows these are not oscillations but transitions between distinct output distributions at each iteration, invisible to L2. Figure~\ref{fig:convergence_monitoring}b confirms that current L2 delta has no predictive power over whether the next iteration will help or hurt, with the two outcomes completely interleaved at every delta value.

A full-dimensional probe reading all $5{,}376$ hidden state dimensions does have access to the granular latent information that L2 lacks; we trained such a probe and confirmed that computational unresolvedness is a real generalising feature of the position-0 latent state. But a latent belief state can be unsettled and safe (the convergent-reasoning questions of Table~\ref{tab:convergent_reasoning}, which arrive at the correct answer at every iteration depth despite ongoing latent computation), or unsettled and vulnerable (correct at the current depth but breaking at the next). Unresolvedness is present in both cases, and the probe cannot separate them. The second case contains the fragile-correctness questions where further iteration shifts the output distribution towards an incorrect answer.

The only way for the probe to prevent overthinking regression is to predict it. This requires predicting a different property from convergence monitoring: the position-$0$ latent state must encode not whether computation has settled but whether additional iteration would break the answer. Specifically, would the answer survive if every subsequent position of the generation were iterated to depth $d{+}1$? At position~$0$, this means the first generated token's latent state must encode predictive information about computation that has not yet occurred across the entire downstream sequence, a substantially stronger property than computational completeness. We find that it does. A probe reading position~$0$ at layer~$15$ prevents every overthinking regression in our evaluation, and the mechanistic and statistical analyses of Sections~\ref{sec:halting_probe_structure} and~\ref{sec:halting_probe_generalisation} confirm it reads a genuine generalising feature, not a memorised lookup. At iteration depth $d$, the probe's target is \textsc{must halt} if the question passes at staged compute depth ${\leq}d$ and fails at flat iter${} = d+1$; otherwise \textsc{safe}.

\subsection{Method}
\label{sec:halting_probe_method}

Due to the availability of flat iteration eval runs, we use GPQA-Diamond for this experiment. \textsc{must halt} labels at depth $d$ require knowing whether the question passes at staged compute depth ${\leq}d$ \emph{and} whether it fails at flat iter${} = d+1$, which requires running the full benchmark at each flat iteration depth independently. The overthinking regression measurement of Section~\ref{sec:overthinking} already runs flat iter${} = 2, 3, 4$ on GPQA-Diamond as its primary output, and the halt labels are constructed from that data. A deployed halting head would be trained on labels generated at scale from training data across domains; the present experiment is designed to establish whether the halt signal exists as a readable feature at position~0, and GPQA-Diamond provides the data to answer that question because the flat-iteration evaluations required for label construction were already computed as part of the overthinking regression analysis. The resulting training set contains $357$ position-0 hidden states from the $121$ recoverable GPQA-Diamond questions across iteration depths and model turns, split between $68$ \textsc{must halt} timesteps (drawn from $48$ of the $121$ questions) and $289$ \textsc{safe} timesteps.

We first trained a $64$-neuron two-layer MLP on this data, $\mathrm{Linear}(5376, 64) \to \mathrm{SiLU} \to \mathrm{Linear}(64, 1)$, $344{,}193$ parameters. Mechanistic analysis of the $64$-neuron probe's learned weights (Section~\ref{sec:halting_probe_structure}) shows it operates as a cooperative ensemble effectively using ten neurons, reading a $107$-dimensional feature in the hidden state identified by inference-only input-dimension ablation (Appendix~\ref{app:halting}). We then retrained a $10$-neuron probe from scratch, $\mathrm{Linear}(5376, 10) \to \mathrm{SiLU} \to \mathrm{Linear}(10, 1)$, $53{,}781$ parameters, which reproduces the same $117/198$ result while making the Cover's theorem bound of Section~\ref{sec:halting_probe_generalisation} directly informative on the output layer. The $10$-neuron probe is the probe we report; the $64$-neuron variant is referenced throughout as the subject of the weight analysis that motivated the reduction.

The probe reads the position-0 hidden state at layer~15 and outputs a scalar logit; iteration halts when the logit exceeds the threshold $\log(0.3/0.7)$. Layer~$15$ was selected from a sweep over layers $\{3, 5, 7, 10, 15, 20, 25, 29\}$ as the shallowest layer that achieves zero overthinks and passes the LOOCV generalisation test at $p < 0.05$ (full sweep in Appendix~\ref{app:halting}). We apply the probe under the same evaluation setup as Section~\ref{sec:evaluation} on GPQA-Diamond. At each iteration the probe inspects the position-0 hidden state; if the output is above threshold, iteration halts at that depth, and that depth is used for the rest of the question.

\subsection{Result}
\label{sec:halting_probe_result}

\begin{table}[H]
\centering
\small
\caption{\textbf{Halt signal probe on GPQA-Diamond ($N = 198$).} The probe at layer~15 closes $16/20 = 80\%$ of the gap between the iter${} = 1$ baseline and the staged compute capacity, while breaking zero correct answers.}
\label{tab:halting_probe}
\begin{tabular}{lcccc}
\toprule
Method & Accuracy & $\Delta$ vs iter${} = 1$ & Overthinks & Missed \\
\midrule
\sst{} iter${} = 1$ & 51.01\% & --- & --- & --- \\
Halt signal probe (layer~15) & \textbf{59.09\%} & $+8.08$pp & $\mathbf{0}$ & $4$ \\
Staged compute capacity ($i_{\max}{=}4$) & 61.11\% & $+10.10$pp & --- & --- \\
\bottomrule
\end{tabular}
\end{table}

The probe reaches $117/198 = 59.09\%$ on GPQA-Diamond, a $+8.08$pp improvement over the iter${} = 1$ baseline (McNemar exact one-sided $p = 1.5 \times 10^{-5}$). Of the $20$ recoverable questions between the iter${} = 1$ baseline and the staged compute capacity, the probe reroutes $16$ to a higher iteration depth at which they pass. Zero correct answers are broken: every discordant pair goes baseline-wrong $\to$ probe-correct, with no baseline-correct $\to$ probe-overthink.

Per-iteration \textsc{must halt} detection is $100\%$ at every training depth ($60/60$ at iter${} = 1$, $4/4$ at iter${} = 2$, $4/4$ at iter${} = 3$). The error mode is uniformly conservative: all $4$ failures are continue questions on which the probe halted at iter${} = 1$ on an already-wrong answer that would only have been corrected at a deeper iteration, leaving the model at its current (wrong) answer rather than disrupting a correct one.

\subsection{Mechanistic structure of the learned probe}
\label{sec:halting_probe_structure}

The mechanistic analysis described here was performed on the $64$-neuron probe trained first, whose larger hidden layer makes per-neuron structure readable. The $64$-neuron probe reaches the same $117/198$ accuracy as the $10$-neuron probe we report ($68/68$ \textsc{must halt}, $257/289$ \textsc{safe}). The structure identified in this section is what motivated the retraining at minimum capacity.

\paragraph{Effective probe dimensionality.} The $64$-neuron probe's hidden layer is effectively a $10$-neuron network. Per-hidden-neuron logit contribution (mean $|\mathrm{post\_silu} \cdot W_2|$ across all evaluation timesteps) concentrates in the first ten of the probe's $64$ hidden neurons: the top four account for $53\%$ of total logit mass, the top eight for $83\%$, the top ten for $93\%$. The remaining $54$ hidden neurons contribute $7\%$ and have no effect on question-level outcomes: zeroing them leaves the $117/198$ accuracy and the $68/68$ \textsc{must halt} detection unchanged, with \textsc{safe} detection shifting only marginally from $257/289$ to $250/289$ (Appendix~\ref{app:halting}). These $10$ probe neurons all read from the same set of hidden-state dimensions, differing in sign patterns rather than encoding ten independent features. This motivated retraining at $10$ neurons, which reproduces the same result and makes Cover's theorem directly applicable (Section~\ref{sec:halting_probe_generalisation}).

\paragraph{Neither half works alone.} Single-neuron ablation of the highest-contribution probe neuron (neuron~44, which by itself carries $20.4\%$ of the total logit mass) collapses the probe to $102/198 = 51.52\%$, near the iter${} = 1$ baseline. The other nine top probe neurons continue to detect all $68/68$ \textsc{must halt} timesteps after the ablation, but \textsc{safe} discrimination collapses from $257/289$ to $10/289$: they classify essentially everything as halt, equivalent to halting at iter${} = 1$ on every question. The reverse ablation, keeping neuron~44 and the $54$ minor neurons while zeroing the other nine top probe neurons, collapses in the opposite direction: $90/198 = 45.45\%$, below the iter${} = 1$ baseline, with perfect \textsc{safe} ($289/289$) but zero \textsc{must halt} detection ($0/68$) and $31$ overthinks from never halting. Neither subset reproduces the probe's accuracy on its own. The probe's decision emerges from the balance between the top-nine halt-detection signal and neuron~44's counterbalancing offset, with the threshold separating \textsc{halt} from \textsc{safe} only when both contributions are present. The effective linear direction of the probe is almost unchanged by removing neuron~44 ($r = 0.9954$ between original and ablated directions); what changes is the logit-magnitude balance needed to maintain threshold separation.

\paragraph{Near-linear operation.} The probe's effective linear direction $W_2^\top W_1$ correlates at $r = 0.905$ with gradient importance on \textsc{must halt} inputs. The SiLU nonlinearity is present but the probe is operating near the linear regime, which directly constrains how much residual nonlinear capacity is available for memorisation beyond the ten-dimensional bottleneck exploited by Section~\ref{sec:halting_probe_generalisation}.

\subsection{Generalisation versus memorisation}
\label{sec:halting_probe_generalisation}

With $53{,}781$ parameters and only $48$ distinct \textsc{must halt} questions in the $357$-timestep training set, the probe has capacity to memorise question-identity patterns and reproduce training accuracy without reading any real feature. The following tests establish that it does not.

\paragraph{Leave-one-out cross-validation.} The probe generalises in $29/48 = 60\%$ of LOOCV folds over the \textsc{must halt} questions. Under the memorisation null, the probe has no information about the held-out question and classifies it at its base prediction rate: $133/357 = 37.3\%$ of all timesteps, including false positives on \textsc{safe} timesteps (full methodology in Appendix~\ref{app:halting_loocv}). A binomial test of $29/48$ at null probability $0.373$ yields $p = 9.4 \times 10^{-4}$, with a one-sided $95\%$ Wilson confidence interval $[48.6\%, 100.0\%]$ on held-out accuracy. The $64$-neuron variant reproduces the same held-out accuracy ($29/48 = 60\%$), confirming that the reduction to ten probe neurons is not an artefact.

\paragraph{Cover's theorem capacity bound.} The $10$-neuron probe's output layer is a linear classifier in ten dimensions. By Cover's theorem~\cite{cover1965} it can shatter at most $11$ points in general position, while the minority class contains $48$ \textsc{must halt} questions ($4.4\times$ the shattering capacity). The bound applies strictly to the output layer; $W_1$ plus the SiLU activation can in principle add nonlinear capacity beyond the ten-dimensional bottleneck. The near-linear operation of the probe ($r = 0.905$, Section~\ref{sec:halting_probe_structure}) constrains how much nonlinear separation the $W_1$ layer can exploit, and Section~\ref{sec:halting_probe_feature} resolves the concern directly by showing that a linear classifier on the essential feature dimensions reads the feature more accurately than the nonlinear probe. LOOCV covers what the capacity bound does not, as an assumption-free empirical test of held-out generalisation on the full network.

\subsection{The 107-dimension feature}
\label{sec:halting_probe_feature}

\begin{figure}[H]
  \centering
  \includegraphics[width=0.65\linewidth]{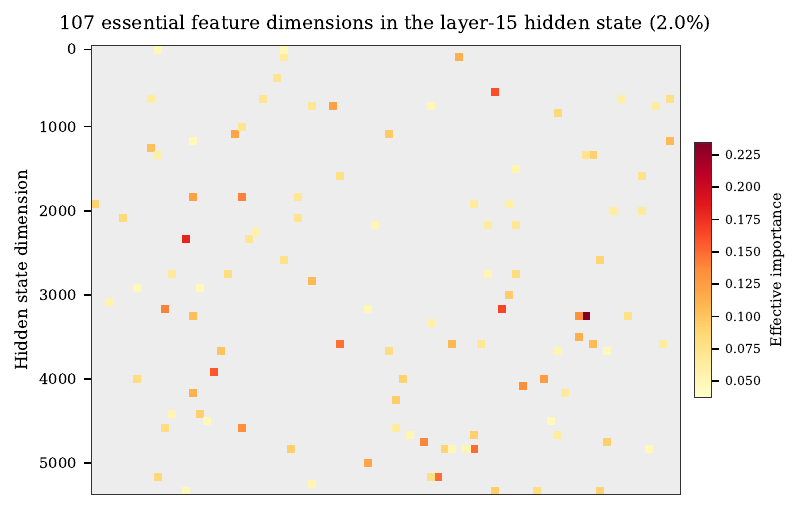}
  \caption{\textbf{The 107 essential hidden-state dimensions at layer~15.} Each cell represents one of the $5{,}376$ hidden-state dimensions ($84 \times 64$ grid, dim~$0$ at bottom-left). Coloured cells are the $107$ dimensions the probe requires; colour intensity indicates effective weight importance. The remaining $5{,}269$ dimensions (grey) can be zeroed with no effect on the probe's evaluation accuracy.}
  \label{fig:feature_position_grid}
\end{figure}

Having established that the probe reads a genuine feature, we characterise it through inference-only input-dimension ablation on the frozen trained probe (Appendix~\ref{app:halting_input_ablation}). All $5{,}376$ hidden-state dimensions are ranked by effective weight importance. Keeping only the top $K$ dimensions (zeroing the rest at inference, no retraining), the minimum $K$ that reproduces the exact baseline ($117/198$, zero overthinks, $4$ missed) is $K = 761$. Greedy pruning within this $761$-dimension set (testing each dimension individually, least important first, keeping it zeroed if the baseline holds) reduces the essential set to $107$ dimensions ($2.0\%$ of the hidden state).

The $107$ essential dimensions are scattered across the full index range of the layer-$15$ hidden state (dim~$9$ to dim~$5{,}364$), with no clustering (Figure~\ref{fig:feature_position_grid}). Zeroing any one of these $107$ dimensions degrades the probe's evaluation accuracy; zeroing the remaining $5{,}269$ dimensions has no effect. The feature is too constrained for a lookup table mapping question identities to halt decisions: $107$ dimensions encoding a binary signal for $48$ questions across $357$ timesteps is a pattern, not a hash. The $10$ probe neurons project this $107$-dimensional pattern into a scalar halt logit through cooperative sign-varying weights.

A linear probe trained directly on these $107$ dimensions ($\mathrm{Linear}(107, 1)$, $108$ parameters) achieves $39/48 = 81\%$ LOOCV accuracy ($p = 4.2 \times 10^{-5}$), exceeding the $29/48$ achieved by the nonlinear probe on the full $5{,}376$-dimensional hidden state. Since the feature is fully readable without nonlinearity, the SiLU activation in the $10$-neuron probe cannot be exploiting capacity beyond what Cover's theorem bounds on its output layer, resolving the caveat raised in Section~\ref{sec:halting_probe_generalisation}. Replacing SiLU with GELU or ReLU in the bottleneck architecture produces identical eval accuracy and LOOCV within one fold, confirming that the activation function choice is irrelevant to feature extraction. On the same trained bottleneck probe, zeroing the $5{,}269$ non-essential input dimensions at inference raises accuracy from the unablated $117/198$ to $118/198$ (zero overthinks, $4$ missed continues), consistent with the probe's training on the full hidden state having absorbed small noise contributions that vanish under ablation. By comparison, a linear probe trained directly on the $107$ dimensions achieves $115/198$ (zero overthinks, $6$ missed continues), three questions below the input-ablated bottleneck probes, identifying threshold calibration as what the nonlinearity provides.

\subsection{Position 0 predicts the downstream trajectory}
\label{sec:halting_probe_payoff}

The probe separates convergent from fragile-correctness trajectories with zero overthinks by reading the basin-shifted latent belief state at position~$0$ (Section~\ref{sec:punctuated_equilibrium}), where the model enters a content-dependent attractor region. The feature it reads is not a convergence signal; it is a prediction about the future. At the first generated token position, before any of the downstream sequence has been produced, $107$ hidden-state dimensions at layer~$15$ already encode whether the eventual answer will survive or break under additional latent computation for every subsequent position. The \sst{} computes this prediction through its own architecture; the probe is a measurement instrument that makes the prediction visible. Whether the specific dimensions carrying this signal for GPQA-Diamond would carry it for other tasks is a question about the universality of the feature, not its existence.

\section{Limitations}
\label{sec:limitations}

The two-pass training method trains at iter${} = 1$ equivalent by design (Section~\ref{sec:two_pass}), and no halting criterion can be learned end-to-end during supervised fine-tuning. Adaptive iteration depth selection requires a separate mechanism added post-hoc, or more naturally co-trained during reinforcement learning where the halting criterion can be learned alongside the task reward without separate label generation. Without adaptive depth selection, applying a uniform iteration depth to all questions produces overthinking regression (Section~\ref{sec:overthinking}).

Section~\ref{sec:halting_probe} establishes that a learned probe can read the halt signal from the position-$0$ latent state with statistically significant generalisation to held-out questions ($p < 0.001$, LOOCV), but the experiment is scoped to feasibility on a single benchmark (GPQA-Diamond) with a small \textsc{must halt} minority ($68$ timesteps from $48$ distinct questions in a $357$-timestep training set). Validating the approach at scale on training data with a large held-out set, or co-training a halting head during RL, is future work.

While the architecture supports deeper iteration, each additional depth incurs a full forward pass through the 27B-parameter model, becoming prohibitively slow for realistic deployment, so evaluation focuses on depths up to $i_{\max} = 4$ and whether accuracy continues to improve beyond this is not established.

Iteration depth at earlier positions propagates through the state stream and shapes downstream computation (Section~\ref{sec:logit_dynamics}), making per-position depth selection an architecturally meaningful option. However, this would require either a halting mechanism at every generated token or detection of basin shift positions for selective depth allocation. These are different experimental designs that could build on the feasibility established in Section~\ref{sec:halting_probe}.

While demonstrated on Gemma~3 27B, the two-pass training method is mathematically designed to apply to any pretrained transformer with a Lipschitz-bounded gated FFN (Appendix~\ref{app:approx_bound}); however, the paper does not claim universal compatibility empirically or that the accuracy increase will transfer with similar effect to other model families or scales, since unforeseen practical considerations may differ across backbones.

The \sst{} uses zero-shot greedy decoding, arguably stricter than the varying protocols under which external comparators were evaluated; however, this protocol heterogeneity is a universal property of cross-paper comparison, not specific to this evaluation. These comparisons provide context for the absolute accuracy on the evaluated benchmarks, not a claim of generalised performance parity with these models across tasks; the controlled architectural comparison is the matched baseline (Section~\ref{sec:baseline_ft}).

This paper's evaluation focuses on reasoning benchmarks, with the state stream's contribution to non-reasoning tasks such as writing or knowledge retrieval remaining future work.

\section{Conclusion}
\label{sec:conclusion}

The State Stream Transformer contributes two things that together turn latent space reasoning into a practical intervention on pretrained models rather than a pretraining commitment. The architecture preserves per-layer latent computation across positions through a nonlinear recurrence on the existing feedforward weights, giving a single mechanism both horizontal persistence across positions and vertical deliberation through iteration. The two-pass parallel training method resolves the cross-position sequential dependency that a nonlinear per-layer recurrence otherwise imposes, replacing it with an associative scan whose approximation is exact to first order in the blend coefficient and whose pass-1 outputs co-adapt with the blend during training. Any suitable pretrained transformer with a gated feedforward network could potentially be turned into an \sst{} by fine-tuning.

When co-trained to use this mechanism, the model organises its latent space into content-dependent semantic basins with sharp transitions at specific positions and stability between them, and the state stream carries each reorganisation forward through the stable regions until the next transition. Every basin shift is a commitment in continuous latent space that reshapes the computational landscape for everything downstream until the next basin shift. When given additional iteration, the model explores genuinely different trajectories through that landscape and resolves them to the same correct conclusion at rates far above chance. At the very first generation position, before any of the downstream sequence has been produced, a small subset of the latent state already encodes a prediction for whether the eventual answer will survive or break under additional iteration depth across the entire generation. The content-dependent basin dynamics, the trajectory encoding at position~0, the convergent exploration across iteration depths, and the predictive signal about future computation are all emergent from co-training with no explicit supervision of planning, halting, or deliberation behaviour. They describe what structured reasoning in continuous latent space actually looks like when the architecture makes it possible and co-training lets the model learn to use it.

The state stream offers a new axis of continuous computation for reasoning in language models, orthogonal to parameter scaling and token-space chain-of-thought and operating on the latent computation those approaches leave untouched. Whether and how this axis compounds with scale, with reinforcement learning, and with deeper iteration on larger models is the natural next question, and one the practicality of the two-pass method makes available to investigate without pretraining from scratch. The mechanistic characterisation presented here also provides a concrete handle for interpretability research on what a transformer is capable of doing when it is given the architectural means to reason in continuous latent space via a state stream.

\bibliographystyle{unsrtnat}
\bibliography{references}

\begin{thebibliography}{54}
\providecommand{\natexlab}[1]{#1}
\providecommand{\url}[1]{\texttt{#1}}
\expandafter\ifx\csname urlstyle\endcsname\relax
  \providecommand{\doi}[1]{doi: #1}\else
  \providecommand{\doi}{doi: \begingroup \urlstyle{rm}\Url}\fi

\bibitem[Fedorenko et~al.(2024)Fedorenko, Piantadosi, and
  Gibson]{fedorenko2024language}
Evelina Fedorenko, Steven~T. Piantadosi, and Edward A.~F. Gibson.
\newblock Language is primarily a tool for communication rather than thought.
\newblock \emph{Nature}, 630\penalty0 (8017):\penalty0 575--586, 2024.
\newblock \doi{10.1038/s41586-024-07522-w}.
\newblock URL \url{https://www.nature.com/articles/s41586-024-07522-w}.

\bibitem[Elhage et~al.(2021)Elhage, Nanda, Olsson, Henighan, Joseph, Mann,
  Askell, Bai, Chen, Conerly, DasSarma, Drain, Ganguli, Hatfield-Dodds,
  Hernandez, Jones, Kernion, Lovitt, Ndousse, Amodei, Brown, Clark, Kaplan,
  McCandlish, and Olah]{elhage2021circuits}
Nelson Elhage, Neel Nanda, Catherine Olsson, Tom Henighan, Nicholas Joseph, Ben
  Mann, Amanda Askell, Yuntao Bai, Anna Chen, Tom Conerly, Nova DasSarma, Dawn
  Drain, Deep Ganguli, Zac Hatfield-Dodds, Danny Hernandez, Andy Jones, Jackson
  Kernion, Liane Lovitt, Kamal Ndousse, Dario Amodei, Tom Brown, Jack Clark,
  Jared Kaplan, Sam McCandlish, and Chris Olah.
\newblock A mathematical framework for transformer circuits, 2021.
\newblock Transformer Circuits Thread,
  \url{https://transformer-circuits.pub/2021/framework/index.html}.

\bibitem[Zhang et~al.(2025)Zhang, L{\'e}vy, d'Ascoli, Rapin, Alario,
  Bourdillon, Pinet, and King]{zhang2025thought}
Mingfang Zhang, Jarod L{\'e}vy, St{\'e}phane d'Ascoli, J{\'e}r{\'e}my Rapin,
  F.-Xavier Alario, Pierre Bourdillon, Svetlana Pinet, and Jean-R{\'e}mi King.
\newblock From thought to action: How a hierarchy of neural dynamics supports
  language production, 2025.
\newblock URL \url{https://arxiv.org/abs/2502.07429}.

\bibitem[Zhu et~al.(2025)Zhu, Peng, Cheng, Qu, Huang, Zhu, Wang, Xue, Zhang,
  Shan, Cai, Kergan, Kembay, Smith, Lin, Nguyen, Pan, Chou, Cai, Wu, Zhao, Liu,
  Yang, Zhou, Zheng, Li, Zhou, Li, Zhang, Liu, Zhang, Huang, and
  Eshraghian]{zhu2025latent}
Rui-Jie Zhu, Tianhao Peng, Tianhao Cheng, Xingwei Qu, Jinfa Huang, Dawei Zhu,
  Hao Wang, Kaiwen Xue, Xuanliang Zhang, Yong Shan, Tianle Cai, Taylor Kergan,
  Assel Kembay, Andrew Smith, Chenghua Lin, Binh Nguyen, Yuqi Pan, Yuhong Chou,
  Zefan Cai, Zhenhe Wu, Yongchi Zhao, Tianyu Liu, Jian Yang, Wangchunshu Zhou,
  Chujie Zheng, Chongxuan Li, Yuyin Zhou, Zhoujun Li, Zhaoxiang Zhang, Jiaheng
  Liu, Ge~Zhang, Wenhao Huang, and Jason Eshraghian.
\newblock A survey on latent reasoning, 2025.
\newblock URL \url{https://arxiv.org/abs/2507.06203}.

\bibitem[Kaplan et~al.(2020)Kaplan, McCandlish, Henighan, Brown, Chess, Child,
  Gray, Radford, Wu, and Amodei]{kaplan2020scaling}
Jared Kaplan, Sam McCandlish, Tom Henighan, Tom~B. Brown, Benjamin Chess, Rewon
  Child, Scott Gray, Alec Radford, Jeffrey Wu, and Dario Amodei.
\newblock Scaling laws for neural language models, 2020.
\newblock URL \url{https://arxiv.org/abs/2001.08361}.

\bibitem[Shazeer et~al.(2017)Shazeer, Mirhoseini, Maziarz, Davis, Le, Hinton,
  and Dean]{shazeer2017moe}
Noam Shazeer, Azalia Mirhoseini, Krzysztof Maziarz, Andy Davis, Quoc Le,
  Geoffrey Hinton, and Jeff Dean.
\newblock Outrageously large neural networks: The sparsely-gated
  mixture-of-experts layer.
\newblock In \emph{International Conference on Learning Representations (ICLR
  2017)}, 2017.
\newblock URL \url{https://arxiv.org/abs/1701.06538}.

\bibitem[Wei et~al.(2022)Wei, Wang, Schuurmans, Bosma, Ichter, Xia, Chi, Le,
  and Zhou]{wei2022chain}
Jason Wei, Xuezhi Wang, Dale Schuurmans, Maarten Bosma, Brian Ichter, Fei Xia,
  Ed~Chi, Quoc Le, and Denny Zhou.
\newblock Chain-of-thought prompting elicits reasoning in large language
  models.
\newblock In \emph{Advances in Neural Information Processing Systems (NeurIPS
  2022)}, 2022.
\newblock URL \url{https://arxiv.org/abs/2201.11903}.

\bibitem[{OpenAI}(2024{\natexlab{a}})]{openai2024o1}
{OpenAI}.
\newblock Learning to reason with llms, 2024{\natexlab{a}}.
\newblock \url{https://openai.com/index/learning-to-reason-with-llms/}.

\bibitem[{DeepSeek-AI}(2025)]{deepseek2025r1}
{DeepSeek-AI}.
\newblock {DeepSeek-R1} incentivizes reasoning in {LLMs} through reinforcement
  learning.
\newblock \emph{Nature}, 645:\penalty0 633--638, 2025.
\newblock \doi{10.1038/s41586-025-09422-z}.
\newblock URL \url{https://www.nature.com/articles/s41586-025-09422-z}.

\bibitem[Snell et~al.(2024)Snell, Lee, Xu, and Kumar]{snell2024scaling}
Charlie Snell, Jaehoon Lee, Kelvin Xu, and Aviral Kumar.
\newblock Scaling {LLM} test-time compute optimally can be more effective than
  scaling model parameters.
\newblock In \emph{International Conference on Learning Representations (ICLR
  2025)}, 2024.
\newblock URL \url{https://arxiv.org/abs/2408.03314}.
\newblock Oral.

\bibitem[Geiping et~al.(2025)Geiping, McLeish, Jain, Kirchenbauer, Singh,
  Bartoldson, Kailkhura, Bhatele, and Goldstein]{geiping2025scaling}
Jonas Geiping, Sean McLeish, Neel Jain, John Kirchenbauer, Siddharth Singh,
  Brian~R. Bartoldson, Bhavya Kailkhura, Abhinav Bhatele, and Tom Goldstein.
\newblock Scaling up test-time compute with latent reasoning: A recurrent depth
  approach.
\newblock In \emph{Advances in Neural Information Processing Systems}, 2025.

\bibitem[Dehghani et~al.(2018)Dehghani, Gouws, Vinyals, Uszkoreit, and
  Kaiser]{dehghani2018universal}
Mostafa Dehghani, Stephan Gouws, Oriol Vinyals, Jakob Uszkoreit, and {\L}ukasz
  Kaiser.
\newblock Universal transformers.
\newblock In \emph{International Conference on Learning Representations (ICLR
  2019)}, 2018.
\newblock URL \url{https://arxiv.org/abs/1807.03819}.

\bibitem[Giannou et~al.(2023)Giannou, Rajput, Sohn, Lee, Lee, and
  Papailiopoulos]{giannou2023looped}
Angeliki Giannou, Shashank Rajput, Jy-yong Sohn, Kangwook Lee, Jason~D. Lee,
  and Dimitris Papailiopoulos.
\newblock Looped transformers as programmable computers.
\newblock In \emph{International Conference on Machine Learning (ICML 2023)},
  2023.
\newblock URL \url{https://arxiv.org/abs/2301.13196}.

\bibitem[Hao et~al.(2024)Hao, Sukhbaatar, Su, Li, Hu, Weston, and
  Tian]{hao2024coconut}
Shibo Hao, Sainbayar Sukhbaatar, DiJia Su, Xian Li, Zhiting Hu, Jason Weston,
  and Yuandong Tian.
\newblock Training large language models to reason in a continuous latent
  space, 2024.
\newblock URL \url{https://arxiv.org/abs/2412.06769}.

\bibitem[Wang et~al.(2025)Wang, Li, Sun, Chen, Liu, Wu, Lu, Song, and
  Abbasi~Yadkori]{wang2025hrm}
Guan Wang, Jin Li, Yuhao Sun, Xing Chen, Changling Liu, Yue Wu, Meng Lu, Sen
  Song, and Yasin Abbasi~Yadkori.
\newblock Hierarchical reasoning model, 2025.
\newblock URL \url{https://arxiv.org/abs/2506.21734}.

\bibitem[Gu and Dao(2024)]{gu2024mamba}
Albert Gu and Tri Dao.
\newblock Mamba: Linear-time sequence modeling with selective state spaces.
\newblock In \emph{Conference on Language Modeling (COLM)}, 2024.
\newblock URL \url{https://arxiv.org/abs/2312.00752}.

\bibitem[Aviss(2025)]{aviss2025sst}
Thea Aviss.
\newblock The state stream transformer: Emergent metacognitive behaviours
  through latent state persistence, 2025.
\newblock URL \url{https://arxiv.org/abs/2501.18356}.

\bibitem[Vaswani et~al.(2017)Vaswani, Shazeer, Parmar, Uszkoreit, Jones, Gomez,
  Kaiser, and Polosukhin]{vaswani2017attention}
Ashish Vaswani, Noam Shazeer, Niki Parmar, Jakob Uszkoreit, Llion Jones,
  Aidan~N. Gomez, {\L}ukasz Kaiser, and Illia Polosukhin.
\newblock Attention is all you need.
\newblock In \emph{Advances in Neural Information Processing Systems (NeurIPS
  2017)}, 2017.
\newblock URL \url{https://arxiv.org/abs/1706.03762}.

\bibitem[Ranzato et~al.(2015)Ranzato, Chopra, Auli, and
  Zaremba]{ranzato2015sequence}
Marc'Aurelio Ranzato, Sumit Chopra, Michael Auli, and Wojciech Zaremba.
\newblock Sequence level training with recurrent neural networks.
\newblock In \emph{International Conference on Learning Representations (ICLR
  2016)}, 2015.
\newblock URL \url{https://arxiv.org/abs/1511.06732}.

\bibitem[Gu et~al.(2021)Gu, Goel, and R{\'e}]{gu2021s4}
Albert Gu, Karan Goel, and Christopher R{\'e}.
\newblock Efficiently modeling long sequences with structured state spaces.
\newblock In \emph{International Conference on Learning Representations (ICLR
  2022)}, 2021.
\newblock URL \url{https://arxiv.org/abs/2111.00396}.

\bibitem[Smith et~al.(2022)Smith, Warrington, and Linderman]{smith2022s5}
Jimmy~T.H. Smith, Andrew Warrington, and Scott~W. Linderman.
\newblock Simplified state space layers for sequence modeling.
\newblock In \emph{International Conference on Learning Representations (ICLR
  2023)}, 2022.
\newblock URL \url{https://arxiv.org/abs/2208.04933}.

\bibitem[Wang et~al.(2024)Wang, Chen, Yuan, Zhang, Li, Peng, and
  Ji]{wang2024codeact}
Xingyao Wang, Yangyi Chen, Lifan Yuan, Yizhe Zhang, Yunzhu Li, Hao Peng, and
  Heng Ji.
\newblock Executable code actions elicit better {LLM} agents.
\newblock In \emph{International Conference on Machine Learning (ICML 2024)},
  2024.
\newblock URL \url{https://arxiv.org/abs/2402.01030}.

\bibitem[Cobbe et~al.(2021)Cobbe, Kosaraju, Bavarian, Chen, Jun, Kaiser,
  Plappert, Tworek, Hilton, Nakano, Hesse, and Schulman]{cobbe2021gsm8k}
Karl Cobbe, Vineet Kosaraju, Mohammad Bavarian, Mark Chen, Heewoo Jun, Lukasz
  Kaiser, Matthias Plappert, Jerry Tworek, Jacob Hilton, Reiichiro Nakano,
  Christopher Hesse, and John Schulman.
\newblock Training verifiers to solve math word problems, 2021.
\newblock URL \url{https://arxiv.org/abs/2110.14168}.

\bibitem[Dettmers et~al.(2023)Dettmers, Pagnoni, Holtzman, and
  Zettlemoyer]{dettmers2023qlora}
Tim Dettmers, Artidoro Pagnoni, Ari Holtzman, and Luke Zettlemoyer.
\newblock {QLoRA}: Efficient finetuning of quantized {LLMs}.
\newblock In \emph{Advances in Neural Information Processing Systems (NeurIPS
  2023)}, 2023.
\newblock URL \url{https://arxiv.org/abs/2305.14314}.

\bibitem[Hendrycks and Gimpel(2016)]{hendrycks2016gelu}
Dan Hendrycks and Kevin Gimpel.
\newblock Gaussian error linear units ({GELUs}), 2016.
\newblock URL \url{https://arxiv.org/abs/1606.08415}.

\bibitem[Hu et~al.(2021)Hu, Shen, Wallis, Allen-Zhu, Li, Wang, Wang, and
  Chen]{hu2021lora}
Edward~J. Hu, Yelong Shen, Phillip Wallis, Zeyuan Allen-Zhu, Yuanzhi Li, Shean
  Wang, Lu~Wang, and Weizhu Chen.
\newblock {LoRA}: Low-rank adaptation of large language models.
\newblock In \emph{International Conference on Learning Representations (ICLR
  2022)}, 2021.
\newblock URL \url{https://arxiv.org/abs/2106.09685}.

\bibitem[Fu et~al.(2026)Fu, You, Chen, Dai, Yang, and Wang]{fu2026think}
Tianyu Fu, Yichen You, Zekai Chen, Guohao Dai, Huazhong Yang, and Yu~Wang.
\newblock {Think-at-Hard}: Dynamic looped transformers for improved reasoning.
\newblock In \emph{Proceedings of the 43rd International Conference on Machine
  Learning}, volume 306 of \emph{Proceedings of Machine Learning Research},
  pages 32129--32154. PMLR, 2026.
\newblock URL \url{https://proceedings.mlr.press/v306/fu26t.html}.

\bibitem[Fan et~al.(2025)Fan, Du, Ramchandran, and Lee]{fan2025looped}
Ying Fan, Yilun Du, Kannan Ramchandran, and Kangwook Lee.
\newblock Looped transformers for length generalization.
\newblock In \emph{The Thirteenth International Conference on Learning
  Representations}, 2025.
\newblock URL \url{https://openreview.net/forum?id=2edigk8yoU}.

\bibitem[{Meta}(2024{\natexlab{a}})]{llama31_70b}
{Meta}.
\newblock Llama 3.1 70b instruct model card, 2024{\natexlab{a}}.
\newblock Hugging Face,
  \url{https://huggingface.co/meta-llama/Llama-3.1-70B-Instruct}.

\bibitem[{Qwen Team}(2024)]{qwen25}
{Qwen Team}.
\newblock Qwen2.5: A party of foundation models, 2024.
\newblock Qwen Blog, \url{https://qwen.ai/blog?id=qwen2.5}.

\bibitem[{Gemma Team}(2025)]{gemma3}
{Gemma Team}.
\newblock Gemma 3 technical report.
\newblock Technical report, Google, 2025.
\newblock URL \url{https://arxiv.org/abs/2503.19786}.

\bibitem[{Meta}(2024{\natexlab{b}})]{llama31_405b}
{Meta}.
\newblock Llama 3.1 405b instruct model card, 2024{\natexlab{b}}.
\newblock Hugging Face,
  \url{https://huggingface.co/meta-llama/Llama-3.1-405B-Instruct}.

\bibitem[{Meta}(2024{\natexlab{c}})]{llama33}
{Meta}.
\newblock Llama 3.3 70b instruct model card, 2024{\natexlab{c}}.
\newblock Hugging Face,
  \url{https://huggingface.co/meta-llama/Llama-3.3-70B-Instruct}.

\bibitem[{OpenAI}(2024{\natexlab{b}})]{gpt4o}
{OpenAI}.
\newblock Gpt-4o system card, 2024{\natexlab{b}}.
\newblock \url{https://openai.com/index/gpt-4o-system-card/}.

\bibitem[{DeepSeek-AI}(2024)]{deepseekv3}
{DeepSeek-AI}.
\newblock Deepseek-v3 technical report, 2024.
\newblock URL \url{https://arxiv.org/abs/2412.19437}.

\bibitem[{Google DeepMind}(2025{\natexlab{a}})]{gemini20flash}
{Google DeepMind}.
\newblock Gemini 2.0 flash model card.
\newblock
  \url{https://storage.googleapis.com/deepmind-media/Model-Cards/Gemini-2-0-Flash-Model-Card.pdf},
  2025{\natexlab{a}}.
\newblock URL
  \url{https://storage.googleapis.com/deepmind-media/Model-Cards/Gemini-2-0-Flash-Model-Card.pdf}.

\bibitem[{Google for Developers}(2025)]{gemini20pro}
{Google for Developers}.
\newblock The {Gemini} 2.0 family expands.
\newblock \url{https://developers.googleblog.com/en/gemini-2-family-expands/},
  2025.
\newblock URL
  \url{https://developers.googleblog.com/en/gemini-2-family-expands/}.

\bibitem[Rein et~al.(2023)Rein, Hou, Stickland, Petty, Pang, Dirani, Michael,
  and Bowman]{rein2023gpqa}
David Rein, Betty~Li Hou, Asa~Cooper Stickland, Jackson Petty, Richard~Yuanzhe
  Pang, Julien Dirani, Julian Michael, and Samuel~R. Bowman.
\newblock {GPQA}: A graduate-level google-proof {Q\&A} benchmark.
\newblock In \emph{Conference on Language Modeling (COLM 2024)}, 2023.
\newblock URL \url{https://arxiv.org/abs/2311.12022}.

\bibitem[{Google DeepMind}(2025{\natexlab{b}})]{gemini25pro}
{Google DeepMind}.
\newblock Gemini 2.5 pro model card, 2025{\natexlab{b}}.
\newblock \url{https://deepmind.google/technologies/gemini/pro/}.

\bibitem[Hendrycks et~al.(2021)Hendrycks, Burns, Kadavath, Arora, Basart, Tang,
  Song, and Steinhardt]{hendrycks2021math}
Dan Hendrycks, Collin Burns, Saurav Kadavath, Akul Arora, Steven Basart, Eric
  Tang, Dawn Song, and Jacob Steinhardt.
\newblock Measuring mathematical problem solving with the {MATH} dataset.
\newblock In \emph{Advances in Neural Information Processing Systems Track on
  Datasets and Benchmarks (NeurIPS 2021)}, 2021.
\newblock URL \url{https://arxiv.org/abs/2103.03874}.

\bibitem[Lightman et~al.(2023)Lightman, Kosaraju, Burda, Edwards, Baker, Lee,
  Leike, Schulman, Sutskever, and Cobbe]{lightman2023verify}
Hunter Lightman, Vineet Kosaraju, Yura Burda, Harri Edwards, Bowen Baker, Teddy
  Lee, Jan Leike, John Schulman, Ilya Sutskever, and Karl Cobbe.
\newblock Let's verify step by step.
\newblock In \emph{International Conference on Learning Representations (ICLR
  2024)}, 2023.
\newblock URL \url{https://arxiv.org/abs/2305.20050}.

\bibitem[Chen et~al.(2021)Chen, Tworek, Jun, Yuan, de~Oliveira~Pinto, Kaplan,
  Edwards, Burda, Joseph, Brockman, et~al.]{chen2021humaneval}
Mark Chen, Jerry Tworek, Heewoo Jun, Qiming Yuan, Henrique~Ponde
  de~Oliveira~Pinto, Jared Kaplan, Harri Edwards, Yuri Burda, Nicholas Joseph,
  Greg Brockman, et~al.
\newblock Evaluating large language models trained on code, 2021.
\newblock URL \url{https://arxiv.org/abs/2107.03374}.

\bibitem[Zhou et~al.(2026)Zhou, Ling, Chen, Wang, Fan, and
  Wang]{zhou2026morethinking}
Shu Zhou, Rui Ling, Junan Chen, Xin Wang, Tao Fan, and Hao Wang.
\newblock When more thinking hurts: Overthinking in {LLM} test-time compute
  scaling, 2026.
\newblock URL \url{https://arxiv.org/abs/2604.10739}.

\bibitem[H{\"a}gele et~al.(2026)H{\"a}gele, Gema, Sleight, Perez, and
  Sohl-Dickstein]{hagele2026hotmess}
Alexander H{\"a}gele, Aryo~Pradipta Gema, Henry Sleight, Ethan Perez, and
  Jascha Sohl-Dickstein.
\newblock The hot mess of {AI}: How does misalignment scale with model
  intelligence and task complexity?, 2026.
\newblock URL \url{https://arxiv.org/abs/2601.23045}.

\bibitem[Hakim(2026)]{hakim2026brevity}
MD~Azizul Hakim.
\newblock Brevity constraints reverse performance hierarchies in language
  models, 2026.
\newblock URL \url{https://arxiv.org/abs/2604.00025}.

\bibitem[Graves(2016)]{graves2016act}
Alex Graves.
\newblock Adaptive computation time for recurrent neural networks, 2016.
\newblock URL \url{https://arxiv.org/abs/1603.08983}.

\bibitem[Cover(1965)]{cover1965}
Thomas~M. Cover.
\newblock Geometrical and statistical properties of systems of linear
  inequalities with applications in pattern recognition.
\newblock \emph{IEEE Transactions on Electronic Computers}, EC-14\penalty0
  (3):\penalty0 326--334, 1965.

\bibitem[Loshchilov and Hutter(2017)]{loshchilov2017adamw}
Ilya Loshchilov and Frank Hutter.
\newblock Decoupled weight decay regularization.
\newblock In \emph{International Conference on Learning Representations (ICLR
  2019)}, 2017.
\newblock URL \url{https://arxiv.org/abs/1711.05101}.

\bibitem[Bai et~al.(2019)Bai, Kolter, and Koltun]{bai2019deq}
Shaojie Bai, J.~Zico Kolter, and Vladlen Koltun.
\newblock Deep equilibrium models.
\newblock In \emph{Advances in Neural Information Processing Systems (NeurIPS
  2019)}, 2019.
\newblock URL \url{https://arxiv.org/abs/1909.01377}.
\newblock Spotlight Oral.

\bibitem[Lim et~al.(2023)Lim, Zhu, Selfridge, and Kasim]{lim2023deer}
Yi~Heng Lim, Qi~Zhu, Joshua Selfridge, and Muhammad~Firmansyah Kasim.
\newblock Parallelizing non-linear sequential models over the sequence length.
\newblock In \emph{International Conference on Learning Representations (ICLR
  2024)}, 2023.
\newblock URL \url{https://arxiv.org/abs/2309.12252}.

\bibitem[Danieli et~al.(2025)Danieli, Rodriguez, Sarabia, Suau, and
  Zappella]{danieli2025pararnn}
Federico Danieli, Pau Rodriguez, Miguel Sarabia, Xavier Suau, and Luca
  Zappella.
\newblock {ParaRNN}: Unlocking parallel training of nonlinear {RNNs} for large
  language models.
\newblock In \emph{International Conference on Learning Representations (ICLR
  2026)}, 2025.
\newblock URL \url{https://arxiv.org/abs/2510.21450}.
\newblock Oral.

\bibitem[Combettes and Pesquet(2019)]{combettes2019lipschitz}
Patrick~L. Combettes and Jean-Christophe Pesquet.
\newblock Lipschitz certificates for layered network structures driven by
  averaged activation operators, 2019.
\newblock URL \url{https://arxiv.org/abs/1903.01014}.

\bibitem[Zhang and Sennrich(2019)]{zhang2019rmsnorm}
Biao Zhang and Rico Sennrich.
\newblock Root mean square layer normalization.
\newblock In \emph{Advances in Neural Information Processing Systems (NeurIPS
  2019)}, 2019.
\newblock URL \url{https://arxiv.org/abs/1910.07467}.

\bibitem[Gouk et~al.(2021)Gouk, Frank, Pfahringer, and
  Cree]{gouk2021regularisation}
Henry Gouk, Eibe Frank, Bernhard Pfahringer, and Michael~J. Cree.
\newblock Regularisation of neural networks by enforcing {Lipschitz}
  continuity.
\newblock \emph{Machine Learning}, 110:\penalty0 393--416, 2021.

\end{thebibliography}

\clearpage
\appendix

\section*{Appendix}

\section{Training}
\label{app:training}

\subsection{Dataset and task formulation}
\label{app:dataset}

The model is fine-tuned on GSM8K grade-school math problems formulated as CodeACT tasks. In the CodeACT paradigm, the model calls tools by emitting executable Python code as its actions rather than producing JSON. Each training example is a multi-turn interaction: the user poses a mathematical question, the model writes Python code delimited by custom \texttt{<execute>}/\texttt{</execute>} tokens, a sandboxed runtime executes the code and returns the output, and the model interprets the result, iterating with further code if needed, before producing a final answer. The \texttt{<execute>} and \texttt{</execute>} tokens are new to the model, mapped to unused vocabulary IDs (262140 and 262141 respectively).
Because the embedding layer and language modelling head share weights, LoRA adaptation of the lm\_head trains both the output distribution over these tokens and their input representations simultaneously.

This formulation decouples reasoning from arithmetic capability. The Python runtime handles computation; what the model must learn is problem decomposition, planning, and interpretation of results. Improvements from the state stream therefore reflect gains in reasoning, not in arithmetic accuracy. The multi-turn structure also provides a natural testbed for the state stream: the model must carry context across turns where tool outputs introduce new information that was not present in the original question.

Training traces were synthetically generated and filtered for quality through automated and manual review. The resulting dataset contains 6{,}579 training and 731 validation examples. Loss is computed only on model-generated tokens; user turns and tool outputs are masked.

\subsection{Training setup}
\label{app:training_setup}

The Gemma~3 27B backbone is frozen and trained via QLoRA. LoRA adapters are applied to every attention and MLP projection and to the language modelling head; the \lsc{} parameters (per-layer blend logits and state normalisation weights, 666{,}624 in total) are trained at full precision alongside. AdamW~\cite{loshchilov2017adamw} is applied with two parameter groups at separate learning rates: $10^{-2}$ for the \lsc{} parameters and $10^{-4}$ for the LoRA adapters. The higher \lsc{} rate reflects its small parameter count and the need for the per-dimension blend coefficients to move meaningfully from their initialisation; the LoRA group instead uses a 10-step linear warmup followed by cosine decay. Training runs on a single NVIDIA RTX PRO 6000 (96\,GB VRAM). Both the \sst{} and matched baseline use chunked cross-entropy loss, offloaded gradient checkpointing, and tiled MLP forward passes; under this matched optimised setup, the measured \sst{} optimiser step takes $2.08\times$ as long as the baseline step (Table~\ref{tab:training_cost}). Full hyperparameters are listed in Table~\ref{tab:hyperparameters}.

\begin{table}[H]
  \centering
  \caption{\textbf{Full training hyperparameters for the \sst{}.} Both the \sst{} and matched fine-tuned baseline (Section~\ref{sec:baseline_ft}) see the same training examples in the same deterministic order (no shuffle) so that the only varied factor between them is the architecture.}
  \label{tab:hyperparameters}
  \small
  \begin{tabular}{@{}l l l@{}}
    \toprule
    \textbf{Group} & \textbf{Hyperparameter} & \textbf{Value} \\
    \midrule
    \multirow{2}{*}{Backbone}
      & Base model              & Gemma 3 27B Instruction-Tuned \\
      & Adaptation              & QLoRA (NF4, frozen base) \\
    \midrule
    \multirow{4}{*}{LoRA adapters}
      & Rank $r$                            & 64 \\
      & Scaling $\alpha$                    & 64 \\
      & Dropout                             & 0.05 \\
      & Target modules                      & attention QKV+O, MLP \{gate, up, down\}, lm\_head \\
    \midrule
    \multirow{4}{*}{\lsc{} parameters$^{\dagger}$}
      & Total parameter count               & 666{,}624 \\
      & Precision                           & full (bf16) \\
      & Per-layer $\alpha$ logit init       & $\theta_{\mathrm{init}} = -1.8$ ($\alpha_{\mathrm{init}} \approx 0.027$) \\
      & Update mode                         & direct (Eq.~\ref{eq:state_update}) \\
    \midrule
    \multirow{6}{*}{Optimisation}
      & Optimiser               & AdamW \\
      & LR (LoRA group)         & $1\!\times\!10^{-4}$, 10-step linear warmup + cosine decay \\
      & LR (\lsc{} group)$^{\dagger}$       & $1\!\times\!10^{-2}$, constant \\
      & Weight decay            & 0 \\
      & Gradient clip (norm)    & 1.0 \\
      & Adam $\beta_1$, $\beta_2$, $\epsilon$ & 0.9, 0.999, $1\!\times\!10^{-8}$ \\
    \midrule
    \multirow{4}{*}{Batching}
      & Micro-batch size        & 1 \\
      & Gradient accumulation   & 16 \\
      & Effective batch size    & 16 \\
      & Max sequence length     & 8{,}192 \\
    \midrule
    \multirow{3}{*}{Stopping}
      & Max steps               & 2{,}000 \\
      & Early stopping patience & 4 (on validation loss) \\
      & Best checkpoint step    & 410 (\sst{}); 280 (baseline) \\
    \midrule
    \multirow{3}{*}{Data}
      & Training examples       & 6{,}579 \\
      & Validation examples     & 731 \\
      & Example ordering        & deterministic (no shuffle), identical across runs \\
    \midrule
    \multirow{5}{*}{Compute}
      & Hardware                & 1$\times$ NVIDIA RTX PRO 6000 (96\,GB) \\
      & Forward passes per step$^{\dagger}$ & 2 (\sst{}); 1 (baseline) \\
      & Memory optimisations              & chunked cross-entropy loss, offloaded gradient \\
      &                                     & checkpointing, tiled MLP forward passes \\
      & Active runtime to best ckpt        & $6.29$\,h (\sst{}); $1.94$\,h (baseline) \\
    \bottomrule
    \multicolumn{3}{l}{\footnotesize $^{\dagger}$\sst{} only; not applicable to the matched fine-tuned baseline.}
  \end{tabular}
\end{table}

\subsection{Dataset format}
\label{app:dataset_format}

Each training example is a single JSON object containing the full multi-turn interaction as a pre-formatted Gemma~3 chat template string. The format uses Gemma~3's native turn delimiters (\texttt{<start\_of\_turn>}, \texttt{<end\_of\_turn>}) and two custom tokens, \texttt{<execute>} (vocabulary ID 262140) and \texttt{</execute>} (vocabulary ID 262141), mapped to unused positions in the Gemma~3 vocabulary. A typical two-turn example follows (6{,}445 of the 6{,}579 training examples have this structure).

\begin{small}
\begin{verbatim}
<bos><start_of_turn>user
Joy can read 8 pages of a book in 20 minutes.
How many hours will it take her to read 120 pages?<end_of_turn>
<start_of_turn>model

<execute>
pages_per_minute = 8 / 20
time_hours = (120 / pages_per_minute) / 60
print(f"Time in hours: {time_hours}")
</execute><end_of_turn>
<start_of_turn>user
<tool_response>
Time in hours: 5.0
</tool_response><end_of_turn>
<start_of_turn>model

<execute>
finish_session("The answer is 5 hours.")
</execute><end_of_turn>
\end{verbatim}
\end{small}

Loss is computed only on model-turn content; all tokens between \texttt{<start\_of\_turn>model} and the corresponding \texttt{<end\_of\_turn>} (inclusive) receive labels, and all other tokens (user turns, tool responses, \texttt{<bos>}, turn delimiters outside model turns) are masked with $-100$. Of the 6{,}579 training examples, 6{,}445 contain two model turns (one code execution, one \texttt{finish\_session} call), 118 contain three model turns (an intermediate code revision), and 16 contain four or five. Median token count is 302 (mean 330, range 113--6{,}025).

\subsection{Relationship to SST V1}
\label{app:v1_comparison}

An earlier version of the \sst{} demonstrated the state stream as a parameter-free intervention on frozen pretrained weights~\cite{aviss2025sst}, but used a fixed scalar blend coefficient across all dimensions, required multiple iterations per token just to produce stable output, and stored an array of prior states in the \lsc{} of which only the most recent was ever used for blending, leaving earlier entries unused in memory. The present work addresses all three limitations: per-dimension learned blend coefficients replace the fixed scalar, the unnecessary array accumulation in the \lsc{} is removed, and a parallel training procedure for the nonlinear recurrence (Section~\ref{sec:training}) makes the \sst{} fully trainable for the first time. Training produces stable single-pass generation, freeing iterations from a stability requirement into a deliberation choice.

\subsection{Architectural ablations}
\label{app:ablations}

Two architectural ablations isolate specific design choices. Both ablations use the same training pipeline, dataset, and hyperparameters as the main \sst{} (Table~\ref{tab:hyperparameters}); the only varied factor in each case is the architectural change described.

\paragraph{Alpha initialisation.} The \sst{} initialises the per-dimension blend logits at $\theta_{\mathrm{init}} = -1.8$, corresponding to $\alpha_{\mathrm{init}} \approx 0.027$ (Section~\ref{sec:alpha}). This value was identified through untrained ablation on two different base models and is validated by training an ablation checkpoint with the logit bias removed (uniform initialisation at $\sigma(0) = 0.5$, midpoint of the $[\alpha_{\min}, \alpha_{\max}]$ range).

\begin{table}[H]
\centering
\small
\caption{\textbf{Alpha initialisation ablation on GPQA-Diamond (staged compute, $i_{\max} = 4$).} The unbiased initialisation underperforms the biased initialisation at every iteration stage. At iter${} = 1$, the unbiased \sst{} falls below the matched fine-tuned baseline ($44.4\%$ vs $45.96\%$), meaning the state stream without a biased initialisation hurts single-pass performance.}
\label{tab:alpha_init_ablation}
\begin{tabular}{lcccc}
\toprule
& iter${} = 1$ & iter${} = 2$ & iter${} = 3$ & Staged $i_{\max}{=}4$ \\
\midrule
\sst{} ($\theta_{\mathrm{init}} = -1.8$) & $51.01\%$ & $56.57\%$ & $57.58\%$ & $\mathbf{61.11\%}$ \\
\sst{} (no bias) & $44.4\%$ & $50.5\%$ & $54.0\%$ & $56.6\%$ \\
\midrule
$\Delta$ & $-6.6$pp & $-6.1$pp & $-3.6$pp & $-4.5$pp \\
\bottomrule
\end{tabular}
\end{table}

The unbiased checkpoint also trained to higher loss ($0.162$ vs $0.152$) and higher validation loss ($0.151$ vs $0.147$), with elevated sentence repetition ($23.1\%$ vs $11.6\%$; Mann--Whitney $p < 0.001$) and more detected loops ($50$ vs $13$ of $198$ questions). Extended training on more data may reduce this gap, but since the biased initialisation already produces better results with less compute, verifying this is of limited practical value.

\paragraph{Linked blend with EMA state propagation.} In the \sst{}, each position's post-feedforward output fully replaces the previous state (Eq.~\ref{eq:state_update}), and the blend coefficient $\boldsymbol{\alpha}_l$ controls only how much of the stored state is read into the current computation. The blend coefficients are independent per-layer vectors, so each of the $62$ layers learns its own per-dimension blend pattern.

An alternative variant changes both decisions. First, the state update becomes an exponential moving average: $\mathbf{C}_{l,t} = \boldsymbol{\alpha}_l \odot \mathbf{C}_{l,t-1} + (1 - \boldsymbol{\alpha}_l) \odot \mathbf{o}_{l,t}$, retaining a fraction of the previous state rather than replacing it. Second, the same $\boldsymbol{\alpha}_l$ is used for both the update retention and the read blend, linking the two operations through a single parameter. This variant was trained with the same pipeline and dataset.

\begin{table}[H]
\centering
\small
\caption{\textbf{Linked blend EMA variant on GPQA-Diamond (staged compute, $i_{\max} = 4$).} The EMA variant drops below the matched baseline at iter${} = 1$ ($40.4\%$ vs $45.96\%$) but recovers at iter${} = 2$ ($51.0\%$). Same training method, different architecture; the instability at iter${} = 1$ isolates the architectural contribution of the direct-update, per-layer-independent design.}
\label{tab:ema_variant_ablation}
\begin{tabular}{lccc}
\toprule
& iter${} = 1$ & iter${} = 2$ & Staged $i_{\max}{=}4$ \\
\midrule
\sst{} (direct update, per-layer $\alpha$) & $51.01\%$ & $56.57\%$ & $\mathbf{61.11\%}$ \\
Linked blend EMA variant & $40.4\%$ & $51.0\%$ & $51.0\%$ \\
Matched fine-tuned baseline & $45.96\%$ & --- & --- \\
\bottomrule
\end{tabular}
\end{table}

\subsection{Training curves}

\begin{figure}[H]
  \centering
  \begin{subfigure}[t]{0.49\linewidth}
    \centering
    \includegraphics[width=\linewidth]{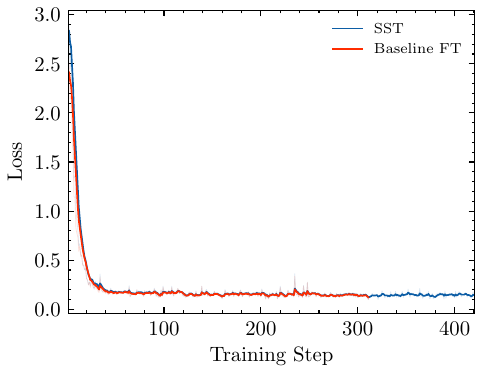}
    \caption{Training loss}
    \label{fig:sst_vs_baseline_training_loss}
  \end{subfigure}
  \hfill
  \begin{subfigure}[t]{0.49\linewidth}
    \centering
    \includegraphics[width=\linewidth]{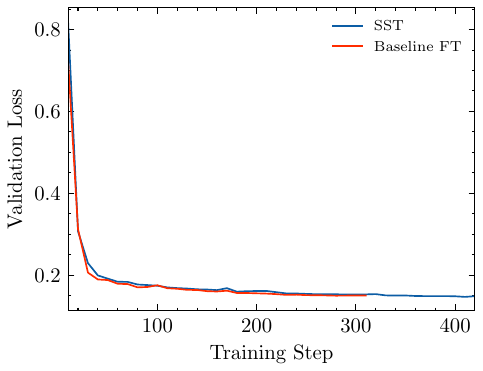}
    \caption{Validation loss}
    \label{fig:sst_vs_baseline_validation_loss}
  \end{subfigure}
  \caption{\textbf{\sst{} vs matched fine-tuned baseline.} Training loss as raw values (light) and exponential moving average (solid); validation loss as evaluated. The baseline converges at a comparable validation loss, confirming it is a fair comparator; the ablation result itself is the \sst{}--baseline delta on downstream evaluations (Section~\ref{sec:evaluation}).}
  \label{fig:sst_vs_baseline_loss}
\end{figure}

\begin{figure}[H]
  \centering
  \begin{subfigure}[t]{0.49\linewidth}
    \centering
    \includegraphics[width=\linewidth]{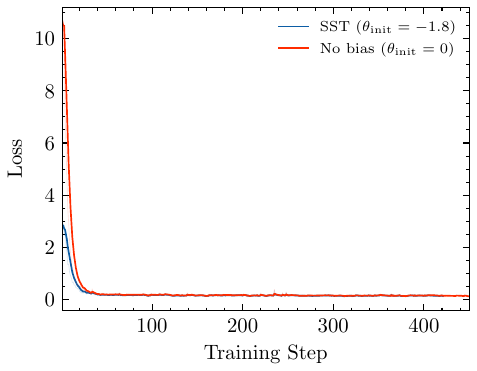}
    \caption{Training loss}
  \end{subfigure}
  \hfill
  \begin{subfigure}[t]{0.49\linewidth}
    \centering
    \includegraphics[width=\linewidth]{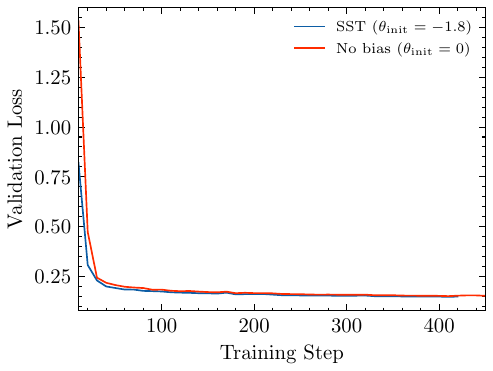}
    \caption{Validation loss}
  \end{subfigure}
  \caption{\textbf{\sst{} vs no-bias ablation variant.} The unbiased checkpoint trains to higher loss and underperforms on downstream evaluations (Table~\ref{tab:alpha_init_ablation}), confirming that a biased initialisation is necessary for stable state stream operation.}
  \label{fig:sst_vs_nobias_loss}
\end{figure}

\begin{figure}[H]
  \centering
  \begin{subfigure}[t]{0.49\linewidth}
    \centering
    \includegraphics[width=\linewidth]{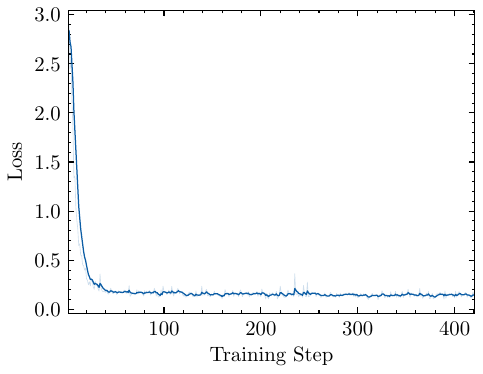}
    \caption{Training loss}
  \end{subfigure}
  \hfill
  \begin{subfigure}[t]{0.49\linewidth}
    \centering
    \includegraphics[width=\linewidth]{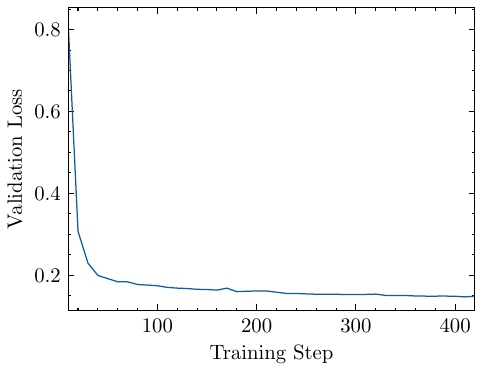}
    \caption{Validation loss}
  \end{subfigure}
  \caption{\textbf{\sst{} training and validation loss (standalone).}}
  \label{fig:sst_loss_standalone}
\end{figure}

\begin{figure}[H]
  \centering
  \begin{subfigure}[t]{0.49\linewidth}
    \centering
    \includegraphics[width=\linewidth]{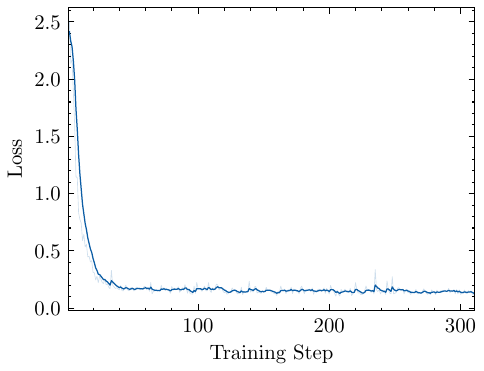}
    \caption{Training loss}
  \end{subfigure}
  \hfill
  \begin{subfigure}[t]{0.49\linewidth}
    \centering
    \includegraphics[width=\linewidth]{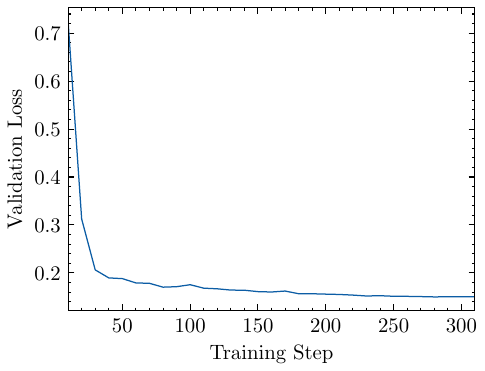}
    \caption{Validation loss}
  \end{subfigure}
  \caption{\textbf{Matched fine-tuned baseline training and validation loss (standalone).}}
  \label{fig:baseline_loss_standalone}
\end{figure}

\subsection{Per-layer alpha analysis}
\label{app:alpha_analysis}

\begin{figure}[H]
  \centering
  \includegraphics[width=0.6\linewidth]{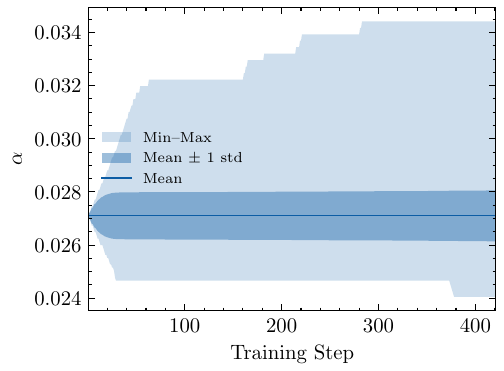}
  \caption{\textbf{Global alpha statistics over training.} Evolution of per-dimension blend coefficient statistics across training steps, showing that the alpha values adapt throughout training rather than remaining at their initialisation.}
  \label{fig:alpha_global_stats_over_training}
\end{figure}

\begin{figure}[H]
  \centering
  \includegraphics[width=\linewidth]{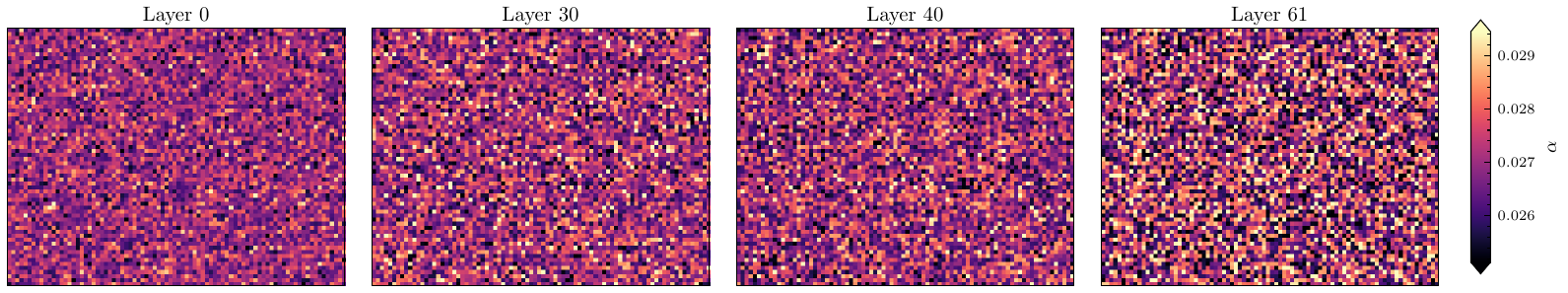}
  \caption{\textbf{Learned $\alpha_l$ at four representative depths.} Each panel shows one layer's 5{,}376-dimensional blend coefficient vector reshaped into a $64 \times 84$ grid with the same hidden-dim index mapped to the same grid position in every panel. Colour encodes raw $\alpha$ on a shared scale (p1--p99 of the full 62-layer matrix). The full 62-layer version is given in Figure~\ref{fig:alpha_panels_raw_full}.}
  \label{fig:alpha_panels_raw_selected}
\end{figure}

\begin{figure}[H]
  \centering
  \includegraphics[width=\linewidth]{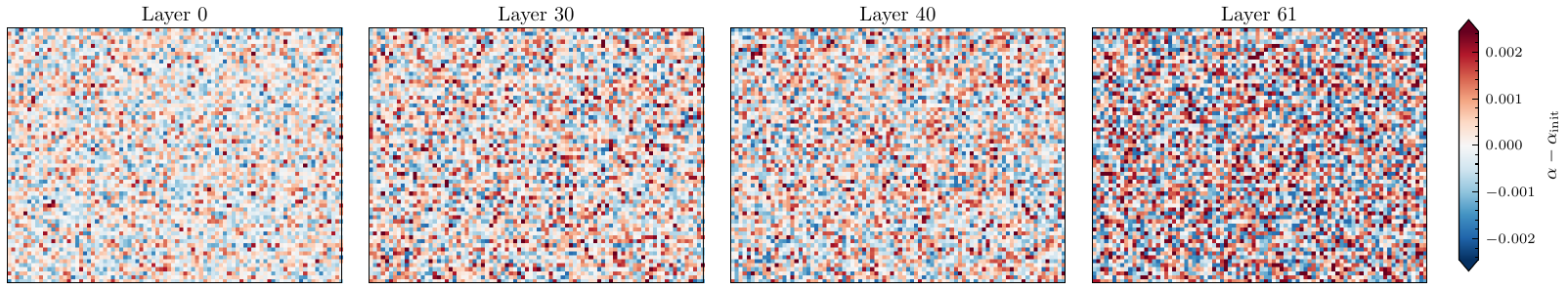}
  \caption{\textbf{Per-layer adaptation patterns at four representative depths.} Each panel shows one layer's deviation from initialisation, $\boldsymbol{\alpha}_l - \alpha_{\mathrm{init}}$, on the same $64 \times 84$ hidden-dim reshape and for the same four layers as Figure~\ref{fig:alpha_panels_raw_selected}. Red cells are dimensions the layer pushed above $\alpha_{\mathrm{init}}$; blue, below; gray, near init. The full 62-layer version is given in Figure~\ref{fig:alpha_panels_deviation_full}.}
  \label{fig:alpha_panels_deviation_selected}
\end{figure}

\begin{figure}[H]
  \centering
  \includegraphics[width=0.5\linewidth]{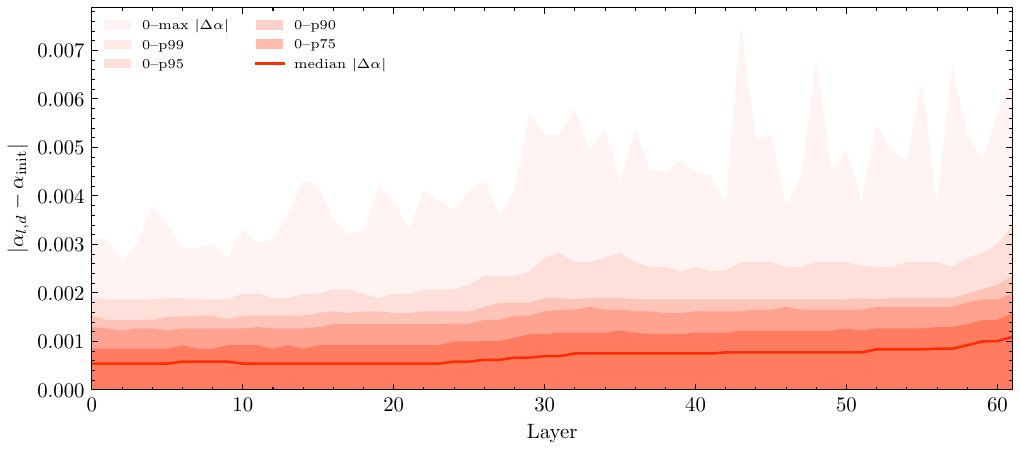}
  \caption{\textbf{Per-layer specialisation profile across depth.} Within-layer percentile bands of $|\alpha_{l,d} - \alpha_{\mathrm{init}}|$ over the 5{,}376 hidden dimensions at each layer. The median (solid red) tracks aggregate adaptation per layer; the upper-percentile bands track the magnitude of the most-adapted dimensions. The four selected layers in Figures~\ref{fig:alpha_panels_raw_selected} and~\ref{fig:alpha_panels_deviation_selected} (0, 30, 40, 61) are drawn from distinct phases of this profile.}
  \label{fig:alpha_specialisation_profile}
\end{figure}

\begin{figure}[H]
  \centering
  \includegraphics[width=\linewidth]{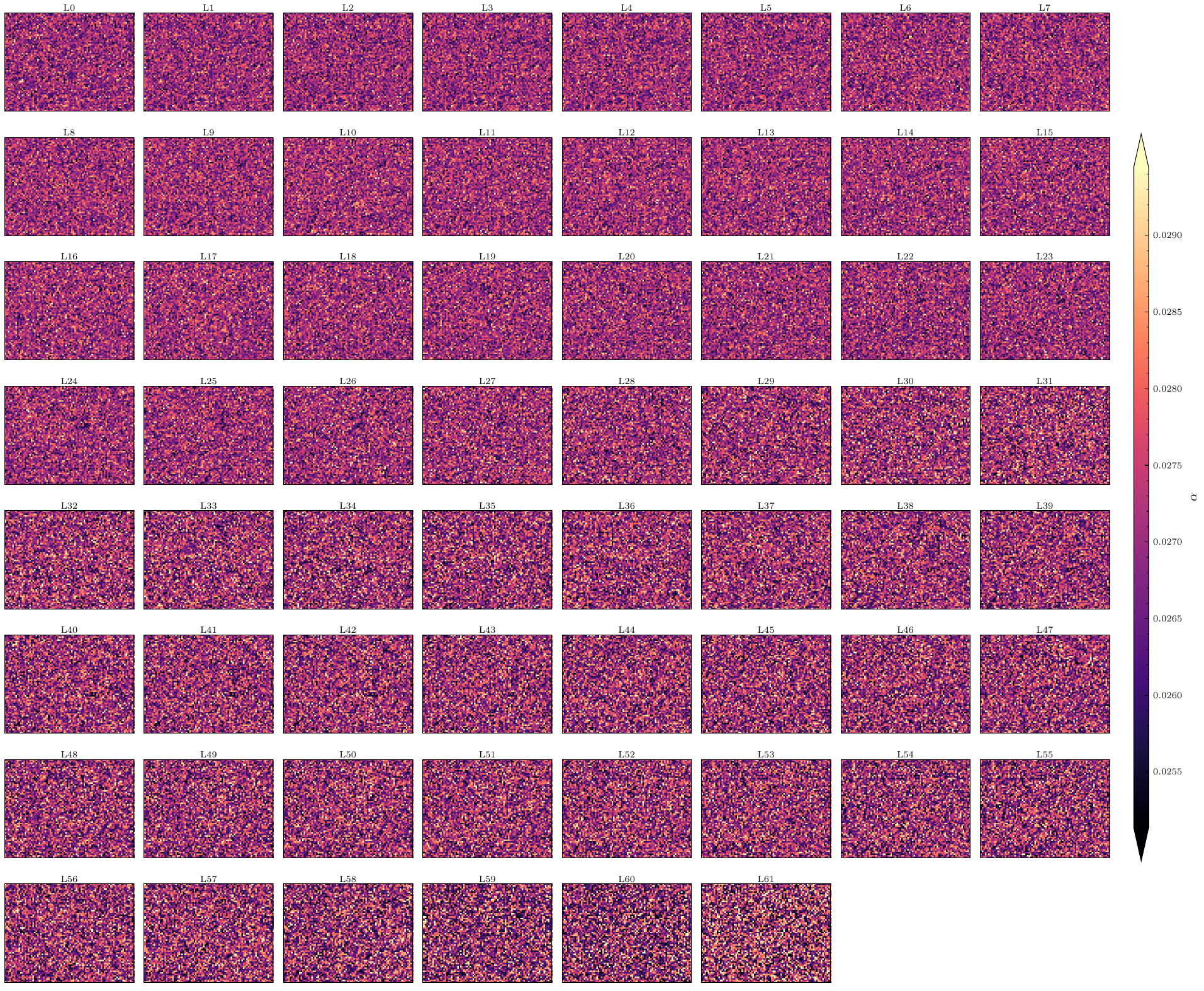}
  \caption{\textbf{Learned alpha values, all 62 layers.} Full version of Figure~\ref{fig:alpha_panels_raw_selected}. Each panel shows one layer's 5{,}376-dimensional alpha vector reshaped into a $64 \times 84$ grid, same hidden-dim mapping and shared colour scale (raw $\alpha$ values, p1--p99 range of the full matrix). Per-layer means remain close to $\alpha_{\mathrm{init}} \approx 0.027$ at every layer; per-layer variation lives in the per-dimension texture rather than overall magnitude. The deviation-from-init view (Figure~\ref{fig:alpha_panels_deviation_full}) removes the shared magnitude and makes this variation directly visible.}
  \label{fig:alpha_panels_raw_full}
\end{figure}

\begin{figure}[H]
  \centering
  \includegraphics[width=\linewidth]{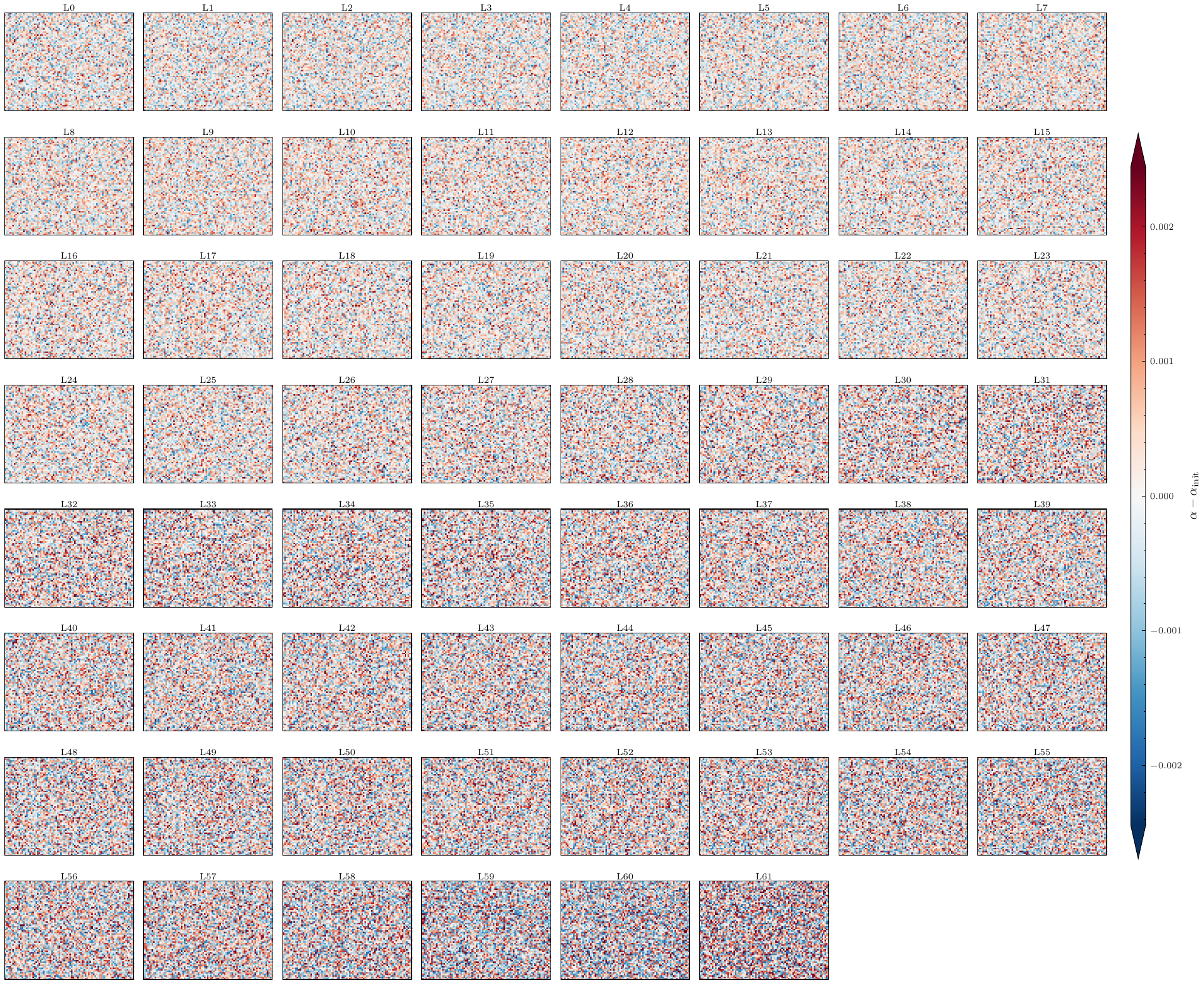}
  \caption{\textbf{Per-layer adaptation patterns, all 62 layers.} Full version of Figure~\ref{fig:alpha_panels_deviation_selected}. Each panel shows one layer's deviation $\boldsymbol{\alpha}_l - \alpha_{\mathrm{init}}$ on the same $64 \times 84$ hidden-dim reshape with a shared diverging colour scale matching the main-text selected-layer figure. The shallow plateau (approximately L0--L15), rising regime (L15--L30), mid-stack peak (L30--L35), partial quiescence (L36--L42), slow climb (L42--L57), and sharp end-stack acceleration (L57--L61) are all visible across the full layer stack.}
  \label{fig:alpha_panels_deviation_full}
\end{figure}

The following figures provide the detailed per-layer analysis of the learned blend coefficients summarised in Section~\ref{sec:learned_alpha}. Figures~\ref{fig:alpha_panels_raw_selected} and~\ref{fig:alpha_panels_deviation_selected} show four representative layers; Figures~\ref{fig:alpha_panels_raw_full} and~\ref{fig:alpha_panels_deviation_full} extend these to all 62 layers. Figure~\ref{fig:alpha_specialisation_profile} shows the magnitude of adaptation as a function of depth.

\section{Two-pass parallel training}
\label{app:two_pass}

\subsection{Parallelisation of the state stream recurrence}
\label{app:parallelisation}

During inference, the cross-position state dependency is not a problem. Autoregressive generation processes one token at a time; each position's state is computed and stored before the next position begins. During training, the model must process all positions in a sequence simultaneously to make efficient use of GPU parallelism. The sequential state dependency prevents this: a naive implementation would require $T$ serial forward passes through every layer, one per position, reducing training to the speed of sequential inference. This kind of sequential dependency is not unique to the state stream; standard autoregressive language modelling faces the same structural problem over the token sequence itself. For token prediction, the dependency is resolved at training time by substituting ground-truth tokens. No such ground truth exists for the state stream; the states must be computed by the model.

The obstacle is specifically the feedforward network in the recurrence loop. The state at position $t$ is not a linear function of the state at position $t{-}1$; it is the output of a nonlinear transformation with unique parameters at every layer (Eqs.~\ref{eq:blend}--\ref{eq:state_update}). The vertical cascade couples these into $L$ nonlinear recurrences, each dependent on the layer below.

\paragraph{Backpropagation through time.} The standard approach to training recurrent systems is to unroll the recurrence and backpropagate through time. In architectures with depth-wise recurrence, such as Universal Transformers, Geiping et al., and Coconut, the recurrence is a loop at each position: a weight-shared block iterates in depth, and the resulting state is discarded after a token is emitted. The next position starts fresh. The recurrence creates no cross-position dependencies, so teacher forcing handles positions and truncated BPTT handles the bounded iteration depth. The \sst{}'s recurrence is fundamentally different: the latent state at each layer persists across positions indefinitely within a generation sequence, creating a dependency chain that spans the full sequence length. Because each layer's recurrent state feeds into the next layer's attention, these $L$ horizontal chains are coupled through the vertical cascade and must be unrolled together across the entire sequence. BPTT through this recurrence requires unrolling all $L$ coupled per-layer recurrences across the entire training sequence, not just a bounded number of iterations per position. This is not impossible, but it is prohibitively expensive and slow. Truncated BPTT limits the unroll window but does not change the fundamental structure of the problem: the computation remains serial within each window.

\paragraph{Implicit fixed-point methods.} Methods such as DEQ~\cite{bai2019deq} avoid unrolling entirely by differentiating through an equilibrium using the implicit function theorem. This requires the architecture to be designed so that the recurrence converges to a fixed point. HRM~\cite{wang2025hrm}, for example, achieves this by having a small weight-shared module iterate within a controlled cycle until convergence, then applying the IFT at the equilibrium to obtain gradients without backpropagating through the iteration history. The \sst{}'s recurrence is across positions, not repeated application of the same function; each position receives a different post-attention input, and the per-layer feedforward networks are unconstrained pretrained nonlinearities with no contraction guarantee. There is no equilibrium to differentiate through.

\paragraph{Newton-based parallel solvers.} DEER~\cite{lim2023deer} and ParaRNN~\cite{danieli2025pararnn} cast nonlinear recurrences as systems of equations and solve them via Newton's method with parallel reductions. These methods require the Jacobian of the recurrence to have exploitable structure (diagonal or block-diagonal) for the Newton step to remain efficient. DEER has been extended with quasi-Newton approximations and Levenberg-Marquardt stabilisation, and ParaRNN scales the approach to 7B-parameter models by exploiting the sparsified Jacobians of custom GRU and LSTM variants. The \sst{}'s feedforward networks use dense $d \times 4d$ gated projections, producing a full $d \times d$ Jacobian that is neither diagonal nor block-diagonal.

\paragraph{Associative scan.} For linear recurrences of the form $\mathbf{s}_t = \mathbf{A}_t \mathbf{s}_{t-1} + \mathbf{B}_t \mathbf{x}_t$, the sequential dependency can be eliminated entirely through an associative scan, reducing $T$ serial steps to $O(\log T)$ parallel operations. This is the mechanism underlying S4~\cite{gu2021s4}, S5~\cite{smith2022s5}, and Mamba~\cite{gu2024mamba}. The \sst{}'s recurrence is not linear; the feedforward nonlinearity breaks the associative structure required by the scan.

\subsection{Why two passes, not more}
\label{app:why_two_passes}

The two-pass method approximates iter${} = 1$: the scan shifts each position's pass-1 output to the next position, so the model is trained to produce correct output from a single blend-feedforward-update cycle. A natural question is whether additional passes could approximate higher iteration depths, exposing the model to iteration during training and enabling end-to-end learning of a halting criterion. However, under teacher forcing the model sees ground-truth tokens at every position regardless of iteration depth. Training with $n$ passes could teach the model to produce correct output at the $n$th iteration given ground-truth context, incentivising it to defer reasoning to later iterations rather than resolving it in the first pass. The result could be a model that \emph{requires} $n$ iterations to perform well, destroying single-pass efficiency. By training at iter${} = 1$ equivalent, the two-pass method forces the model to be maximally capable on a single forward pass. Iteration at inference is then additional compute with the potential to improve an already-capable base, rather than a requirement for baseline competence. The consequence is that no halting criterion can be learned end-to-end during supervised fine-tuning; adaptive depth selection must be added separately, either post-hoc (Section~\ref{sec:halting_probe}) or through reinforcement learning.

\subsection{Two-pass approximation error bound}
\label{app:approx_bound}

The two-pass training method (Section~\ref{sec:two_pass}) substitutes pass-1 outputs for the true sequential states. This subsection derives the $O(\alpha^2)$ error bound on the resulting approximation.

\paragraph{Setup.} At layer $l$, position $t$, the true sequential recurrence computes:
\begin{align}
\tilde{\mathbf{h}}_{l,t} &= (\mathbf{1} - \boldsymbol{\alpha}_l) \odot \mathbf{h}_{l,t} + \boldsymbol{\alpha}_l \odot \mathrm{RMSNorm}(\mathbf{C}_{l,t-1}) \label{eq:approx_blend} \\
\mathbf{C}_{l,t} &= \tilde{\mathbf{h}}_{l,t} + \mathrm{FFN}_l(\mathrm{RMSNorm}(\tilde{\mathbf{h}}_{l,t})) \label{eq:approx_state}
\end{align}
where $\mathbf{C}_{l,t-1}$ depends on the full recursive history. The two-pass method substitutes the pass-1 output $\mathbf{o}^{(1)}_{l,t-1}$, computed without the blend, for $\mathbf{C}_{l,t-1}$.

\paragraph{Step 1: the blend perturbation is $O(\alpha)$.} Pass~1 computes $\mathbf{o}^{(1)}_{l,t}$ with the blend disabled, so the FFN input is $\mathbf{h}_{l,t}$. The true sequential computation blends before the FFN:
\[
\tilde{\mathbf{h}}_{l,t} = \mathbf{h}_{l,t} + \boldsymbol{\alpha}_l \odot \bigl(\mathrm{RMSNorm}(\mathbf{C}_{l,t-1}) - \mathbf{h}_{l,t}\bigr)
\]
The difference between the blended and unblended FFN inputs is $\boldsymbol{\alpha}_l \odot (\mathrm{RMSNorm}(\mathbf{C}_{l,t-1}) - \mathbf{h}_{l,t})$. Since $\boldsymbol{\alpha}_l \in [0.024, 0.035]$ (learned range from the trained checkpoint; architectural constraint $[\alpha_{\min}, \alpha_{\max}] = [0.015, 0.10]$) and the state-hidden difference is bounded by the activation scale, this perturbation is $O(\alpha)$.

\paragraph{Step 2: the FFN + residual preserves the $O(\alpha)$ bound.} The post-FFN computation is $f(\tilde{\mathbf{h}}) = \tilde{\mathbf{h}} + \mathrm{FFN}_l(\mathrm{RMSNorm}(\tilde{\mathbf{h}}))$. The FFN in Gemma~3 is a gated projection:
\[
\mathrm{FFN}(\mathbf{x}) = \bigl(\mathrm{GELU}_{\tanh}(\mathbf{x}W_g) \odot \mathbf{x}W_u\bigr)W_d
\]
where $W_g, W_u \in \mathbb{R}^{d \times 4d}$ and $W_d \in \mathbb{R}^{4d \times d}$. We establish that $\mathrm{FFN}_l$ has a finite Lipschitz constant $L_l$ by verifying each component operation:

\begin{itemize}[leftmargin=*, itemsep=4pt]
\item \textbf{Linear maps} $\mathbf{x} \mapsto \mathbf{x}W$ are Lipschitz with constant equal to the spectral norm $\sigma_1(W)$, the largest singular value~\cite{combettes2019lipschitz}. This is finite for any matrix with finite entries.

\item \textbf{GELU (tanh approximation)} has Lipschitz constant $\approx 1.129$, attained at $x = \sqrt{2}$ where the derivative $\Phi(x) + x\phi(x)$ reaches its maximum. The tanh approximation preserves this bound.

\item \textbf{RMSNorm}~\cite{zhang2019rmsnorm} maps $\mathbf{x} \mapsto \boldsymbol{\gamma} \odot \mathbf{x} / \mathrm{RMS}(\mathbf{x})$, projecting onto a hypersphere of radius $\sqrt{d}$ (pre-gamma) for any nonzero input. This constrains the output to a \textbf{bounded domain} regardless of input magnitude, and the radial projection is Lipschitz on $\{\mathbf{x} : \|\mathbf{x}\| \geq r\}$ for any $r > 0$ (satisfied by non-degenerate hidden states in trained transformers).

\item \textbf{Element-wise gating product} $\mathbf{a} \odot \mathbf{b}$: the product of two Lipschitz functions is Lipschitz on a bounded domain with constant $M_f L_g + M_g L_f$, where $M_f, M_g$ are function bounds and $L_f, L_g$ are Lipschitz constants. The bounded domain is enforced by the preceding RMSNorm. (This is false on unbounded domains; e.g.\ $f(x) = x$ is Lipschitz but $f(x)^2$ is not.)

\item \textbf{Composition:} the composition of finitely many Lipschitz functions is Lipschitz with constant equal to the product of the individual constants.

\item \textbf{Residual connection:} $f(\mathbf{x}) = \mathbf{x} + g(\mathbf{x})$ has Lipschitz constant $1 + L_g$ by the triangle inequality~\cite{gouk2021regularisation}.
\end{itemize}

Since all component constants are finite, the full post-FFN computation $f$ has finite Lipschitz constant $1 + L_l$:
\[
\|f(\tilde{\mathbf{h}}_{l,t}) - f(\mathbf{h}_{l,t})\| \leq (1 + L_l)\,\|\tilde{\mathbf{h}}_{l,t} - \mathbf{h}_{l,t}\|
\]
Since $\|\tilde{\mathbf{h}}_{l,t} - \mathbf{h}_{l,t}\| = O(\alpha)$ from Step~1, the post-FFN output difference is $O(\alpha)$:
\[
\mathbf{o}^{(1)}_{l,t} = \mathbf{C}_{l,t} + O(\alpha)
\]
The argument holds for any Lipschitz activation function. SiLU has Lipschitz constant ${\approx}1.100$ and ReLU has $L = 1$, so the bound applies to any pretrained backbone the \sst{} could be applied to.

\paragraph{Step 3: blending the approximation gives $O(\alpha^2)$ error.} Pass~2 blends the pass-1 output (the $O(\alpha)$ approximation) with weight $\boldsymbol{\alpha}_l$:
\[
\boldsymbol{\alpha}_l \odot \mathrm{RMSNorm}\bigl(\mathbf{o}^{(1)}_{l,t-1}\bigr) = \boldsymbol{\alpha}_l \odot \mathrm{RMSNorm}\bigl(\mathbf{C}_{l,t-1} + O(\alpha)\bigr)
\]
RMSNorm is Lipschitz, so $\mathrm{RMSNorm}(\mathbf{C} + O(\alpha)) = \mathrm{RMSNorm}(\mathbf{C}) + O(\alpha)$. Multiplying by $\boldsymbol{\alpha}_l = O(\alpha)$:
\[
\boldsymbol{\alpha}_l \odot O(\alpha) = O(\alpha^2)
\]
The error in the pass-2 blended hidden state relative to the true sequential computation is $O(\alpha^2)$. With learned $\alpha \in [0.024, 0.035]$, $\alpha^2 \in [5.8 \times 10^{-4},\, 1.2 \times 10^{-3}]$.

\paragraph{Scope of the bound.} This analysis assumes fixed weights. In practice, the LoRA adapters, blend parameters, and state normalisation weights are trained jointly across both passes, with gradients flowing from pass~2 through the scan into pass~1 (Section~\ref{sec:two_pass}). The two passes co-adapt, tightening the effective approximation beyond the fixed-weight bound.

\section{Extended mechanism analysis}
\label{app:mechanism}

\subsection{Top-$k$ overlap metric}
\label{app:topk_overlap}

Given two hidden state vectors $\mathbf{u}, \mathbf{v} \in \mathbb{R}^d$, the top-$k$ overlap is the fraction of the $k$ largest-magnitude dimensions shared between them:
\[
\mathrm{overlap}_k(\mathbf{u}, \mathbf{v}) = \frac{|\mathrm{topk}(\mathbf{u}) \cap \mathrm{topk}(\mathbf{v})|}{k},
\]
where $\mathrm{topk}(\mathbf{x})$ returns the set of indices of the $k$ dimensions with the largest $|x_i|$. This measures which dimensions are most active in each representation, ignoring the sign and magnitude of the activations themselves. Overlap of $1.0$ means the two vectors have the same $k$ most-active dimensions; overlap of $0.0$ means the active sets are disjoint.

We use $k = 1024$ throughout, approximately $19\%$ of the $d = 5{,}376$ hidden dimensions. This choice is large enough that stable positions consistently produce overlap above $0.99$ (the metric is not saturated), while sensitive enough that basin shift positions drop sharply to as low as $0.30$. Alternative values of $k$ preserve the separation between stable and low-overlap positions; $k = 1024$ provides the widest dynamic range in the separation.

\subsection{Basin shift classification via GMM}
\label{app:gmm}

The threshold separating stable from basin-shift positions in Sections~\ref{sec:punctuated_equilibrium}--\ref{sec:logit_dynamics} is derived from a two-component Gaussian mixture model fitted to the full top-$1024$ overlap distribution. The two-component choice reflects the bimodal structure visible in the raw overlap data of Figure~\ref{fig:overlap_heatmap}: stable positions with overlap near $1.0$ across every layer, and low-overlap positions appearing as vertical streaks of dark cells cascading through the layer stack. For all $198$ questions, $10$ generated positions, and $62$ layers, the iter${} = 1$ vs iter${} = 4$ overlap is computed, yielding $N = 122{,}760$ values. A two-component GMM is fitted via expectation-maximisation (scikit-learn \texttt{GaussianMixture}, seed $42$, $500$ maximum iterations):

\begin{table}[H]
\centering
\small
\begin{tabular}{lccc}
\toprule
Component & Mean & Std & Weight \\
\midrule
Stable & $0.990$ & $0.004$ & $86.2\%$ \\
Low-overlap & $0.869$ & $0.092$ & $13.8\%$ \\
\bottomrule
\end{tabular}
\end{table}

The crossover threshold ($0.976$) is the overlap value at which the two components have equal posterior probability. The order-of-magnitude difference in standard deviation ($0.004$ vs $0.092$) reflects the two distinct regimes: stable positions cluster tightly near $1.0$, while low-overlap positions span a wide range of divergence magnitudes.

\paragraph{Robustness to component count.} As a sanity check on the two-component fit, GMMs were also fitted at $3$--$5$ components. The additional components subdivide the low-overlap tail into sub-populations of different magnitudes, consistent with the content-dependent variation described in Section~\ref{sec:punctuated_equilibrium} (per-question low-overlap counts ranging from $1$ to $7$, per-position rates varying from $4\%$ to $42\%$, overlap values spanning $0.30$--$0.97$). The crossover threshold varies by less than $0.003$ across the $2$--$5$ component fits, so the partition between stable and basin-shift positions is robust to component count.

\begin{figure}[H]
  \centering
  \includegraphics[width=0.7\linewidth]{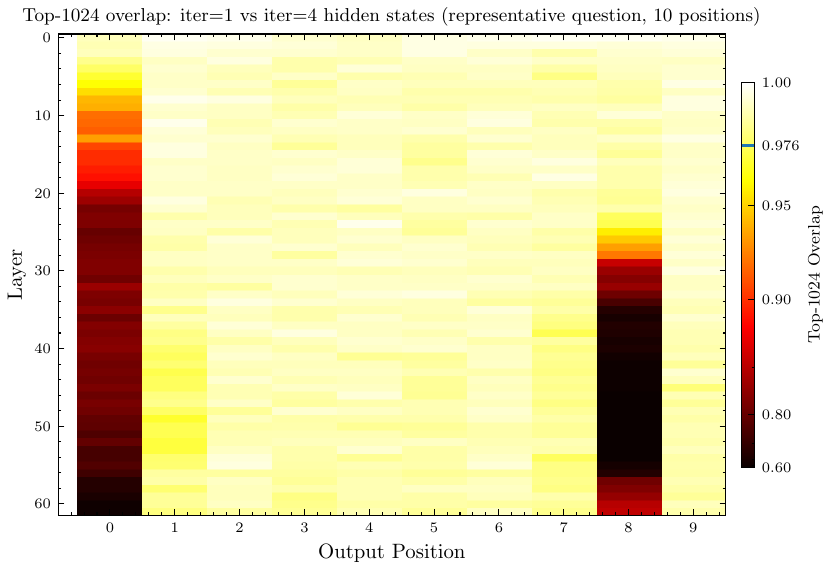}
  \caption{\textbf{Top-1024 overlap heatmap, zoomed to the first 10 generated positions} (companion to Figure~\ref{fig:overlap_heatmap}). Same metric and threshold; a separate question chosen to display the basin-shift cascade structure at the per-position scale. Position-0 universal basin shift, content-dependent shifts at later positions, stable positions in between.}
  \label{fig:overlap_heatmap_zoom}
\end{figure}

\begin{figure}[H]
  \centering
  \includegraphics[width=\linewidth]{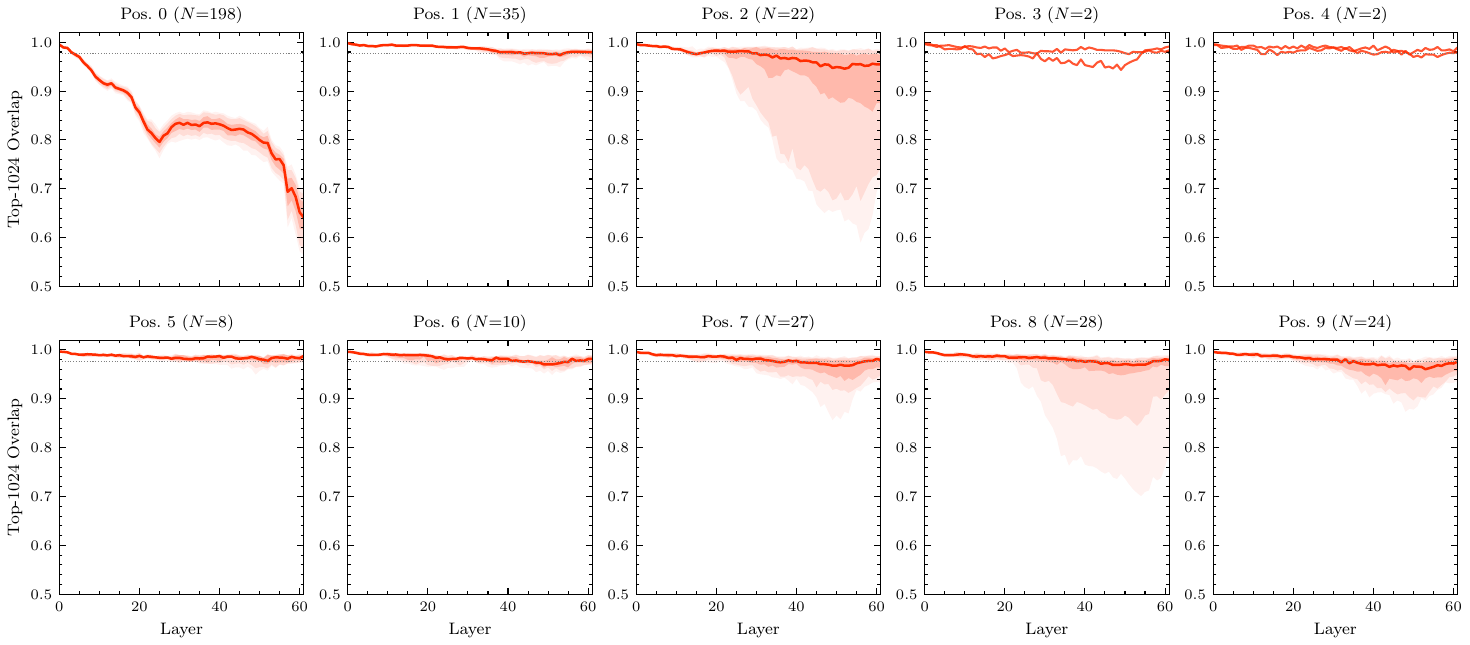}
  \caption{\textbf{Layer profile at basin shift positions, all sequence positions.} Same metric as Figure~\ref{fig:layer_profile}, shown separately for each of the first 10 generated positions. For $N \geq 5$: median, IQR, p10--p90, and p5--p95 bands. For $N < 5$: individual traces. Position~0 is universal and has a distinct layer profile, with the representational reorganisation spanning the full layer stack. Positions 3--4 are near-immune to basin shifts; their rare occurrences barely cross the threshold. Position~2 is bimodal: most occurrences are shallow, but a content-dependent subset shows reorganisation as deep as position~0 (p5--p95 band reaching below 0.4). Positions 5--9 share a common structure: near-flat early layers, onset at the layer~25 feedforward activity band, and a trough in the middle-to-late layers. The character and depth of the reorganisation varies by question content within each position, as reflected in the width of the percentile bands.}
  \label{fig:layer_profile_per_position}
\end{figure}

\begin{figure}[H]
  \centering
  \includegraphics[width=\linewidth]{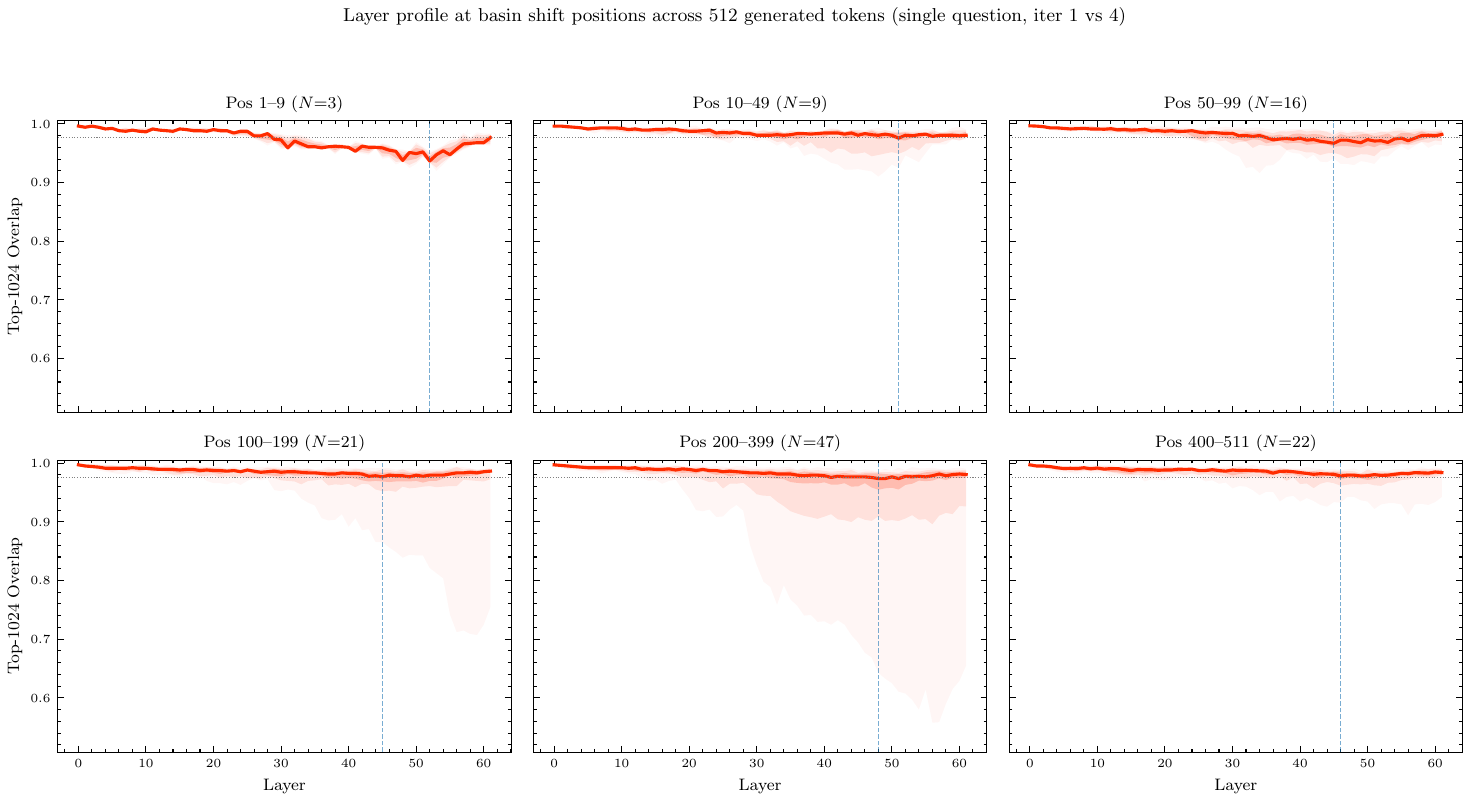}
  \caption{\textbf{Layer profile over 512 generated positions.} Same metric as Figure~\ref{fig:layer_profile}, extended to 512 positions of generated output. The punctuated equilibrium structure persists well beyond the 10-position window analysed in Section~\ref{sec:punctuated_equilibrium}, with basin shift positions continuing to appear at content-dependent intervals throughout the generation.}
  \label{fig:layer_profile_512pos}
\end{figure}

\subsection{Logit dynamics methodology}
\label{app:logit_methodology}

The logit dynamics analysis of Section~\ref{sec:logit_dynamics} uses two complementary comparison methods, determined by whether the token history is shared across the measurements being compared.

\paragraph{Cross-run comparisons (pre-divergence).} At pre-divergence positions (defined in Section~\ref{sec:punctuated_equilibrium}), the token history is identical across independent runs at different iteration depths, so any difference in the output distribution is causally attributable to the state stream. The analysis compares the final-iteration output of each independent run across all six pairwise comparisons ($1$v$2$, $1$v$3$, $1$v$4$, $2$v$3$, $2$v$4$, $3$v$4$). Each comparison yields a set of clean position records; a position can be pre-divergence for one pair but not another, so each pair is tracked independently with its own clean range. The top-1 to top-2 logprob gap is measured in the lower-iteration run's output distribution. Overlap regime (low-overlap or stable) is assigned per position using the top-$1024$ overlap from the higher-iteration run's layer states, thresholded at $0.976$ (Appendix~\ref{app:gmm}). Cross-run comparisons measure the total state stream effect: the combined consequence of additional vertical passes at the current position and the horizontal propagation of differently-iterated states from all prior positions. These two axes cannot be separated in the cross-run measurement (Section~\ref{sec:punctuated_equilibrium}). The general characterisation of the phenomenon~(1--2) uses all six pairs; the cross-run side of the output-distribution reorganisation comparison~(5) restricts to the three $1{\to}X$ pairs ($1$v$2$, $1$v$3$, $1$v$4$) to match the within-run structure where iter${} = 1$ is always the baseline.

\paragraph{Within-run comparisons (all positions).} Comparing between iterations within a single run isolates the local vertical iteration effect. Each run at iteration depth $d$ stores the output distribution at every iteration from $1$ to $d$; comparing iteration $i$ to iteration $j$ within the same run shares both the token history and the horizontally-propagated states from prior positions, since these are identical across all iterations within the run. Any difference is therefore attributable to the additional vertical passes at the current position alone. The within-run analysis uses all six structurally-matched comparisons: iter${} = 1$ vs iter${} = 2$ from the iter${} = 2$, $3$, and $4$ runs; iter${} = 1$ vs iter${} = 3$ from the iter${} = 3$ and $4$ runs; and iter${} = 1$ vs iter${} = 4$ from the iter${} = 4$ run. Because all iterations within the same run share the same generated token sequence, within-run comparisons are not restricted to pre-divergence positions; all positions are valid. This produces a larger sample than the comparable cross-run analysis (which uses three $1{\to}X$ pairs for structural equivalence) because each within-run $1$v$N$ comparison exists in every run that contains both iterations (giving six comparisons per question), and because cross-run must discard positions after the first argmax divergence where different runs have generated different tokens. This method is used for the gap distributions~(3) and output-distribution reorganisation~(5) of Section~\ref{sec:logit_dynamics}, where the claim is about what each overlap regime does locally.

\paragraph{Within-run consecutive transitions.} The later-iteration dynamics~(6) of Section~\ref{sec:logit_dynamics} use a separate within-run analysis: consecutive iteration pairings ($1\!\rightarrow\!2$, $2\!\rightarrow\!3$, $3\!\rightarrow\!4$) within the iter${} = 4$ run. This measures whether later vertical passes continue, reverse, or reinforce the changes made by earlier passes at the same position. The same token-history-sharing property applies. The per-token direction-consistency and residual-drift statistics in paragraph~(6) additionally use a cross-run analysis restricted to the $950$ positions that are pre-divergence for all three consecutive cross-run pairs simultaneously, where token history is shared across the four runs and the state stream therefore remains the sole source of variation.

\subsection{Logit dynamics tables}
\label{app:logit_tables}

\begin{table}[H]
\caption{\textbf{Argmax change rates by overlap regime (cross-run, all six pairs).} Low-overlap positions are $3.8\times$ more likely to change the argmax than stable positions.}
\label{tab:argmax_concentration}
\centering\small
\begin{tabular}{lccc}
\toprule
 & $N$ records & Argmax changes & Rate [$95\%$ CI] \\
\midrule
Low-overlap & 2{,}090 & 251 & $12.0\%$ [$10.7\%$, $13.5\%$] \\
Stable & 6{,}489 & 224 & $3.5\%$ [$3.0\%$, $3.9\%$] \\
\midrule
Total & 8{,}579 & 475 & \\
\multicolumn{4}{l}{OR $= 3.82$, $p < 0.001$} \\
\bottomrule
\end{tabular}
\end{table}

\begin{table}[H]
\caption{\textbf{Exact tie rates and logprob shifts at argmax changes (cross-run, all six pairs).} Only $11.6\%$ of argmax changes are exact ties; the remainder override a token with strictly higher probability.}
\label{tab:decision_boundary}
\centering\small
\begin{tabular}{lcc}
\toprule
 & Count & Rate [$95\%$ CI] \\
\midrule
Exact ties (gap $= 0$) & 55/475 & $11.6\%$ [$8.8\%$, $14.8\%$] \\
Non-ties (gap $> 0$) & 420/475 & $88.4\%$ [$85.2\%$, $91.2\%$] \\
\midrule
\multicolumn{3}{l}{\emph{Logprob shift of top-1 token between iteration depths}} \\
At argmax changes & \multicolumn{2}{l}{median $0.87$ nats ($e^{0.87} = 2.39\times$), mean $0.92$ nats} \\
At non-changes & \multicolumn{2}{l}{median $0.05$ nats ($e^{0.05} = 1.05\times$), mean $0.12$ nats} \\
\bottomrule
\end{tabular}
\end{table}

\begin{table}[H]
\caption{\textbf{Convergent reasoning on GPQA-Diamond.} Among the 60 questions correct at all four iteration depths, breakdown by the number of unique text traces across the four depths. A question demonstrates convergent reasoning only if all four depth outputs are (a) the correct answer and (b) pairwise distinct text traces. The observed rate is compared against the chance baseline of independent random 4-option guessing across the four depths.}
\label{tab:convergent_reasoning}
\centering\small
\begin{tabular}{lr}
\toprule
 & Count \\
\midrule
Questions correct at all four depths & 60 \\
\quad with 4 pairwise distinct text traces (convergent reasoning) & \textbf{53} \\
\quad with 3 unique text traces (one depth pair identical) & 7 \\
\quad with 2 or fewer unique text traces (deterministic repetition) & 0 \\
\midrule
Observed convergent-reasoning rate & $53/198 = 26.77\%$ \\
Bootstrap $95\%$ CI on observed rate ($10{,}000$ resamples) & $[20.71\%, 32.83\%]$ \\
\midrule
Chance baseline: $P(\text{all 4 correct}) = (1/4)^4 = 1/256$ & $0.39\%$ \\
Null expected count out of 198 & $0.77$ \\
Observed / chance ratio & $68.53\times$ \\
Binomial exact test (one-sided greater) & $p = 7.62 \times 10^{-80}$ \\
\bottomrule
\end{tabular}
\end{table}

\begin{table}[H]
\caption{\textbf{Position 0 divergence patterns.} All 198 questions show a basin shift at position 0, producing three distinct downstream outcomes.}
\label{tab:pos0_divergence}
\centering\small
\begin{tabular}{lcc}
\toprule
Outcome & $N$ & $\%$ \\
\midrule
\multicolumn{3}{l}{All $198$ questions show basin shift at position $0$} \\
\midrule
Diverge at position 0 & 75 & $37.9\%$ \\
Diverge at positions 1--9 & 66 & $33.3\%$ \\
Diverge beyond 10-position window & 57 & $28.8\%$ \\
\midrule
\multicolumn{3}{l}{All $57$ late-divergence confirmed different text at different depths} \\
\bottomrule
\end{tabular}
\end{table}

\begin{table}[H]
\caption{\textbf{Basin shift activity across consecutive iteration pairings.} Within the iter${=}4$ run ($N=1{,}980$), basin shifts can be sustained, one-shot, or late-onset.}
\label{tab:basin_shift_activity}
\centering\small
\begin{tabular}{lc}
\toprule
\multicolumn{2}{l}{\emph{Basin shift activity (iter${=}4$ run, $N = 1{,}980$)}} \\
\midrule
Sustained (all 3 pairings) [$95\%$ CI] & $186$ ($9.4\%$ [$8.1\%$, $10.8\%$]) \\
At $1{\to}2$ only & $226$ ($11.4\%$ [$10.1\%$, $12.9\%$]) \\
New at $2{\to}3$ (stable at $1{\to}2$) & $15$ ($0.8\%$ [$0.4\%$, $1.3\%$]) \\
New at $3{\to}4$ (stable at $1{\to}2$) & $14$ ($0.7\%$ [$0.4\%$, $1.2\%$]) \\
\bottomrule
\end{tabular}
\end{table}

\begin{table}[H]
\caption{\textbf{Later-iteration computation by regime.} Cumulative logprob shift trajectories, argmax reinforcement at basin shift positions, and question-level stabilisation.}
\label{tab:later_iter_computation}
\centering\small
\begin{tabular}{lcc}
\toprule
 & Basin shift & Stable \\
\midrule
\multicolumn{3}{l}{\emph{Cumulative logprob shift direction across iterations $2 \to 3 \to 4$}$^{\dagger}$} \\
$N$ (logit trajectories) & 15{,}647 & 31{,}257 \\
Cumulative ratio ($2{\to}4$ / $2{\to}3$) [$95\%$ CI] & $1.081$ [$1.062$, $1.101$] & $0.878$ [$0.867$, $0.888$] \\
Logits continuing same direction ($> 1.0$) & $27.3\%$ & $19.2\%$ \\
Logits reversing $> 50\%$ of shift & --- & $31.9\%$ \\
Logits retaining within $10\%$ & --- & $33.5\%$ \\
Logits losing $10\%$--$50\%$ & --- & $15.4\%$ \\
Direction reversal rate & --- & $61.4\%$ ($N = 46{,}904$) \\
Cumulative $2{\to}4$ / $2{\to}3$ (all positions) & \multicolumn{2}{c}{$0.95$} \\
Regime difference & \multicolumn{2}{c}{$p < 10^{-56}$, CIs do not overlap} \\
\midrule
\multicolumn{3}{l}{\emph{Basin shift argmax reinforcement ($N=81$ positions changed at $1{\to}2$)}$^{\ddagger}$} \\
Re-changed at $2{\to}3$ [$95\%$ CI] & \multicolumn{2}{c}{$4/81$ ($4.9\%$ [$1.4\%$, $12.2\%$])} \\
Re-changed at $3{\to}4$ [$95\%$ CI] & \multicolumn{2}{c}{$5/81$ ($6.2\%$ [$2.0\%$, $13.8\%$])} \\
Later-transition argmax changes (within iter${=}4$ run) & \multicolumn{2}{c}{$37$ ($19$ at $2{\to}3$, $18$ at $3{\to}4$; vs $88$ at $1{\to}2$)} \\
\midrule
\multicolumn{3}{l}{\emph{Argmax stabilisation after $1 \to 2$ change (cross-run, $N = 139$ questions)}} \\
Changed at $1{\to}2$, then stabilised & \multicolumn{2}{c}{$113/139$ ($81.3\%$)} \\
Changed at $1{\to}2$, further changes & \multicolumn{2}{c}{$26/139$ ($18.7\%$)} \\
\bottomrule
\multicolumn{3}{l}{\footnotesize $^{\dagger}$Cross-run, restricted to positions pre-divergence across all three consecutive run pairs ($N=950$). $^{\ddagger}$Within iter${=}4$ run.}
\end{tabular}
\end{table}

\begin{table}[H]
\caption{\textbf{Output-distribution reorganisation at low-overlap argmax changes.} Within-run isolates local iteration; cross-run measures total state stream effect.}
\label{tab:reorg_lo_argmax}
\centering\small
\begin{tabular}{lcc}
\toprule
Low-overlap argmax changes & Within ($N=439$) & Cross ($N=237$) \\
\midrule
\multicolumn{3}{l}{\emph{Top-100 token replacement}} \\
Mean [$95\%$ CI] & $53.8$ [$52.9$, $54.7$] & $53.3$ [$52.1$, $54.5$] \\
p25 / med / p75 & 47 / 54 / 60 & 47 / 54 / 59 \\
Range & 1--76 & 13--76 \\
\midrule
\multicolumn{3}{l}{\emph{Top-5 tokens that remain in top-100}} \\
Rank shift (med / p95 / max) & $1.0$ / $6.0$ / $36$ & $1.0$ / $6.0$ / $36$ \\
Logprob shift range (nats) & $[-3.78, +6.82]$ & $[-3.78, +6.82]$ \\
\midrule
\multicolumn{3}{l}{\emph{Argmax suppression (nats)}} \\
Mean [$95\%$ CI] & $-1.21$ [$-1.25$, $-1.17$] & $-1.16$ [$-1.22$, $-1.10$] \\
Median & $-1.11$ & $-1.07$ \\
IQR & $[-1.55, -0.88]$ & $[-1.53, -0.82]$ \\
Range & $[-2.68, -0.06]$ & $[-2.68, -0.13]$ \\
Every instance negative & Yes & Yes \\
\midrule
\multicolumn{3}{l}{\emph{New argmax winner}} \\
Rank in original dist & 1--50 (med 4) & 1--50 (med 4) \\
In original top-100 & 439/439 & 237/237 \\
\midrule
\multicolumn{3}{l}{Tokens replaced Mann-Whitney $p < 0.001$, suppression $p < 0.001$.} \\
\bottomrule
\end{tabular}
\end{table}

\begin{table}[H]
\caption{\textbf{Output-distribution reorganisation at stable argmax changes.} The within-run/cross-run divergence (median 1 vs 20 tokens replaced) isolates horizontal propagation from upstream basin shifts.}
\label{tab:reorg_st_argmax}
\centering\small
\begin{tabular}{lcc}
\toprule
Stable argmax changes & Within ($N=81$) & Cross ($N=180$) \\
\midrule
\multicolumn{3}{l}{\emph{Top-100 token replacement}} \\
Mean [$95\%$ CI] & $1.0$ [$0.8$, $1.2$] & $23.0$ [$21.5$, $24.5$] \\
p25 / med / p75 & 0 / 1 / 2 & 15 / 20 / 28 \\
Range & 0--3 & 10--56 \\
\midrule
\multicolumn{3}{l}{\emph{Top-5 tokens that remain in top-100}} \\
Rank shift (med / p95 / max) & $0.0$ / $1.0$ / $2$ & $1.0$ / $3.0$ / $76$ \\
Logprob shift range (nats) & $[-0.17, +0.16]$ & $[-3.41, +7.46]$ \\
\midrule
\multicolumn{3}{l}{\emph{Argmax suppression (nats)}} \\
Mean [$95\%$ CI] & $-0.068$ [$-0.073$, $-0.062$] & $-0.623$ [$-0.701$, $-0.545$] \\
Median & $-0.0625$ & $-0.44$ \\
IQR & $[-0.078, -0.055]$ & $[-0.80, -0.30]$ \\
Range & $[-0.16, -0.02]$ & $[-3.41, +0.07]$ \\
Every instance negative & Yes & No \\
\midrule
\multicolumn{3}{l}{\emph{New argmax winner}} \\
Rank in original dist & 2 (always) & 1--4 (med 2) \\
In original top-100 & 81/81 & 180/180 \\
\midrule
\multicolumn{3}{l}{Tokens replaced Mann-Whitney $p < 0.001$, suppression $p < 0.001$.} \\
\bottomrule
\end{tabular}
\end{table}

\begin{table}[H]
\caption{\textbf{Low-overlap positions, all (including non-flip).} Token replacement and argmax change rates across the full population.}
\label{tab:reorg_lo_all}
\centering\small
\begin{tabular}{lcc}
\toprule
Low-overlap all positions & Within ($N=2{,}101$) & Cross ($N=1{,}046$) \\
\midrule
\multicolumn{3}{l}{\emph{Top-100 token replacement}} \\
Mean [$95\%$ CI] & $32.7$ [$31.6$, $33.8$] & $48.2$ [$47.4$, $49.0$] \\
p25 / med / p75 & 3 / 46 / 57 & 41 / 50 / 57 \\
Range & 0--76 & 11--80 \\
\midrule
Argmax change rate [$95\%$ CI] & $20.9\%$ [$19.2$, $22.7$] & $22.7\%$ [$20.2$, $25.3$] \\
\bottomrule
\end{tabular}
\end{table}

\begin{table}[H]
\caption{\textbf{Stable positions, all (including non-flip).} The within-run/cross-run gap (median 1 vs 20) confirms horizontal propagation dominates stable-position reorganisation.}
\label{tab:reorg_st_all}
\centering\small
\begin{tabular}{lcc}
\toprule
Stable all positions & Within ($N=9{,}779$) & Cross ($N=1{,}885$) \\
\midrule
\multicolumn{3}{l}{\emph{Top-100 token replacement}} \\
Mean [$95\%$ CI] & $1.3$ [$1.3$, $1.3$] & $22.0$ [$21.6$, $22.5$] \\
p25 / med / p75 & 1 / 1 / 2 & 16 / 20 / 26 \\
Range & 0--18 & 7--63 \\
\midrule
Argmax change rate [$95\%$ CI] & $0.8\%$ [$0.7$, $1.0$] & $9.5\%$ [$8.3$, $11.0$] \\
\bottomrule
\end{tabular}
\end{table}

\begin{table}[H]
\caption{\textbf{Causal ordering of hidden state change and argmax divergence.} Low overlap always precedes or coincides with the first argmax disagreement; no exceptions in 198 questions.}
\label{tab:causal_ordering}
\centering\small
\begin{tabular}{lcc}
\toprule
Group & $N$ & Pattern \\
\midrule
Simultaneous (low overlap + argmax flip at pos 0) & 75 & $37.9\%$ \\
Low overlap precedes argmax flip (1--9 pos earlier) & 66 & $33.3\%$ \\
Low overlap at pos 0, divergence beyond window & 57 & $28.8\%$ \\
\midrule
\multicolumn{3}{l}{Exceptions (argmax flip before low overlap): $0/198$} \\
\bottomrule
\end{tabular}
\end{table}

\begin{table}[H]
\caption{\textbf{Gap distributions at argmax-changing positions by regime (within-run).} Low-overlap positions override large gaps; stable positions change only at exact ties or the minimum representable gap.}
\label{tab:gap_distributions}
\centering\small
\begin{tabular}{lcc}
\toprule
Within-run (iter${=}1$ vs iter${=}2,3,4$) & Low-overlap ($N=439$) & Stable ($N=81$) \\
\midrule
Exact ties (gap $= 0$) [$95\%$ CI] & $5.0\%$ [$3.2\%$, $7.5\%$] & $54.3\%$ [$42.9\%$, $65.4\%$] \\
Gap $< 0.25$ nats [$95\%$ CI] & $15.3\%$ [$12.0\%$, $19.0\%$] & $100\%$ [$95.5\%$, $100\%$] \\
Gap $< 1.0$ nat [$95\%$ CI] & $39.0\%$ [$34.4\%$, $43.7\%$] & $100\%$ [$95.5\%$, $100\%$] \\
Gap $> 1.0$ nat [$95\%$ CI] & $58.3\%$ [$53.5\%$, $63.0\%$] & $0\%$ [$0\%$, $4.5\%$] \\
Mean gap (nats) [$95\%$ CI] & $1.61$ [$1.49$, $1.74$] & $0.057$ [$0.044$, $0.071$] \\
Std & $1.34$ & $0.062$ \\
Range & $[0.00, 6.375]$ & $[0.00, 0.125]$ \\
\midrule
\multicolumn{3}{l}{Gap: Mann-Whitney $p < 0.001$. Exact tie rate: Fisher's exact OR $= 0.044$, $p < 0.001$.} \\
\bottomrule
\end{tabular}
\end{table}

\subsection{Blend perturbation exceeds bf16 precision}
\label{app:precision_proof}

\paragraph{Null hypothesis.} The between-iteration hidden state differences are deterministic floating point error propagation through the forward pass, not real FFN computation on different inputs.

\paragraph{bf16 machine epsilon.} The gap between $1.0$ and the next representable bf16 value, computed via \texttt{torch.nextafter(1.0, 2.0)} in bf16, is exactly $2^{-7} = 1/128 = 0.0078125$. Adding $2^{-7}$ to $1.0$ in bf16 produces $1.0078125$; adding $2^{-8}$ produces $1.0$ (the increment vanishes). Machine epsilon is therefore $\epsilon = 2^{-7}$, and the maximum relative rounding error on a value $x$ is $|x| \times \epsilon$.

\paragraph{Analytical bound from the learned weights.} The blend (Eq.~\ref{eq:blend}) modifies the FFN input at layer $l$, dimension $d$ by $\alpha_{l,d} \times (\mathrm{RMSNorm}(\mathbf{C}_{l,t-1}) - \mathbf{h}_{l,t})_d$. For this perturbation to be indistinguishable from bf16 rounding error on $\mathbf{h}_{l,t,d}$:
\[
\alpha_{l,d} \times |\Delta_d| \leq \epsilon \times |\mathbf{h}_{l,t,d}|
\]
In the most favourable case for the null ($|\Delta_d| = |\mathbf{h}_{l,t,d}|$), the activation magnitude cancels and the condition reduces to $\alpha_{l,d} \leq \epsilon$. The $333{,}312$ learned blend coefficients were extracted from the trained checkpoint by computing $\alpha_{l,d} = \alpha_{\min} + (\alpha_{\max} - \alpha_{\min}) \cdot \sigma(\theta_{l,d})$ from the stored logit vectors $\theta_{l,d}$ at each of the $62$ layers. They range from $0.024$ to $0.035$ (p1--p99: $0.025$--$0.029$). The minimum value is $3.08\epsilon$. Zero of the $333{,}312$ values fall below $3\epsilon$. The blend coefficient alone exceeds bf16 precision at every dimension of every layer, establishing that whenever the state-hidden difference $\Delta_d$ is non-trivial, the perturbation to the FFN input is above the precision floor.

\paragraph{Empirical measurement at basin shift positions.} The analytical bound establishes that the blend coefficient is above precision. The empirical measurement confirms that the realised perturbation (blend coefficient times state-hidden difference, propagated through the FFN and vertical cascade) exceeds precision by a wide margin at the positions that drive the largest representational changes and the most consequential downstream effects on the output distribution.

The measurement is restricted to basin shift positions (top-$1024$ overlap $< 0.976$; Section~\ref{sec:punctuated_equilibrium}). This restriction requires justification because it could be mistaken for selecting positions where deltas are large then measuring that deltas are large. The two metrics are structurally independent: the basin shift classification measures \emph{set membership} (which dimensions appear in the top-$1024$ by magnitude, Appendix~\ref{app:topk_overlap}), while the precision test measures \emph{magnitude ratio} (the absolute delta at each dimension divided by the bf16 rounding error at that dimension's activation scale). A position can have low overlap (many dimensions swap in and out of the top-$1024$) while having small absolute deltas (the swapping dimensions have similar magnitudes), and vice versa. Basin shift positions are where the causal claim of Section~\ref{sec:punctuated_equilibrium} applies; these are the positions where the state stream reorganises the representation, so they are where the precision question is relevant. Relatively stable positions also contribute computational work (Section~\ref{sec:logit_dynamics}), but the basin shift positions drive the most consequential downstream changes and provide the strongest test of the null.

For all $198$ GPQA-Diamond questions, all $10$ generated positions, the iter${} = 4$ evaluation run stores per-iteration hidden states at all $62$ layers. Basin shift positions are identified from the top-$1024$ overlap between iter${} = 1$ and iter${} = 4$ across layers $20$--$35$ (the peak FFN activity band), yielding $353$ basin shift positions out of $1{,}980$ total ($17.8\%$). At each basin shift position, layer, and dimension, the ratio $|\delta_d| / (\epsilon \times |\mathbf{h}^{(\mathrm{iter}=1)}_{l,d}|)$ is computed, where $\delta_d$ is the between-iteration delta. This produces $117{,}588{,}532$ per-dimension measurements across the iter~$1 \to 2$ transition.

\paragraph{Null threshold.} Under the rounding null, bf16 round-to-nearest produces rounding error bounded by $\frac{1}{2}\epsilon|x|$ per operation. The delta between two iterations involves two forward passes through the same FFN on slightly different inputs; the rounding errors are deterministic and correlated (same weights, same operations, same rounding mode). In the worst case of maximally anticorrelated rounding (one pass rounds up, the other rounds down on the same operation), the delta from rounding alone is at most $\epsilon|x|$ per dimension, so the ratio should be ${\leq}1$. Testing at a null proportion of $0.5$ (at most half of dimensions exceed $1\times$) is therefore generous to the null.

\begin{table}[H]
\centering
\small
\caption{\textbf{Per-dimension delta-to-precision ratio at basin shift positions, iter~$\mathbf{1 \to 2}$.} Ratio $|\delta_d| / (\epsilon \times |h_d|)$ summarised across $353$ basin shift positions $\times$ $5{,}376$ dimensions at each layer. Under the rounding null, this ratio should be ${\leq}1$.}
\label{tab:precision_proof}
\begin{tabular}{ccccccc}
\toprule
Layer & Median & p25 & p75 & p95 & Frac $> 1\times$ & $p$ \\
\midrule
$0$  & $2.9\times$ & $0.9\times$ & $4.2\times$ & $14\times$ & $71.9\%$ & ${<}10^{-300}$ \\
$15$ & $11.1\times$ & $3.2\times$ & $31.7\times$ & $163\times$ & $90.4\%$ & ${<}10^{-300}$ \\
$30$ & $17.7\times$ & $4.6\times$ & $49.3\times$ & $248\times$ & $92.6\%$ & ${<}10^{-300}$ \\
$45$ & $25.2\times$ & $5.7\times$ & $56.4\times$ & $239\times$ & $93.3\%$ & ${<}10^{-300}$ \\
$61$ & $28.9\times$ & $4.8\times$ & $83.0\times$ & $397\times$ & $92.0\%$ & ${<}10^{-300}$ \\
\midrule
All $62$ layers & $15.4\times$ & & & & $91.3\%$ & ${<}10^{-300}$ \\
\bottomrule
\end{tabular}
\end{table}

Across all $62$ layers combined, $91.3\%$ of per-dimension measurements exceed the bf16 precision floor (one-sided binomial test against null proportion $0.5$, $p < 10^{-300}$; $95\%$ CI on fraction above $1\times$: $[0.913, 1.0]$). All $62$ layers individually reach significance at $p < 0.001$. The median ratio grows from $2.9\times$ at layer~$0$ to $28.9\times$ at layer~$61$, confirming that the vertical cascade amplifies the perturbation through successive layers of FFN computation. Later iteration transitions ($2 \to 3$, $3 \to 4$) show smaller margins at the early layers, consistent with the first iteration performing the bulk of the representational reorganisation, but remain significant ($p < 10^{-300}$, $62.5\%$ and $61.4\%$ above $1\times$ globally) with all layers from $12$+ reaching individual significance.

\section{Evaluation methodology}
\label{app:eval_methodology}

The evaluation is built on a controlled architectural comparison between the \sst{} and the matched fine-tuned baseline (Section~\ref{sec:baseline_ft}). The two models share every aspect of training except the architecture itself. The data pipeline applies no shuffle; example ordering is a deterministic function of the training step, so both models process identical examples in identical order throughout training. This is stronger than seed-matched reproducibility, where a shared random seed produces the same shuffled sequence. Here there is no stochastic element in the data pipeline to control for. The resulting training dynamics confirm the design: both models follow the same loss trajectory at different magnitudes (Figure~\ref{fig:sst_vs_baseline_loss}), and the baseline converges at a comparable validation loss, establishing that the two models had equivalent training experiences and that any difference in downstream performance reflects the architectural change alone.

All evaluation uses greedy argmax decoding, making generation fully deterministic within a fixed execution configuration. There are three reasons for this choice. First, it makes the matched comparison exact. Each model produces a single output per question, and each paired outcome is a direct observation of model capability. If the \sst{} answers a question correctly and the baseline does not, that is a real discordant pair. Under stochastic evaluation, each model's accuracy on a question becomes a probability estimated from finite samples, and the paired comparison must distinguish the true capability difference from estimation variance in both rates simultaneously. Greedy decoding eliminates this estimation layer; the comparison operates on actual outputs, not on estimates of output distributions. Second, the mechanistic analysis of Section~\ref{sec:mechanism} compares the model's behaviour across iteration depths, and any stochastic element in generation would introduce variation indistinguishable from the state stream's effect. Greedy decoding ensures that differences between iteration depths within an evaluation protocol are causally attributable to the state stream alone. Third, greedy decoding is the strategy coherent with the \sst{}'s reasoning mechanism. The state stream deliberates in latent space before committing to a token, reshaping the output distribution through the blend-feedforward-update cycle at every layer (Section~\ref{sec:logit_dynamics}). The argmax of the resulting distribution is the product of that deliberation. Stochastic sampling from the post-deliberation distribution would partially undo this by sometimes selecting tokens that the latent computation specifically moved away from. The remaining sources of non-determinism in argmax are tie-breaking among near-equal logits and batch-size-dependent numerical variation in scaled dot-product attention. All headline and staged-compute evaluations use a fixed batch size of $10$ on the same physical GPU. This fixes the fused-attention execution path and makes greedy generation reproducible within that protocol. The independent flat-depth sweep in Section~\ref{sec:overthinking} used a different batch size because evaluating every question at every depth under the original configuration was infeasible within the available single-GPU evaluation budget. Because changing batch configuration can alter floating-point execution and subsequent autoregressive trajectories, the flat and staged results are treated as separate evaluation protocols rather than interchangeable accuracy estimates. Empirical validation of the headline configuration is reported in Appendix~\ref{app:determinism}.

The \sst{} is evaluated at iteration depths 1 through 4. The architecture's latent compute ceiling may be higher at deeper iteration counts, but each additional iteration is a full forward pass through a 27B-parameter model. Beyond 4 iterations, the compute cost per token exceeds what is practical for production inference. The evaluation focuses on the iteration range that is actually deployable, because reasoning improvements that require impractical compute budgets are not meaningful for a feasible architecture.

Correctness adjudication for GSM8K, MATH-500, and GPQA-Diamond uses an LLM judge (Claude Opus~4.6) provided with the model's full trace and the benchmark's ground-truth answer; the judge emits a discrete correctness flag. We prefer this over regex-based answer extraction, which is brittle in the presence of natural-language reasoning chains and tends to produce false negatives whenever the model's surface form deviates from the expected pattern. To control for judge stochasticity, every benchmark evaluation was scored twice and inspected for any answer-flip between the two runs; no answers flipped. Cases the judge did not flag as fully certain were then human-verified; across the four-benchmark evaluation this changed two questions, both instances where the judge had incorrectly marked a correct answer as wrong because the model's solution disagreed with the benchmark's labelled ground truth. HumanEval is scored by unit-test execution and does not use the judge.

\section{Determinism validation}
\label{app:determinism}

The headline and staged-compute evaluations use greedy argmax decoding on the same physical GPU (NVIDIA RTX PRO 6000) with a fixed batch size of $10$ (Section~\ref{sec:evaluation}). Fixing the batch size ensures that scaled dot-product attention follows the same kernel path on every run, eliminating batch-size-dependent numerical variation within this protocol. To verify that the forward pass is deterministic across repeated runs at a fixed configuration, we run two tests. First, a single GPQA-Diamond question is answered 5 times at iter${} = 4$ with 1{,}024 tokens per run at batch size 1: all 5 runs produce identical token sequences. Second, 10 questions are generated as a batch of 10, 5 times at iter${} = 4$ with 512 tokens per question: all 5 runs produce identical outputs across all 5{,}087 tokens per run.

\section{Inference latency and cost}
\label{app:inference_cost}

Only failures restart at the next depth: $198 \rightarrow 97 \rightarrow 86 \rightarrow 84$ questions are evaluated at depths 1--4, newly resolving $101$, $11$, $2$, and $7$. Exhaustively measuring the capacity bound therefore costs
\[
198(1)+97(2)+86(3)+84(4)=986
\]
stack-pass equivalents per generated token, averaging $4.98$ per question with a maximum of $10$. If the final question-level depth were selected before generation, the corresponding execution cost would be
\[
[101(1)+11(2)+2(3)+84(4)]/198=2.35
\]
forward passes per generated token, averaged across questions, with a maximum of $4$.

We measured latency to token emission using the evaluated checkpoints at batch size $1$, with accelerated scaled dot-product attention, KV caching, and CUDA synchronisation on the NVIDIA RTX PRO 6000. Table~\ref{tab:inference_latency} reports the median over 36 post-prefill token emissions per condition.

\begin{table}[H]
\centering
\small
\caption{\textbf{Measured inference latency at each iteration depth.} Median latency over 36 post-prefill token emissions per condition, measured at batch size $1$ using accelerated scaled dot-product attention, KV caching, and CUDA synchronisation on the NVIDIA RTX PRO 6000. Ratios are relative to the matched baseline.}
\label{tab:inference_latency}
\begin{tabular}{lcc}
\toprule
Configuration & ms/token & vs baseline \\
\midrule
Matched baseline & 50.1 & $1.00\times$ \\
\sst{} iter${} = 1$ & 57.9 & $1.15\times$ \\
\sst{} iter${} = 2$ & 112.2 & $2.24\times$ \\
\sst{} iter${} = 3$ & 167.1 & $3.33\times$ \\
\sst{} iter${} = 4$ & 222.0 & $4.43\times$ \\
\bottomrule
\end{tabular}
\end{table}

These measurements characterise the current non-optimised PyTorch research implementation, which has no \sst{}-specific fused GEMM or CUDA kernels for the blend--normalise--state-update path. They report current implementation latency rather than an accuracy-per-compute comparison or a projection of optimised deployment performance.

\section{Halting probe details}
\label{app:halting}

This appendix contains full tables and supporting data for the halting probe analysis of Section~\ref{sec:halting_probe}.

\subsection{Probe training}
\label{app:halting_training}

Both the $10$-neuron reported probe and the $64$-neuron variant used for the mechanistic analysis are trained with the identical pipeline. Architecture: $\mathrm{Linear}(5376, d) \to \mathrm{SiLU} \to \mathrm{Linear}(d, 1)$ with $d \in \{10, 64\}$. Optimiser: Adam at learning rate $10^{-3}$, batch size $32$, $60$ epochs, binary cross-entropy loss. Class balance: the minority class (\textsc{must halt}) is duplicated to match majority count, then a weighted random sampler draws equal expected counts of each class. Random seed $42$. No regularisation beyond the bottleneck. Inference halts when the output logit exceeds $\log((1 - 0.7)/0.7)$.

Training data comes from the $121$ recoverable GPQA-Diamond questions (those with a correct iteration depth within $i_{\max} = 4$, identified from the staged compute results of Section~\ref{sec:capacity}). Each model turn contributes one training timestep per iteration depth at which a hidden state exists up to that question's correct depth. Labels are \textsc{must halt} at depth $d$ if the question passes at staged depth ${\leq}d$ and fails at flat depth $d+1$; \textsc{safe} otherwise. This produces $68$ \textsc{must halt} and $289$ \textsc{safe} timesteps across $357$ total.

\subsection{Layer comparison}

\begin{table}[H]
\centering
\small
\caption{\textbf{Halt signal probe layer sweep ($64$-neuron probe).} Evaluation and LOOCV results for each tested layer. Layer~$15$ selected for zero overthinks \emph{and} statistically significant LOOCV; layers $20$ and $29$ hit higher accuracy but LOOCV does not rule out memorisation at those depths. LOOCV $p$-value: one-sided binomial test at null probability equal to the probe's base prediction rate (see Appendix~\ref{app:halting_loocv}). The $10$-neuron probe achieves the same $29/48$ LOOCV with a stronger $p = 9.4 \times 10^{-4}$ due to a lower base prediction rate ($37.3\%$ vs $42.1\%$).}
\label{tab:halting_probe_layer_sweep}
\begin{tabular}{cccccccc}
\toprule
Layer & Correct & Accuracy & $\Delta$ vs iter${}{=}1$ & OT & Miss & LOOCV & LOOCV $p$ \\
\midrule
$3$  & $111/198$ & $56.06\%$ & $+5.05$pp & $5$ & $5$ & --- & --- \\
$5$  & $114/198$ & $57.58\%$ & $+6.57$pp & $2$ & $5$ & --- & --- \\
$7$  & $113/198$ & $57.07\%$ & $+6.06$pp & $2$ & $6$ & $33/45 = 73\%$ & $0.114$ \\
$10$ & $116/198$ & $58.59\%$ & $+7.58$pp & $1$ & $4$ & --- & --- \\
$\mathbf{15}$ & $\mathbf{117/198}$ & $\mathbf{59.09\%}$ & $\mathbf{+8.08}$\textbf{pp} & $\mathbf{0}$ & $\mathbf{4}$ & $\mathbf{29/48 = 60\%}$ & $\mathbf{0.008}$ \\
$20$ & $119/198$ & $60.10\%$ & $+9.09$pp & $1$ & $1$ & $21/48 = 44\%$ & $0.331$ \\
$25$ & $115/198$ & $58.08\%$ & $+7.07$pp & $0$ & $6$ & --- & --- \\
$29$ & $119/198$ & $60.10\%$ & $+9.09$pp & $0$ & $2$ & $22/48 = 46\%$ & $0.233$ \\
\bottomrule
\end{tabular}
\end{table}

Head-to-head layer~$15$ versus layer~$7$ (McNemar paired test): $p = 0.19$, $95\%$ CI on net advantage $[-2.2, +4.9]$ questions, $5$ discordant pairs. The two layers are not statistically distinguishable on accuracy; layer~$15$ is preferred for zero overthinks and significant LOOCV at the conventional $0.05$ threshold. Layer~$7$'s LOOCV is at $p = 0.114$ on a smaller test set ($N = 45$ rather than $48$ because its two overthinks shift which questions count as \textsc{must halt}), which falls just short of significance.

\subsection{Per-iteration detection and depth breakdown}

\begin{table}[H]
\centering
\small
\caption{\textbf{Per-iteration \textsc{must halt} detection and \textsc{safe} correctness at layer~$15$ ($64$-neuron probe).} Both the $64$-neuron and $10$-neuron probes detect all $68/68$ \textsc{must halt} timesteps. The $10$-neuron probe has slightly lower overall \textsc{safe} correct-left-alone ($224/289$ vs $257/289$), reflecting additional harmless false halts that do not convert any correct answer to an overthink.}
\label{tab:halting_probe_per_iter}
\begin{tabular}{lcccc}
\toprule
Depth & \textsc{must halt} total & \textsc{must halt} detected & \textsc{safe} total & \textsc{safe} left alone \\
\midrule
iter${} = 1$ & $60$ & $60$ ($100\%$) & $212$ & $185$ ($87\%$) \\
iter${} = 2$ & $4$  & $4$  ($100\%$) & $43$  & $38$  ($88\%$) \\
iter${} = 3$ & $4$  & $4$  ($100\%$) & $17$  & $17$  ($100\%$) \\
iter${} = 4$ & $0$  & ---            & $17$  & $17$  ($100\%$) \\
\midrule
Total        & $68$ & $68$ ($100\%$) & $289$ & $257$ ($89\%$) \\
\bottomrule
\end{tabular}
\end{table}

Iteration-depth breakdown of probe decisions across the $117$ correctly-answered questions at layer~$15$: $27$ at iter${} = 1$, $74$ at iter${} = 2$, $9$ at iter${} = 3$, $7$ at iter${} = 4$. Most correctness concentrates at iter${} = 2$, consistent with the overthinking-regression profile of Section~\ref{sec:overthinking}: iter${} = 2$ recovers the largest set of questions that iter${} = 1$ misses without yet incurring the deeper-iteration regressions.

\begin{figure}[H]
  \centering
  \includegraphics[width=\linewidth]{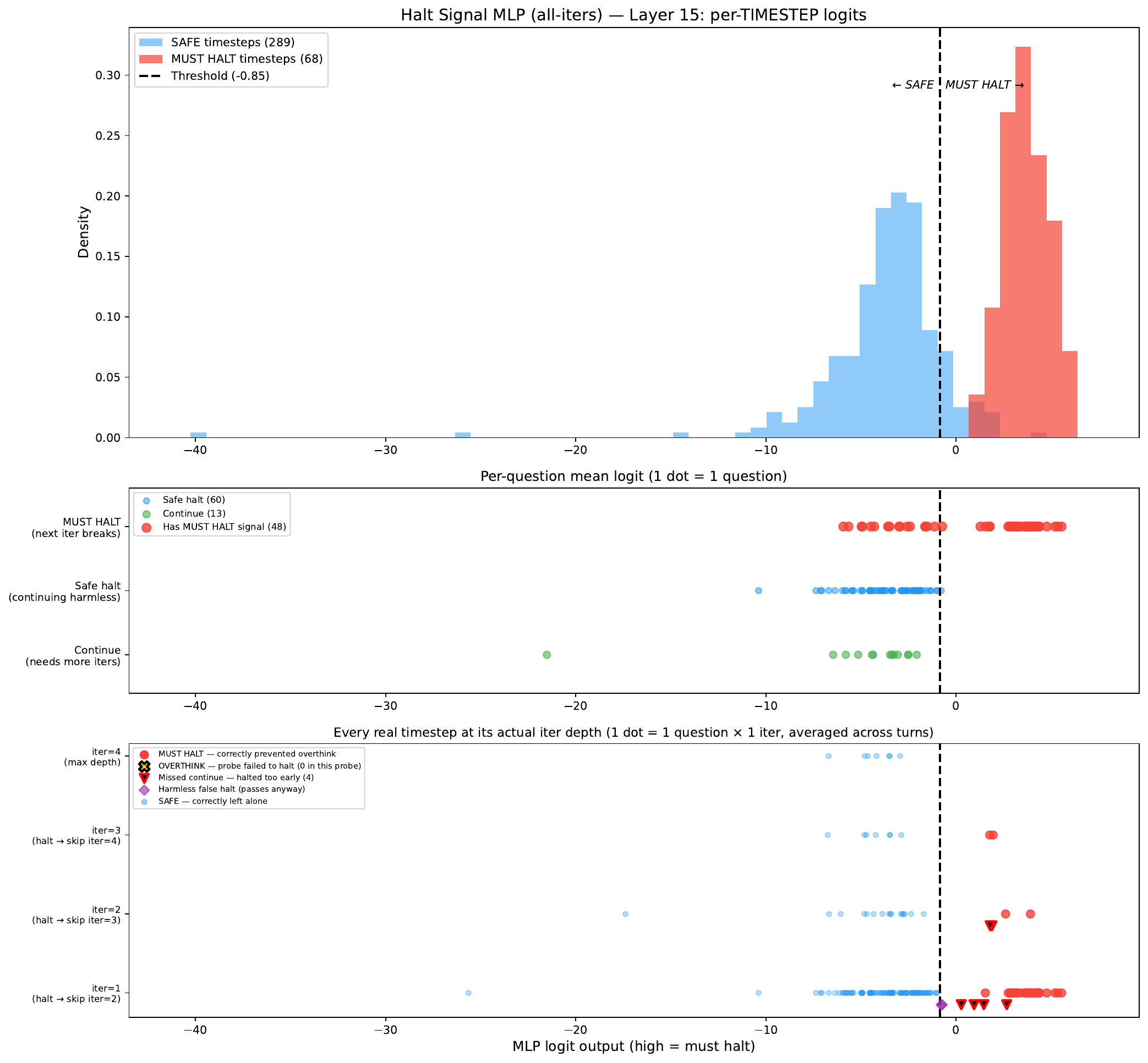}
  \caption{\textbf{Layer~15 halt signal probe output across all evaluation timesteps.} \textbf{Top:} per-timestep histogram. \textsc{must halt} timesteps (mean $+3.66$, std $1.16$) separate from \textsc{safe} timesteps (mean $-3.72$, std $3.52$) with a gap of $7.38$ between class means. \textbf{Middle:} per-question strip plot. Each point is the mean logit across that question's evaluated timesteps: questions containing a \textsc{must halt} state occupy the higher-logit region, while safe-halt and continue questions cluster below the threshold. \textbf{Bottom:} every evaluation timestep at its actual iteration depth, with markers for correctly detected halts (red), harmless false halts (purple), and missed continues (black).}
  \label{fig:halt_signal}
\end{figure}

\subsection{Ablation sweep}

\begin{table}[H]
\centering
\small
\caption{\textbf{Ablation sweep on the $64$-neuron probe.} Keep top-$N$ neurons by contribution, zero the rest. Top $10$ reproduce the full result exactly. Beyond $10$, additional neurons improve only marginal \textsc{safe} detection with no effect on question-level outcomes. The two single-component ablations below show that neither half of the top-$10$ ensemble works alone.}
\label{tab:halting_probe_ablation_sweep}
\begin{tabular}{lcccccc}
\toprule
Configuration & Correct & Accuracy & OT & Miss & \textsc{halt} det. & \textsc{safe} det. \\
\midrule
Top $10$ neurons                & $117/198$ & $59.09\%$ & $0$ & $4$ & $68/68$ & $250/289$ \\
Top $15$ neurons                & $117/198$ & $59.09\%$ & $0$ & $4$ & $68/68$ & $255/289$ \\
Top $20$ neurons                & $117/198$ & $59.09\%$ & $0$ & $4$ & $68/68$ & $257/289$ \\
Top $30$ neurons                & $117/198$ & $59.09\%$ & $0$ & $4$ & $68/68$ & $257/289$ \\
Full probe (all $64$)           & $117/198$ & $59.09\%$ & $0$ & $4$ & $68/68$ & $257/289$ \\
\midrule
Remove neuron $44$ (top-$9$ only) & $102/198$ & $51.52\%$ & $0$ & $19$ & $68/68$ & $10/289$ \\
Keep only $44$ + minor (zero other top-$9$) & $90/198$ & $45.45\%$ & $31$ & $0$ & $0/68$ & $289/289$ \\
\midrule
iter${} = 1$ baseline           & $101/198$ & $51.01\%$ & --- & --- & --- & --- \\
\bottomrule
\end{tabular}
\end{table}

\subsection{Probe evaluation procedure}
\label{app:halting_eval}

The probe is evaluated on the full $198$-question GPQA-Diamond benchmark under the same evaluation setup as Section~\ref{sec:evaluation}. For each question, the \sst{} generates a multi-turn agentic response. At each model turn, the probe inspects the position-$0$ hidden state at layer~$15$ after each iteration of the recurrence. If the probe's output logit exceeds the threshold $\log(0.3/0.7)$ (halt when $P(\text{halt}) > 0.3$), iteration halts at that depth, and that depth is used for the remainder of the question. If the probe does not halt within $i_{\max} = 4$ iterations, the model uses iter${} = i_{\max}$. This procedure is identical for the baseline evaluation, the LOOCV folds, and the input-dimension ablation conditions; the only variable is the probe's weights (for LOOCV) or the hidden-state dimensions available to the probe (for ablation).

\subsection{LOOCV methodology}
\label{app:halting_loocv}

Each of the $48$ \textsc{must halt} questions is held out in turn. For each fold:

\begin{enumerate}[leftmargin=*, itemsep=2pt]
    \item Remove all timesteps belonging to the held-out question from the training set.
    \item Train a fresh probe from scratch on the remaining timesteps (identical pipeline to Appendix~\ref{app:halting_training}: same architecture, seed, epochs, class balancing).
    \item Run the full GPQA-Diamond evaluation with the freshly trained probe (procedure in Appendix~\ref{app:halting_eval}). Record whether the held-out question is correctly classified.
\end{enumerate}

Result: $29$ of $48$ held-out \textsc{must halt} questions are correctly classified ($60\%$).

\paragraph{Null hypothesis and statistical test.} The null hypothesis is memorisation: the probe stores the training \textsc{must halt} patterns as a lookup table and reads no genuine feature of the hidden state. Under this null, the probe has no information about the held-out question (it was excluded from training). On an input it has not memorised, the probe classifies it as \textsc{must halt} at its base prediction rate.

This base rate is measured empirically: when the full probe (trained on all $357$ timesteps) is run on the complete evaluation, it classifies $133$ of $357$ timesteps as \textsc{must halt}. Of these $133$, $68$ are true \textsc{must halt} timesteps (correctly detected) and $65$ are \textsc{safe} timesteps incorrectly classified as \textsc{must halt} (false positives). The base prediction rate is therefore $133/357 = 37.3\%$.

Under the memorisation null, each LOOCV fold is an independent trial with success probability $0.373$. A one-sided binomial test of $29$ successes in $48$ trials at $p = 0.373$:

\[
P(X \geq 29 \mid X \sim \mathrm{Binomial}(48, 0.373)) = 9.4 \times 10^{-4}
\]

The memorisation null is rejected at $p < 0.001$. The probe generalises to held-out questions at a rate significantly above its base prediction rate, confirming it reads a genuine feature of the position-$0$ latent state.

\subsection{Input-dimension ablation: isolating the feature}
\label{app:halting_input_ablation}

The probe-neuron ablation of Section~\ref{sec:halting_probe_structure} establishes that $10$ of the $64$ probe neurons carry the halt signal. A separate question is which of the $5{,}376$ hidden-state dimensions these $10$ neurons read from. All ablation below is \textbf{inference-only on the frozen trained probe}: the probe is trained once on unmasked data and never retrained. For each ablation condition, the specified dimensions are zeroed in the position-$0$ hidden states before they are fed to the frozen probe during evaluation (per-question, per-turn, per-iteration, identical to Section~\ref{sec:halting_probe_result}). The results measure what the trained probe actually relies on, not what a retrained probe can learn to compensate for.

\paragraph{Step 1: constrain the range.} All $5{,}376$ dimensions are ranked by effective weight importance ($\sum_j |W_{2,j}| \cdot |W_{1,j,i}|$). For each $K$, only the top-$K$ dimensions are kept and all others are zeroed at inference. $K$ is swept to find the minimum that reproduces the baseline.

\begin{table}[H]
\centering
\small
\caption{\textbf{Top-$K$ input-dimension sweep (inference-only, frozen probe).} Keeping the top-$K$ dimensions by weight importance and zeroing the rest. $K = 761$ is the minimum that reproduces the exact baseline.}
\label{tab:halting_probe_topk_sweep}
\begin{tabular}{cccccc}
\toprule
$K$ & Correct & Accuracy & OT & Miss & Baseline match \\
\midrule
$500$ & $115/198$ & $58.08\%$ & $2$ & $4$ & no \\
$750$ & $115/198$ & $58.08\%$ & $2$ & $4$ & no \\
$\mathbf{761}$ & $\mathbf{117/198}$ & $\mathbf{59.09\%}$ & $\mathbf{0}$ & $\mathbf{4}$ & \textbf{yes} \\
$775$ & $117/198$ & $59.09\%$ & $0$ & $4$ & yes \\
$1000$ & $117/198$ & $59.09\%$ & $1$ & $3$ & no \\
$1500$ & $99/198$ & $50.00\%$ & $22$ & $0$ & no \\
$5376$ & $117/198$ & $59.09\%$ & $0$ & $4$ & yes \\
\bottomrule
\end{tabular}
\end{table}

The collapse at $K = 1500$ reflects probe weight calibration: dimensions ranked $1000$--$1500$ contribute signal that shifts the probe's threshold when included without the compensating dimensions ranked $1500$+. The boundary $K = 761$ was found by binary search between $K = 750$ and $K = 775$.

\paragraph{Step 2: greedy pruning to find the essential minimum.} Starting from the $761$-dimension set, each dimension is tested individually (least important first). For each: zero it at inference (keeping everything already pruned zeroed). If baseline holds ($117/198$, OT $= 0$, Miss $= 4$), keep it zeroed (redundant). If baseline breaks, mark it as essential and restore it.

\begin{table}[H]
\centering
\small
\caption{\textbf{Greedy pruning result.} Of $761$ candidate dimensions, $654$ are redundant and $107$ are essential. The $107$ span the full index range (dim~$9$ to dim~$5{,}364$), with $43$ in the top $100$ by weight importance and $21$ in ranks $500$--$761$.}
\label{tab:halting_probe_greedy_prune}
\begin{tabular}{lc}
\toprule
& Count \\
\midrule
Starting set (top-$761$ by weight importance) & $761$ \\
Pruned (redundant) & $654$ \\
\textbf{Essential (cannot remove)} & $\mathbf{107}$ \\
\midrule
Final verification & $117/198$, OT $= 0$, Miss $= 4$ \\
\bottomrule
\end{tabular}
\end{table}

The $107$ essential dimensions constitute $2.0\%$ of the $5{,}376$-dimensional hidden state. They are scattered across the full range (not clustered), with median weight-importance rank $166$. No individual dimension dominates: the greedy pruning removes each dimension only when the cumulative loss from prior removals makes that specific dimension's contribution critical for maintaining the threshold.

\subsection{The four missed questions}

All four errors at layer~$15$ are continue questions halted by the probe at iter${} = 1$. All four were failed by the iter${} = 1$ baseline already, so the probe's halt at iter${} = 1$ leaves each at its (wrong) iter${} = 1$ answer rather than routing to a deeper iteration where the question would have passed. None are broken correct answers. The conservative failure mode is mechanistically attributable to the cooperative-ensemble structure of Section~\ref{sec:halting_probe_structure}: by default the probe produces a halt logit sufficiently below threshold to route to continuation, and \textsc{must halt} requires the balanced contribution of the top-nine halt-detection neurons and neuron~$44$'s counterbalancing offset to cross threshold, which these four continue questions do on false-positive halt evidence at iter${} = 1$, halting before reaching the depth they needed.

\end{document}